\documentclass[10pt,twocolumn,letterpaper]{article}

\usepackage[pagenumbers]{cvpr} 

\usepackage{graphicx}
\usepackage{amsmath}
\usepackage{amssymb}
\usepackage{booktabs,caption}
\usepackage{multirow}
\usepackage{bm}
\usepackage{dsfont}
\usepackage[linesnumbered,ruled,vlined]{algorithm2e}
\usepackage{makecell}
\usepackage[flushleft]{threeparttable}
\usepackage{enumitem}
\usepackage[accsupp]{axessibility} 

\graphicspath{{figures/}}
\usepackage[pagebackref,breaklinks,colorlinks]{hyperref}

\def\ie{\emph{i.e.\ }}

\usepackage[capitalize]{cleveref}
\crefname{section}{Sec.}{Secs.}
\Crefname{section}{Section}{Sections}
\Crefname{table}{Table}{Tables}
\crefname{table}{Tab.}{Tabs.}

\begin{document}

\title{NTIRE 2022 Challenge on Efficient Super-Resolution: Methods and Results}

\author{Yawei Li$^*$ \and
Kai Zhang$^*$ \and
Radu Timofte$^*$ \and
Luc Van Gool$^*$ \and
Fangyuan Kong \and
Mingxi Li \and 
Songwei Liu \and
Zongcai Du \and
Ding Liu \and 
Chenhui Zhou \and
Jingyi Chen \and
Qingrui Han \and
Zheyuan Li \and
Yingqi Liu \and
Xiangyu Chen \and
Haoming Cai \and
Yu Qiao \and
Chao Dong \and
Long Sun \and 
Jinshan Pan \and 
Yi Zhu \and 
Zhikai Zong \and
Xiaoxiao Liu \and 
Zheng Hui \and 
Tao Yang \and
Peiran Ren \and
Xuansong Xie \and
Xian-Sheng Hua \and
Yanbo Wang \and
Xiaozhong Ji \and 
Chuming Lin \and 
Donghao Luo \and 
Ying Tai \and 
Chengjie Wang \and
Zhizhong Zhang \and
Yuan Xie \and
Shen Cheng \and
Ziwei Luo \and
Lei Yu \and
Zhihong Wen \and
Qi Wu1 \and
Youwei Li \and
Haoqiang Fan \and
Jian Sun \and
Shuaicheng Liu \and
Yuanfei Huang \and
Meiguang Jin \and 
Hua Huang \and
Jing Liu \and
Xinjian Zhang \and
Yan Wang \and
Lingshun Long \and
Gen Li \and
Yuanfan Zhang \and 
Zuowei Cao \and
Lei Sun \and
Panaetov Alexander \and
Yucong Wang \and
Minjie Cai \and
Li Wang \and 
Lu Tian \and
Zheyuan Wang \and 
Hongbing Ma \and 
Jie Liu \and
Chao Chen \and
Yidong Cai \and
Jie Tang \and
Gangshan Wu \and
Weiran Wang \and
Shirui Huang \and
Honglei Lu \and
Huan Liu \and 
Keyan Wang \and 
Jun Chen \and
Shi Chen \and
Yuchun Miao \and
Zimo Huang \and
Lefei Zhang \and
Mustafa Ayazoğlu \and 
Wei Xiong \and
Chengyi Xiong \and
Fei Wang \and
Hao Li \and 
Ruimian Wen \and
Zhijing Yang \and
Wenbin Zou    \and
Weixin Zheng \and 
Tian Ye \and
Yuncheng Zhang \and
Xiangzhen Kong \and
Aditya Arora \and
Syed Waqas Zamir \and
Salman Khan \and
Munawar Hayat \and
Fahad Shahbaz Khan \and 
Dandan Gao\and
Dengwen Zhou\and
Qian Ning \and
Jingzhu Tang \and 
Han Huang \and 
Yufei Wang \and 
Zhangheng Peng \and
Haobo Li \and
Wenxue Guan \and 
Shenghua Gong \and 
Xin Li \and 
Jun Liu \and
Wanjun Wang \and
Dengwen Zhou \and
Kun Zeng \and
Hanjiang Lin \and
Xinyu Chen \and
Jinsheng Fang
}
\maketitle

\let\thefootnote\relax\footnotetext{$^*$ Y. Li (yawei.li@vision.ee.ethz.ch, Computer Vision Lab, ETH Zurich), K. Zhang, R. Timofte, and L. Van Gool were the challenge organizers, while the other authors participated in the challenge. Appendix~\ref{sec:teams} contains the authors' teams and affiliations.
NTIRE 2022 webpage: \url{https://data.vision.ee.ethz.ch/cvl/ntire22/}. Code: \url{https://github.com/ofsoundof/NTIRE2022_ESR}.}

\begin{abstract}
This paper reviews the NTIRE 2022 challenge on efficient single image super-resolution with focus on the proposed solutions and results.
The task of the challenge was to super-resolve an input image with a magnification factor of $\times$4 based on pairs of low and corresponding high resolution images. The aim was to design a network for single image super-resolution that achieved improvement of efficiency measured according to several metrics including runtime, parameters, FLOPs, activations, and memory consumption while at least maintaining the PSNR of 29.00dB on DIV2K validation set. IMDN is set as the baseline for efficiency measurement. The challenge had 3 tracks including the main track (runtime), sub-track one (model complexity), and sub-track two (overall performance). In the main track, the practical runtime performance of the submissions was evaluated. The rank of the teams were determined directly by the absolute value of the average runtime on the validation set and test set.
In sub-track one, the number of parameters and FLOPs were considered. And the individual rankings of the two metrics were summed up to determine a final ranking in this track.
In sub-track two, all of the five metrics mentioned in the description of the challenge including runtime, parameter count, FLOPs, activations, and memory consumption were considered. Similar to sub-track one, the rankings of five metrics were summed up to determine a final ranking.
The challenge had 303 registered participants, and 43 teams made valid submissions. They gauge the state-of-the-art in efficient single image super-resolution.
\end{abstract}

\section{Introduction}
\label{sec:introduction}

Single image super-resolution (SR) aims at recovering a high-resolution (LR) image from a single low-resolution (LR) image that undergoes certain degradation process. Before the deep learning era, the problem of image SR is tackled by reconstruction-based~\cite{capel2003computer,farsiu2004fast,protter2008generalizing} and exampled-based methods~\cite{freeman2002example,timofte2013anchored,timofte2014a+,romano2016raisr}. With the thriving of deep learning, SR is frequently tackled by solutions based on deep neural networks~\cite{dong2014learning,kim2016accurate,ledig2017photo,lim2017enhanced,zhang2019deep,liang2021swinir}. 

For image SR, it is assumed that the LR image is derived after two major degradation processes including blurring and down-sampling, namely,
\begin{equation}
    \mathbf{y} = (\mathbf{x} * \mathbf{k})\downarrow_s.
\end{equation}
where $*$ denotes the convolution operation between the LR image and the blur kernel and $\downarrow_s$ is the down-sampling operation with a down-scaling factor of $\times s$. Depending on the blur kernel and the down-sampling operation, image SR could be classified into several standard problems. And among them, bicubic down-sampling with different down-scaling factors ($\times 2$, $\times 3$, $\times 4$, $\times 8$, or even $\times 16$) is the most frequently used degradation model. This classical standard degradation model allows direct comparison between different image SR methods, which also serves as a test bed to validate the advantage a newly proposed SR method. 

With the fast development of hardware technologies, it becomes possible to train much larger and deeper neural networks for image SR, which contributes significantly to the performance boost of the proposed solutions. Almost each breakthrough in the field of image SR comes with a more complex deep neural network~\cite{dong2014learning,kim2016accurate,ledig2017photo,lim2017enhanced,zhang2018residual,liang2021swinir}. Apart from the development of large models with high performance, a parallel direction is design efficient deep neural networks for single image SR~\cite{dong2016accelerating,shi2016real,li2018carn,hui2018fast,IMDN,gu2019fast,zhang2021edge}. In~\cite{dong2016accelerating,shi2016real}, the proposed networks extracted features directly from the LR images instead of the bicubic interpolation of the LR image, which saved the computation by almost a factor of $s^2$. This laid the foundation for later design of neural networks for image SR. Later works focused on the design of basic building blocks from different perspectives~\cite{li2018carn,hui2018fast,IMDN}. In~\cite{gu2019fast}, a new activation function, namely multi-bin trainable linear unit was proposed to increase the nonlinear modeling capacity of shallow network. In~\cite{zhang2021edge}, an edge-oriented convolution block is proposed for real-time SR on mobile devices.

Besides the manual design of deep neural networks, there are a plethora of works that try to improve the efficiency of deep neural networks via network pruning~\cite{he2018amc,liu2019metapruning,li2020dhp,li2021heterogeneity,li2022revisiting}, low-rank filter decomposition~\cite{zhang2016accelerating,jaderberg2014speeding,li2020group,li2019learning}, network quantization~\cite{han2015deep,courbariaux2015binaryconnect}, neural architecture search~\cite{liu2019darts,song2020efficient,wu2021neural}, and knowledge distillation~\cite{hinton2015distilling,tung2019similarity}. Among those network compression works, a couple of them have been successfully applied to image SR~\cite{li2019learning,li2020dhp,li2021heterogeneity,song2020efficient,wu2021neural}.

The efficiency of deep neural network could be measured in different metrics including runtime, number of parameters, computational complexity (FLOPs), activations, and memory consumption, which affect the deployment of deep neural network in different aspects. Among them, runtime is the most direct indicator of the efficiency of a network and thus is used as the main efficiency evaluation metric. The number of activations and parameters is related to memory consumption. And a higher memory consumption means that additional memory devices are needed to store the activations and parameters during the inference. Increased computational complexity is related to higher energy consumption, which could shorten the battery life of mobile devices. At last, the number of parameters is also related to AI chip design. More parameters mean larger chip area and increased cost of the designed AI devices.

Jointly with the 2022 New Trends in Image Restoration and Enhancement (NTIRE 2022) workshop, we organize the challenge on efficient super-resolution. The task of the challenge is to super-resolve an LR image with a magnification factor of $\times$4 by a network that reduces one or several aspects such as runtime, parameters, FLOPs, activations and memory consumption, while at least maintaining PSNR of 29.00dB on the DIV2K validation set. The challenge aims to seek advanced and novel solutions for efficient SR, to benchmark their efficiency, and identify the general trends for the design of efficient SR networks.

\section{NTIRE 2022 Efficient Super-Resolution Challenge}
This challenge is one of the NTIRE 2022 associated challenges on: spectral recovery~\cite{arad2022ntirerecovery}, 
spectral demosaicing~\cite{arad2022ntiredemosaicing},
perceptual image quality assessment~\cite{gu2022ntire},
inpainting~\cite{romero2022ntire},
night photography rendering~\cite{ershov2022ntire},
efficient super-resolution~\cite{li2022ntire},
learning the super-resolution space~\cite{lugmayr2022ntire},
super-resolution and quality enhancement of compressed video~\cite{yang2022ntire},
high dynamic range~\cite{perezpellitero2022ntire},
stereo super-resolution~\cite{wang2022ntire},
burst super-resolution~\cite{bhat2022ntire}.

The objectives of this challenge are:
(i) to advance research on efficient SR; (ii) to compare the efficiency of different methods and (iii) to offer an opportunity for academic and industrial attendees to interact and explore collaborations. This section details the challenge itself.

\subsection{DIV2K Dataset~\cite{agustsson2017ntire}}
Following~\cite{agustsson2017ntire,zhang2019aim,zhang2020aim}, the DIV2K dataset is adopted for the challenge. The dataset contains 1,000 DIVerse 2K resolution RGB images, which are divided into a training set with 800 images, a validation set with 100 images, and a testing with 100 images.
The corresponding LR DIV2K in this challenge is the bicubicly downsampled counterpart with a down-scaling factor $\times 4$. The validation set is already released to the participants.
The testing HR images are hidden from the participants during the whole challenge.

\subsection{IMDN Baseline Model}
\label{sec:baseline_model}

The IMDN~\cite{IMDN} serves as the baseline model in this challenge. The aim is to improve its efficiency in terms of runtime, number of parameters, FLOPs, number of activations, and GPU memory consumption while maintaining a PSNR performance of 29.00dB on the validation set. The IMDN model uses a $3 \times 3$ convolution to extract features from the LR RGB images, which is followed by 8 information multi-distillation blocks. The information multi-distillation block contains 4 stages that progressive refine the feature representation in the block. In each stage, the input feature from the previous stage is split along the channel dimension, leading two separated features. Among the two features, one is bypassed to the end of the block and the other one is fed to the next stage for the calculation of high-level feature. The bypassed features from the 4 stages are concatenated along the channel dimension and combined by a $1 \times 1 $ convolution. The final upsampler only consists of one trainable convolutional layer to expand the feature dimension. Pixel-shuffle is used to recover the high-resolution grid of the image. This design is considered to save as many parameters as possible.

The baseline IMDN is provided by the winner of the AIM 2019 Challenge on Constrained Super-Resolution~\cite{zhang2019aim}. The quantitative performance and efficiency metrics  of IMDN are given in \cref{table_track1} and summarized as follows. (1) The number of parameters is 0.894M. (2) The average PSNRs on validation and testing sets of DIV2K are 29.13dB and 28.78dB, respectively. (3) The runtime averaged on the validation and test set with PyTorch 1.11.0, CUDA Toolkit 10.2, cuDNN 7.6.2 and a single Titan Xp GPU is 50.86 ms. (4) The number of FLOPs for an input of size $256\times256$ is 58.53G. (5) The number of activations (\ie the number of elements  in  all  convolutional layer outputs) for an input of size $256\times256$ is 154.14M. (5) The maximum GPU memory consumption during the inference on the DIV2K validation set is 471.76M. (6) The number of convolutional layers is 43.

\subsection{Tracks and Competition}
The aim of this challenge is to devise a network that reduces one or several aspects such as runtime, parameters, FLOPs, activations and memory consumption while at least maintaining the PSNR of 29.00dB on the validation set. The challenge is divided into three tracks according to the 5 evaluation metrics.

\medskip
\noindent{\textbf{Main Track: Runtime Track.}} In this track, the practical runtime performance of the submissions is evaluated. The rankings of the teams are determined directly by the absolute value of the average runtime on the validation set and test set.

\medskip
\noindent{\textbf{Sub-Track 1: Model Complexity Track.}} In this track, the number of parameters and FLOPs are considered. And the rankings of the two metrics are summed up to determine a final ranking in this track.

\medskip
\noindent{\textbf{Sub-Track 2: Overall Performance Track.}} In this track, all of the five metrics mentioned in the description of the challenge including runtime, parameters, FLOPs, activations, and GPU memory are considered. Similar to Sub-Track 1, the rankings of five metrics are summed up to determine a final ranking in this track.

\medskip
\noindent{\textbf{Ranking statistic}}
When determine the ranking in the case of multiple metrics, the individual rankings of different metrics are summed up, which constitutes a ranking statistic of the metrics. This idea is similar to that behind Spearman's correlation. That is, instead of using the absolute values, a ranking statistic could remove the influence of unit and at the same time be good enough to distinguish different entries.

\medskip
\noindent{\textbf{Challenge phases }}
\textit{(1) Development and validation phase:} The participants had access to the 800 pairs of LR/HR training image pairs and 100 pairs of LR/HR validation images of the DIV2K dataset.  The IMDN model, pretrained parameters, and validation demo script are given on GitHub (\url{https://github.com/ofsoundof/IMDN}), allowing the participants to benchmark the runtime of their models on their system. 
The participants could upload the HR validation results on the evaluation server to calculate the PSNR of the super-resolved image produced by their models to get immediate feedback. The number of parameters and runtime was computed by the participant.
\textit{(2) Testing phase:} 
In the final test phase, the participants were granted access to the 100 LR testing images. The HR ground-truth images are hidden for the participants. The participants then submitted their super-resolved results to the Codalab evaluation server and e-mailed the code and factsheet to the organizers.
The organizers verified and ran the provided code to obtain the final results. Finally, the participants received the final results at the end of the challenge.

\begin{table*}[!t]
\caption{Results of NTIRE 2022 Efficient SR Challenge. The underscript numbers in parentheses following each metric value denotes the ranking of the solution in terms of that metric. ``Ave. Time'' is averaged on DIV2K validation and test datasets. ``\#Params'' denotes the total number of parameters. ``FLOPs'' is the abbreviation for floating point operations. ``\#Acts'' measures the number of elements of all outputs of convolutional layers. ``GPU Mem.'' represents maximum GPU memory consumption according to the PyTorch function $\texttt{torch.cuda.max\_memory\_allocated()}$ during the inference on DIV2K validation set. ``\#Conv'' represents the number of convolutional layers. ``FLOPs'' and ``\#Acts'' are tested on an LR image of size 256$\times$256. \textbf{This is not a challenge for PSNR improvement. The ``validation/testing PSNR'' and ``\#Conv'' are not ranked}.
}
\label{table_track1}
\centering
\begin{threeparttable}
\resizebox{\linewidth}{!}
{
\begin{tabular}{l||rrr|ll|llrrr|l}
\toprule
Team                  & \makecell{Main \\ Track} & \makecell{Sub- \\ Track 1}  & \makecell{Sub- \\Track 2}   & \makecell{PSNR \\ {[Val.]}} & \makecell{PSNR \\ {[Test]}} & \makecell{Ave. \\ Time [ms]}  & \makecell{\#Params \\ {[M]}}      & \makecell{FLOPs \\ {[G]}}        & \makecell{\#Acts \\ {[M]}}         &\makecell{ GPU Mem. \\ {[M]}}           & \#Conv  \\
\midrule
ByteESR               & 1          & 22$_{(11)}$ & 33$_{(2)}$   & 29.00    & 28.72     & 27.11$_{(1)}$   & 0.317$_{(11)}$ & 19.70$_{(11)}$  & 80.05$_{(6)}$   & 377.91$_{(4)}$   & 39    \\
NJU\_Jet              & 2          & 37$_{(18)}$ & 44$_{(6)}$   & 29.00    & 28.69     & 28.07$_{(2)}$   & 0.341$_{(18)}$ & 22.28$_{(19)}$  & 72.09$_{(4)}$   & 204.60$_{(1)}$   & 34    \\
NEESR                 & 3          & 10$_{(4)}$  & 27$_{(1)}$   & 29.01    & 28.71     & 29.97$_{(3)}$   & 0.272$_{(4)}$  & 16.86$_{(6)}$   & 79.59$_{(5)}$   & 575.99$_{(9)}$   & 59    \\
Super                 & 4          & 26$_{(12)}$ & 55$_{(10)}$  & 29.00    & 28.71     & 32.09$_{(4)}$   & 0.326$_{(14)}$ & 20.06$_{(12)}$  & 93.82$_{(10)}$  & 663.07$_{(15)}$  & 59    \\
MegSR                 & 5          & 18$_{(9)}$  & 43$_{(5)}$   & 29.00    & 28.68     & 32.59$_{(5)}$   & 0.290$_{(9)}$  & 17.70$_{(9)}$   & 91.72$_{(8)}$   & 640.63$_{(12)}$  & 64    \\
rainbow               & 6          & 16$_{(8)}$  & 34$_{(3)}$   & 29.01    & 28.74     & 34.10$_{(6)}$   & 0.276$_{(6)}$  & 17.98$_{(10)}$  & 92.80$_{(9)}$   & 309.23$_{(3)}$   & 59    \\
VMCL\_Taobao          & 7          & 29$_{(14)}$ & 57$_{(11)}$  & 29.01    & 28.68     & 34.24$_{(7)}$   & 0.323$_{(13)}$ & 20.97$_{(16)}$  & 98.67$_{(11)}$  & 633.00$_{(10)}$  & 40    \\
Bilibili AI           & 8          & 15$_{(7)}$  & 41$_{(4)}$   & 29.00    & 28.70     & 34.67$_{(8)}$   & 0.283$_{(8)}$  & 17.61$_{(7)}$   & 90.50$_{(7)}$   & 633.74$_{(11)}$  & 64    \\
NKU-ESR               & 9          & 12$_{(5)}$  & 48$_{(7)}$   & 29.00    & 28.66     & 34.81$_{(9)}$   & 0.276$_{(7)}$  & 16.73$_{(5)}$   & 111.12$_{(13)}$ & 662.51$_{(14)}$  & 65    \\
NJUST\_RESTORARION    & 10         & 54$_{(27)}$ & 89$_{(15)}$  & 28.99    & 28.68     & 35.76$_{(10)}$  & 0.421$_{(28)}$ & 27.67$_{(26)}$  & 108.66$_{(12)}$ & 643.95$_{(13)}$  & 52    \\
TOVBU                 & 11         & 43$_{(21)}$ & 96$_{(19)}$  & 29.00    & 28.71     & 38.32$_{(11)}$  & 0.376$_{(23)}$ & 22.38$_{(20)}$  & 113.55$_{(15)}$ & 867.17$_{(27)}$  & 64    \\
Alpan Team            & 12         & 18$_{(10)}$ & 51$_{(9)}$   & 29.01    & 28.75     & 39.63$_{(12)}$  & 0.326$_{(15)}$ & 12.31$_{(3)}$   & 115.52$_{(16)}$ & 439.37$_{(5)}$   & 132   \\
Dragon                & 13         & 38$_{(19)}$ & 70$_{(13)}$  & 29.01    & 28.69     & 41.80$_{(13)}$  & 0.358$_{(20)}$ & 21.11$_{(18)}$  & 120.15$_{(17)}$ & 260.00$_{(2)}$   & 131   \\
TieGuoDun Team        & 14         & 54$_{(27)}$ & 104$_{(21)}$ & 28.95    & 28.65     & 42.35$_{(14)}$  & 0.433$_{(29)}$ & 27.10$_{(25)}$  & 112.03$_{(14)}$ & 788.13$_{(22)}$  & 64    \\
HiImageTeam       & 15         & 7$_{(3)}$   & 70$_{(13)}$  & 29.00    & 28.72     & 47.75$_{(15)}$  & 0.242$_{(3)}$  & 14.51$_{(4)}$   & 151.36$_{(23)}$ & 861.84$_{(25)}$  & 100   \\
xilinxSR              & 16         & 66$_{(34)}$ & 107$_{(22)}$ & 29.05    & 28.75     & 48.20$_{(16)}$  & 0.790$_{(34)}$ & 51.76$_{(32)}$  & 136.31$_{(18)}$ & 471.37$_{(7)}$   & 38    \\
cipher                & 17         & 50$_{(24)}$ & 111$_{(23)}$ & 29.00    & 28.72     & 51.42$_{(17)}$  & 0.407$_{(26)}$ & 25.25$_{(24)}$  & 155.35$_{(24)}$ & 770.82$_{(20)}$  & 67    \\
NJU\_MCG              & 18         & 13$_{(6)}$  & 66$_{(12)}$  & 28.99    & 28.71     & 52.02$_{(18)}$  & 0.275$_{(5)}$  & 17.65$_{(8)}$   & 212.35$_{(27)}$ & 511.08$_{(8)}$   & 84    \\
IMGWLH                & 19         & 34$_{(17)}$ & 91$_{(17)}$  & 29.01    & 28.72     & 56.34$_{(19)}$  & 0.362$_{(21)}$ & 20.10$_{(13)}$  & 136.35$_{(19)}$ & 753.02$_{(19)}$  & 113   \\
imglhl                & 20         & 45$_{(22)}$ & 92$_{(18)}$  & 29.03    & 28.75     & 56.88$_{(20)}$  & 0.381$_{(24)}$ & 23.26$_{(21)}$  & 144.05$_{(21)}$ & 451.21$_{(6)}$   & 127   \\
whu\_sigma            & 21         & 63$_{(32)}$ & 132$_{(30)}$ & 29.02    & 28.73     & 61.04$_{(21)}$  & 0.705$_{(33)}$ & 43.88$_{(30)}$  & 142.91$_{(20)}$ & 1011.54$_{(28)}$ & 64    \\
Aselsan Research      & 22         & 27$_{(13)}$ & 98$_{(20)}$  & 29.02    & 28.73     & 63.18$_{(22)}$  & 0.317$_{(12)}$ & 20.71$_{(15)}$  & 206.05$_{(26)}$ & 799.52$_{(23)}$  & 134   \\
Drinktea              & 23         & 59$_{(31)}$ & 121$_{(27)}$ & 29.00    & 28.70     & 75.52$_{(23)}$  & 0.589$_{(31)}$ & 36.92$_{(28)}$  & 148.05$_{(22)}$ & 734.54$_{(17)}$  & 67    \\
GDUT\_SR              & 24         & 50$_{(24)}$ & 136$_{(31)}$ & 29.05    & 28.75     & 75.70$_{(24)}$  & 0.414$_{(27)}$ & 24.80$_{(23)}$  & 260.05$_{(28)}$ & 1457.98$_{(34)}$ & 195   \\
Giantpandacv          & 25         & 63$_{(32)}$ & 150$_{(34)}$ & 29.07    & 28.76     & 87.87$_{(25)}$  & 0.683$_{(32)}$ & 45.07$_{(31)}$  & 361.23$_{(31)}$ & 1272.95$_{(31)}$ & 122   \\
neptune               & 26         & 39$_{(20)}$ & 123$_{(29)}$ & 28.99    & 28.69     & 101.69$_{(26)}$ & 0.316$_{(10)}$ & 38.03$_{(29)}$  & 269.48$_{(29)}$ & 1179.05$_{(29)}$ & 45    \\
XPixel                & 27         & 3$_{(1)}$   & 49$_{(8)}$   & 29.01    & 28.69     & 140.47$_{(27)}$ & 0.156$_{(1)}$  & 9.50$_{(2)}$    & 65.76$_{(3)}$   & 729.94$_{(16)}$  & 43    \\
NJUST\_ESR            & 28         & 3$_{(1)}$   & 89$_{(15)}$  & 28.96    & 28.68     & 164.80$_{(28)}$ & 0.176$_{(2)}$  & 8.73$_{(1)}$    & 160.43$_{(25)}$ & 1346.74$_{(33)}$ & 25    \\
TeamInception         & 29         & 57$_{(30)}$ & 146$_{(33)}$ & 29.12    & 28.82     & 171.56$_{(29)}$ & 0.505$_{(30)}$ & 32.42$_{(27)}$  & 502.27$_{(34)}$ & 866.16$_{(26)}$  & 74    \\
cceNBgdd              & 30         & 33$_{(16)}$ & 114$_{(24)}$ & 28.97    & 28.67     & 180.60$_{(30)}$ & 0.339$_{(16)}$ & 21.11$_{(17)}$  & 404.16$_{(33)}$ & 739.65$_{(18)}$  & 197   \\
ZLZ                   & 31         & 55$_{(29)}$ & 118$_{(26)}$ & 29.00    & 28.72     & 183.43$_{(31)}$ & 0.372$_{(22)}$ & 64.45$_{(33)}$  & 57.51$_{(2)}$   & 1244.23$_{(30)}$ & 16    \\
Express               & 32         & 31$_{(15)}$ & 117$_{(25)}$ & 29.04    & 28.77     & 203.16$_{(32)}$ & 0.339$_{(17)}$ & 20.41$_{(14)}$  & 325.53$_{(30)}$ & 853.27$_{(24)}$  & 148   \\
Just Try              & 33         & 70$_{(35)}$ & 170$_{(35)}$ & 29.12    & 28.81     & 247.90$_{(33)}$ & 0.832$_{(35)}$ & 135.30$_{(35)}$ & 392.43$_{(32)}$ & 2387.93$_{(35)}$ & 207   \\
ncepu\_explorers      & 34         & 47$_{(23)}$ & 137$_{(32)}$ & 29.09    & 28.79     & 317.66$_{(34)}$ & 0.390$_{(25)}$ & 23.73$_{(22)}$  & 994.25$_{(35)}$ & 771.54$_{(21)}$  & 374   \\
mju\_mnu              & 35         & 53$_{(26)}$ & 121$_{(27)}$ & 29.06    & 28.79     & 332.28$_{(35)}$ & 0.345$_{(19)}$ & 78.81$_{(34)}$  & 46.76$_{(1)}$   & 1310.72$_{(32)}$ & 40    \\ \midrule

\multicolumn{12}{c}{The following methods are not ranked since their validation/testing PSNR are not on par with the baseline.}\\ \midrule

Virtual\_Reality Team &            &              &               & 27.35    & 27.26     & 2231.32          & 0.423           & 423.16           & 2731.08          & 3336.88*          & 82    \\
NTU607QCO-ESR         &            &              &               & 27.79    & 27.61     & 38.85            & 0.433           & 27.06            & 108.89           & 776.38            & 60    \\
Strong Tiger          &            &              &               & 29.00    & 28.61     & 34.92            & 0.560           & 36.64            & 78.91            & 641.13            & 23    \\
VAP                   &            &              &               & 29.01    & 28.47     & 23.96            & 0.175           & 10.83            & 70.93            & 507.64            & 63    \\
Multicog              &            &              &               & 28.38    & 28.16     & 207.98           & 0.312           & 37.67            & 430.23           & 1461.57           & 130   \\
Set5baby Team         &            &              &               & 28.92    & 28.62     & 83.44            & 0.223           & 13.98            & 229.07           & 797.25            & 88    \\
NWPU\_SweetDreamLab   &            &              &               & 28.47    & 28.23     & 31.19            & 0.193           & 11.73            & 90.50            & 633.10            & 76    \\
SSL                   &            &              &               & 28.72    & 28.44     & 64.71            & 0.290           & 18.95            & 150.60           & 675.41            & 48   \\
\midrule

RFDN AIM2020 Winner   &            &              &               & 29.04    & 28.75     & 41.97            & 0.433           & 27.10            & 112.03           & 788.13            & 64    \\ 
IMDN\_baseline        &            &              &               & 29.13    & 28.78     & 50.86            & 0.894           & 58.53            & 154.14           & 471.76            & 43    \\ \bottomrule
\end{tabular}
}
\begin{tablenotes}
\small
\item[*] This solution uses too much GPU memory. Images are cropped to $256 \times 256$ with 32 overlapping pixels during inference.
\end{tablenotes}
\end{threeparttable}
\end{table*}

\medskip
\noindent{\textbf{Evaluation protocol }}
The quantitative evaluation metrics include validation and testing PSNRs, runtime, number of parameters, number of FLOPs, number of activations, and maximum GPU memory consumed during inference. The PSNR was measured by first discarding the 4-pixel boundary around the images.
The the average runtime during the inference on the 100 LR validation images and the 100 LR testing images is computed. The best runtime among three consecutive trails is selected as the final result. The average runtime on the validation set and testing set is used as the final runtime indicator. The maximum GPU memory consumption is recorded during the inference. The FLOPs, activations are evaluated on an input image of size $256\times256$.
Among the above metrics, the runtime is regarded as the most important one. During the challenge, the participants are required to maintain the PSNR of 29.00dB on the validation set. For the final ranking, a tiny accuracy drop is tolerated. To be specific, submissions with PSNR higher than 28.95dB on the validation set and 28.65dB on the testing set could enter the final ranking. The constraint on the testing set avoids overfitting on the validation set.
A code example for calculating these metrics is available at \url{https://github.com/ofsoundof/NTIRE2022_ESR}. The code of the submitted solutions and the pretrained weights are also available in this repository.

\section{Challenge Results}
\cref{table_track1} reports the final test results and rankings of the teams. The solutions with validation PSNR lower than 28.95dB and test PSNR lower than 28.65dB are not ranked. 
The results of the baseline method IMDN~\cite{IMDN} and the overall first place winner team in AIM 2020 Efficient SR challenge~\cite{RFDN} are also reported for comparison. The methods evaluated in \cref{table_track1} are briefly described in Sec.~\ref{sec:methods_and_teams} and the team members are listed in Appendix~\ref{sec:teams}.
According to \cref{table_track1}, we can have the following observations. 

\textbf{Main Track: Runtime Track.}
First, the ByteESR team is the overall first place winner in the main track of this efficient SR challenge. The NJU\_Jet team and NEESR team win the overall second place and overall third place, respectively. The average runtime of the first three solutions is below 30 ms and is very close to each other. The first 12 teams proposed a solution with average runtime lower than 40 ms. In addition, the solution proposed by the 13-th team is also faster than the AIM 2020 winnder RFDN~\cite{RFDN}.

\textbf{Sub-Track 1: Model Complexity Track}
For this track, there are two first place winners including XPixel and NJUST\_ESR. The solution proposed by XPixel team has slightly fewer parameters while the solution proposed by NJUST\_ESR team has fewer computation. The HiImageTeam team achieves the third place in this track. The number of parameters of 9 solutions are lower than 0.3 M, which is a significant improvement compared with the AIM 2019 constrained SR challenge winner IMDN and the AIM 2020 efficient SR challenge winner RFDN. As for the computational complexity, the FLOPs of 11 solutions is lower than 20G and 27 solutions have fewer FLOPs than RFDN.

\textbf{Sub-Track 2: Overall Performance Track}
The NEESR team is the first place winner in this track. ByteESR and rainbow are the second and third place winner in this track, respectively.
The solutions proposed by mju\_mnu, ZLZ, and NJUST\_ESR team have the least number of activations. Meanwhile,
the solutions proposed by NJU\_Jet, Dragon, and rainbow team are among the most memory-efficient solutions.

The xilinxSR team achieved the best PSNR fidelity metric (29.05dB on the validation set and 29.75dB on test set) among the solutions that outperforms the baseline network IMDN in terms of runtime and number of parameters. 
When comparing this solution and the baseline solution IMDN, it is observed that IMDN has a larger PSNR improvement on the validation set than on the test set. 
On the other hand, it is also observed that, compared with the solutions proposed by TeamInception and Just Try, IMDN has similar PSNR on the validation set but achieves lower PSNR on the test set.
Such phenomenons indicate that IMDN is more in favor of the PSNR on validation set.
This is also the reason why the baseline PSNR on the validation set is set to 29.00dB rather than 29.13dB.

\subsection{Architectures and main ideas}
During this challenge, a couple of techniques are proposed to improve the efficiency of deep neural network for image SR while maintaining the performance as much as possible. Depending on the metrics that a team wants to optimize, different solutions are proposed. In the following, some typical ideas are listed.
\begin{enumerate}
    \item \textbf{Modifying the architecture of information multi-distillation block (IMDB) and residual feature distillation block (RFDB) is the mainstream technique.} The IMDB and RFDB modules come from the first place winners of the AIM 2019 constrained image SR challenge and the AIM 2020 efficient image SR challenge. Thus, some teams of this challenge start from modifying those two basic architectures. The first place winner ByteESR in the runtime main track proposed a residual local feature block (RLFB) to replace the RFDB and IMDB modules. The main difference is the removal of the concatenation operation and the associated $1\times1$ convolutional feature distillation layers. This is optimized especially out of runtime considerations. Besides, the team also reduced the number of convolutions in the ESA module.
    
    \item \textbf{Multi-stage information distillation might influence the inference speed of the super efficient models.} It is observed that the first two place solutions in the runtime main track do not contain multi-stage information distillation blocks in the network. It is also reported in other work~\cite{zhang2021edge} that using too many skip connections and associated $1\times1$ information distillation layers could harm the runtime performance.
    
    \item \textbf{Reparameterization could bring slight performance improvement.} A couple of teams tried to reparameterize a normal convolutional layer with multiple basic operations ($3\times3$ convolution, $1\times1$ operation, first and second order derivative operators, skip connections) during network training. During inference, the multiple operations that reparameterize a convolution could be merged back to a single convolution. It is demonstrated that, this technique could bring slight PSNR gain. For example, NJU\_Jet replaced a normal residual block with a reparameterized residual block. The NEESR team used edge-oriented convolution block to reparameterize a normal convolution.
    
    \item \textbf{Filter decomposition methods could effectively reduce the model complexity.} Filter decomposition method generally refers to the replacement of a normal convolution by a couple of lightweight convolutions (depthwise, $1\times1$, $1\times3$ and $3\times1$ convolutions). For example, the XPixel used the combination of depthwise convolution and $1\times1$ convolution. The NJUST\_ESR team also used the inverted residual block. And the solutions proposed by the two team won the first two places in the model complexity track.

    \item \textbf{Network pruning began to play a role.} It is observed that a couple of teams used network pruning techniques to slightly compress a baseline network. For example, the ByteESR team slightly pruned the number of channels in their network from 48 to 46 at the final training stage. The MegSR team normalized the weight parameters and applied learnable parameters to the normalized channels. They pruned the network according to the magnitude of these parameters. The xilinxSR team also tried to prune the IMDB modules.
    
    \item \textbf{Activation function is an important factor.} It is observed that some team used advanced activation function in their network. For example, the rainbow team used SiLU activation function for each convolution except the last $1\times 1$ convolution. A lot of teams also used GeLU activation function.
    
    \item \textbf{Design of loss functions is also among the consideration.} Loss function is also an important element for the success of an efficient SR network. While most of the teams used L1 or L2 loss, some teams also demonstrated using a more advanced loss function could bring marginal PSNR gain. For example, the ByteESR team used contrastive loss to improve the PSNR by 0.01dB - 0.02dB on different validation sets. The NKU-ESR team proposed edge-enhanced gradient-variance loss.
    
    \item \textbf{Advanced training strategy guarantees the performance of the network}. The advanced training strategy contains many aspects of the training setting. For example, most of the teams prolonged the training. Since the size of models is mostly small, training with both large patch size and batch size becomes possible. Periodic learning rate scheduler and cosine learning rate scheduler are used by some team, which could help the training to step outside of the local minima. The training of winner solutions typically contains several stages. Advanced tuning of the network architecture such as pruning and merging of reparameterized operations is used at the final fine-tuning stages.  
    
    \item \textbf{Various other techniques are also attempted.} Some teams also proposed solutions based on neural architecture search, vision transformers, and even fast Fourier transform.
\end{enumerate}

\subsection{Participants}
\begin{figure}[!ht]
    \centering
    \includegraphics[width=1.0\linewidth]{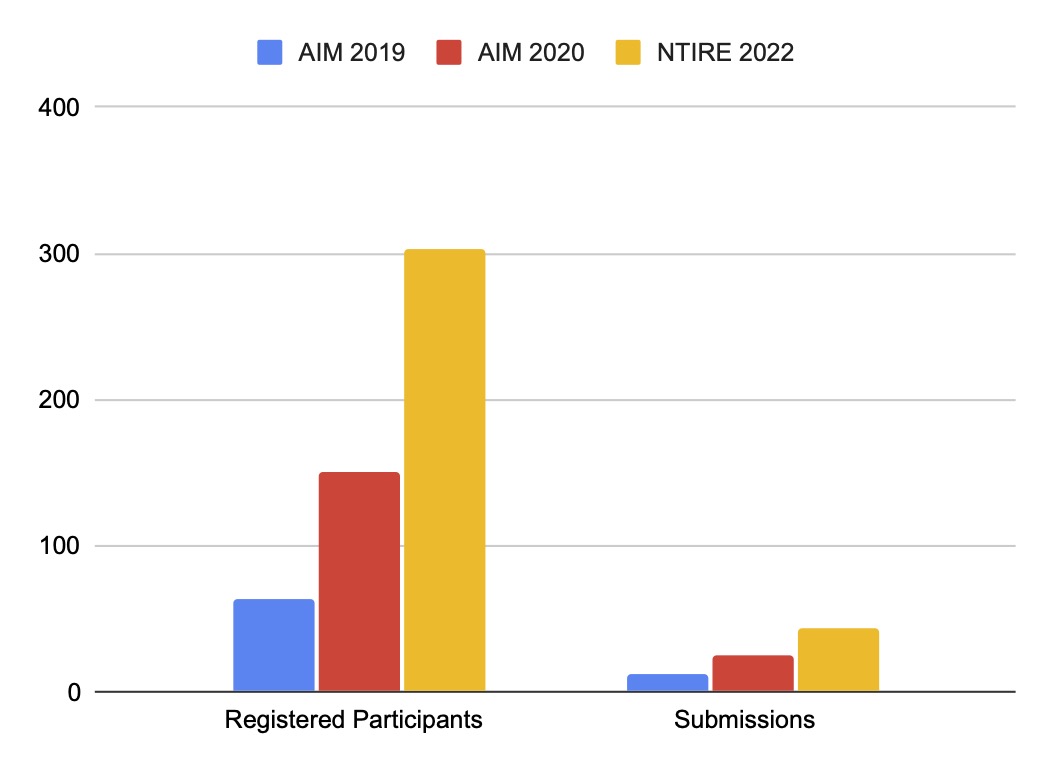}
    \caption{Number of participants and valid submission during the past three challenges.}
    \label{fig:participants}
\end{figure}

This year we see a continuous growth of the efficient SR challenge with more participants and valid submission. As shown in \cref{fig:participants}, the number of registered participants grows from 64 in AIM 2019, 150 in AIM 2020, and finally 303 in this year. Meanwhile, the number of valid submission also grows from 12 in AIM 2019, 25 in AIM 2020, and 43 this year. 

\subsection{Fairness}
To maintain the fairness of the efficient SR challenge, a couple of rules about fair and unfair tricks are set. Most of the rules are about the dataset used to train the network.
First, training with additional external dataset such as Flickr2K is allowed. Secondly, training with the additional DIV2K validation set including either of the HR or LR images is not allowed. This is because the validation set is used to examine the overall performance and generalizability of the proposed network. Thirdly, training with the DIV2K test LR images is not allowed. Fourthly, training with advanced data augmentation strategy during training is regarded as a fair trick. 

\begin{figure*}[!htb]
  \centering
  \subfloat[][]{
    \includegraphics[width=0.30\linewidth]{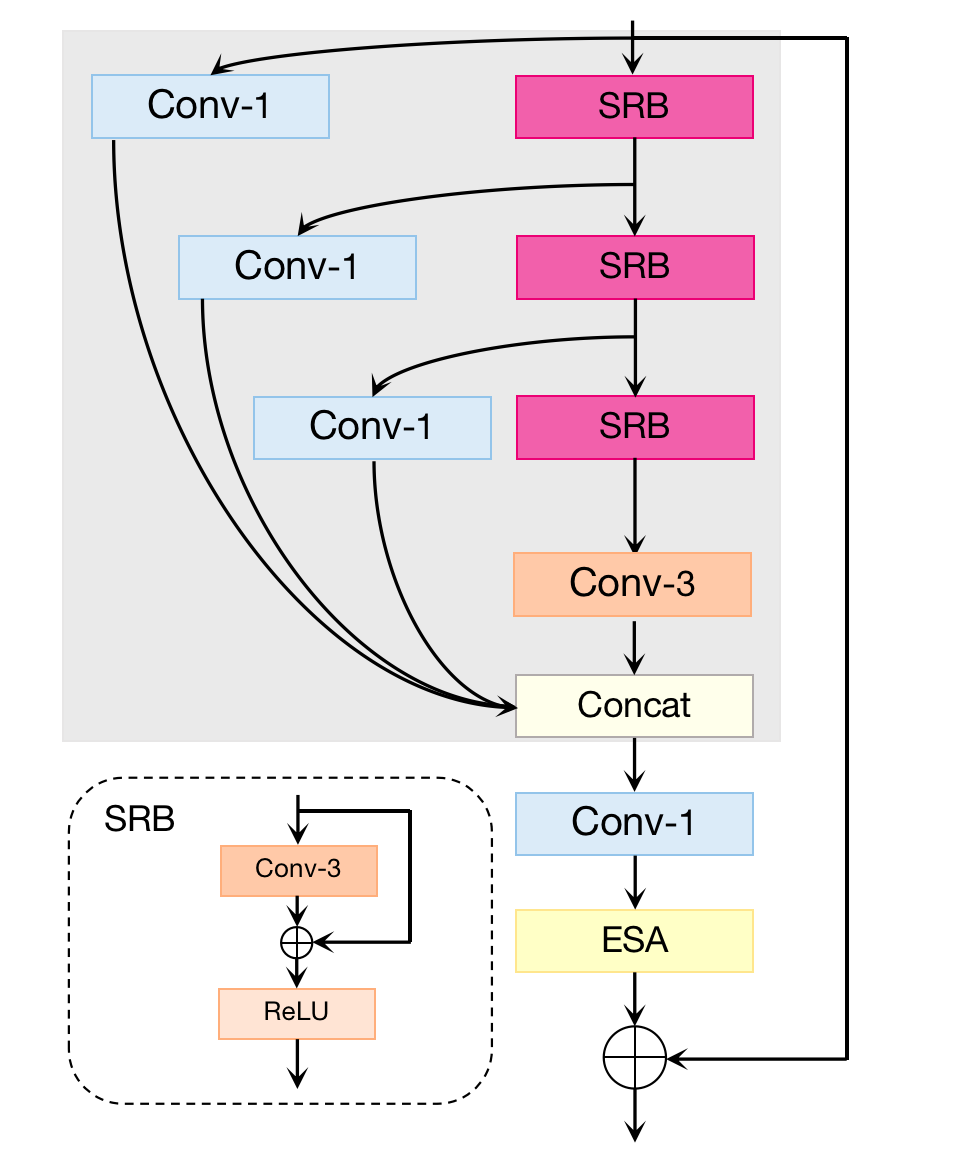}
    \label{fig:RFDBarchitecture}
  }
  \qquad
  \subfloat[][]{
    \includegraphics[width=0.20\linewidth]{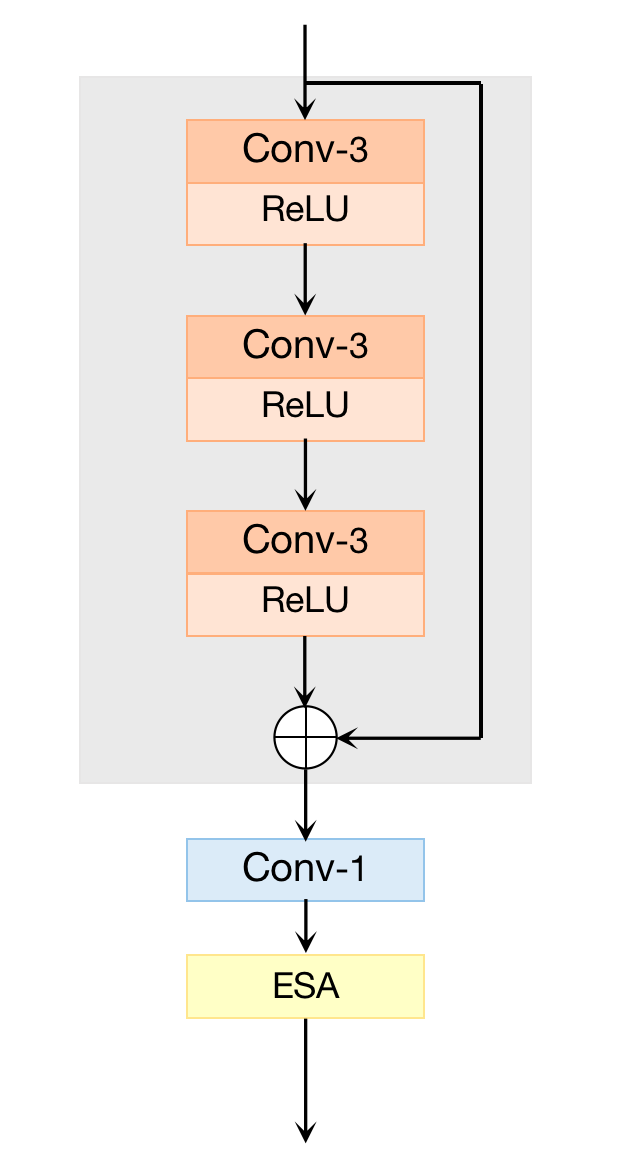}
    \label{fig:RLFBarchitecture}
  }
  \qquad
  \subfloat[][]{
    \includegraphics[width=0.15\linewidth]{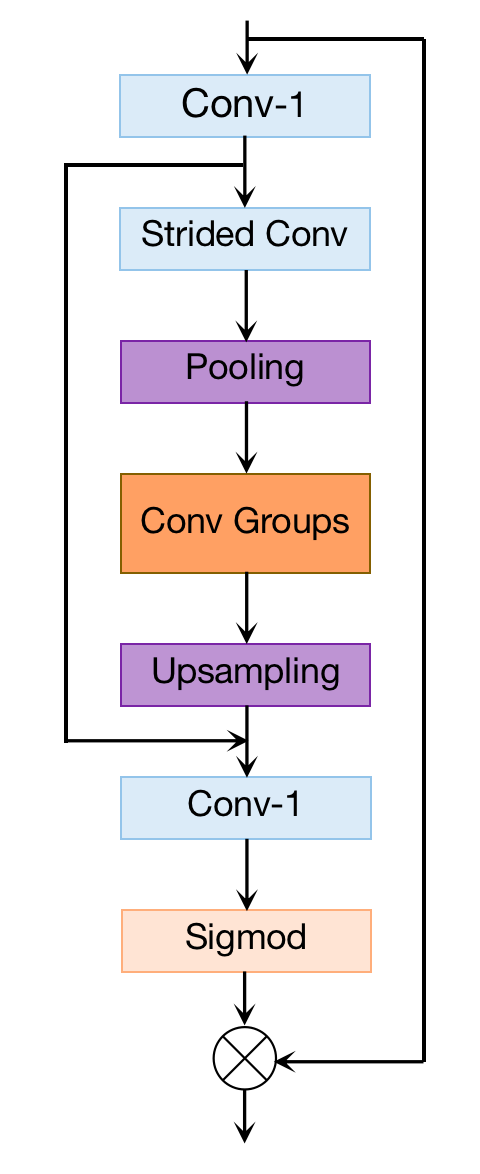}
    \label{fig:ESAarchitecture}
  }
  \caption{\textit{ByteESR Team:} (a) Residual feature distillation block (RFDB). (b) Residual local feature block (RLFB). (c) Enhanced Spatial Attention (ESA).}
\label{fig:blockarchitecture}
\end{figure*}

\subsection{Conclusions}
Based on the aforementioned analysis of the efficient SR challenge results, the following conclusions can be drawn.

\begin{enumerate}
    \item The efficient image SR community is growing. This year the challenge had 303 registered participants, and received 43 valid submissions, which is a significant boost compared with the previous years.
    
    \item The family of the proposed solutions during this challenge keep to push the frontier of the research and implementation of efficient images SR. 
    
    \item In conjunction with the previous series of the efficient SR challenge including AIM 2019 Constrained SR Challenge~\cite{zhang2019aim} and AIM 2020 Efficient SR Challenge~\cite{zhang2020aim}, the proposed solutions make new records of network efficiency in term of metrics such as runtime and model complexity while maintain the accuracy of the network.
    
    \item There is a divergence between the actual runtime and theoretical model complexity of the proposed networks. This shows that the theoretical model complexity including FLOPs and the number of parameters do not correlate well with the actual runtime at least on GPU infrastructures.
    
    \item In the meanwhile, new developments in the efficient SR field are also observed, which include but not limited to the following aspects.
    \begin{itemize}
        \item The effectiveness of multi-stage information distillation mechanism is challenged by the first two place solutions in the runtime main track. 
        \item Other techniques such as contrastive loss, network pruning, and convolution reparameterization began to play a role for efficient SR.
    \end{itemize}

\end{enumerate}

\section{Challenge Methods and Teams}
\label{sec:methods_and_teams}

\subsection{ByteESR}

\begin{figure}[!ht]
  \centering
  \includegraphics[width=0.98\linewidth]{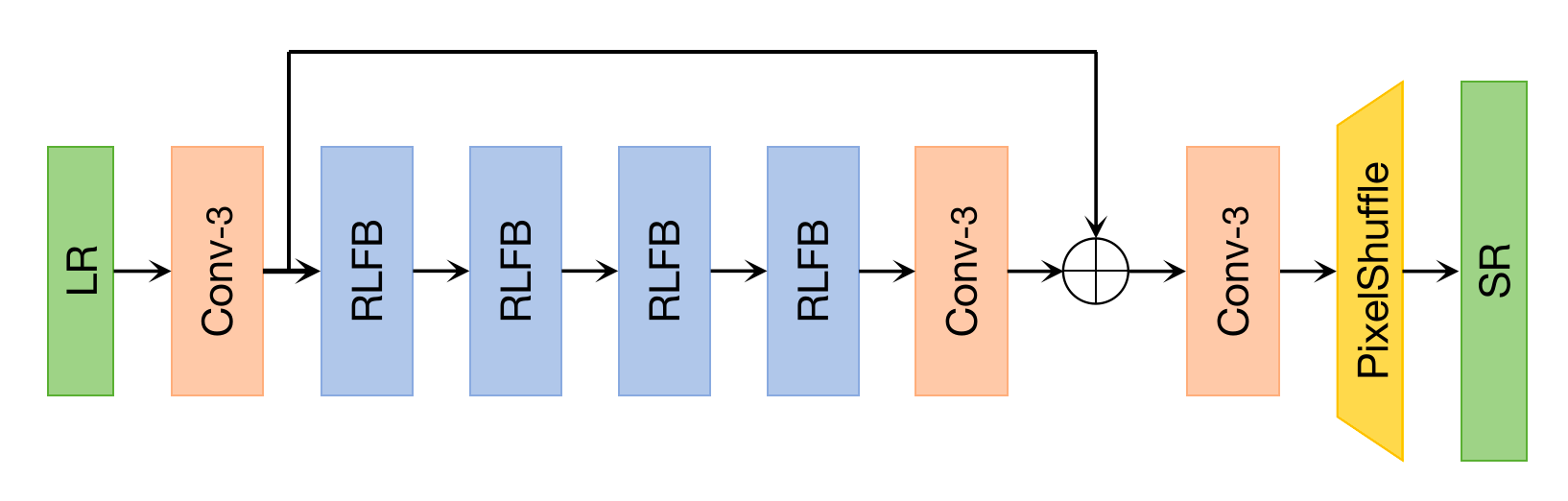}
   \caption{\textit{ByteESR Team:} The architecture of residual local feature network (RLFN).}
   \label{fig:modelarchitecture}
\end{figure}

\textbf{Network Architecture}
The ByteESR Team proposed Residual Local Feature Network  (RLFN) for Efficient Super-Resolution. As shown in \cref{fig:modelarchitecture}, the proposed RLFN uses one of the basic SR architecture, which is similar to IMDN~\cite{IMDN} and other methods~\cite{lim2017enhanced,ESRGAN}. The difference is RLFN uses four residual local feature block (RLFB) as the building blocks.

The proposed RLFN is modified from residual feature distillation block (RFDB)~\cite{RFDN}. 
As shown in \cref{fig:RFDBarchitecture}, RFDB uses three Conv-1 for feature distillation, and all the distilled features are concatenated together. Although aggregating multiple layers of distilled features can result in more powerful feature, concatenation accounts for most of the inference time. Based on the consideration of reducing inference time and memory, RLFB (see \cref{fig:RLFBarchitecture}) removes the concatenation layer and the related feature distillation layers and replaces them with an addition for local feature learning. Besides, in RLFB, the Conv Groups in ESA~\cite{RFANet} (see \cref{fig:ESAarchitecture}) is simplified to one Conv-3 to decrease the model depth and complexity.

    \begin{figure*}[htbp]
	    \centering
	    \includegraphics[width=0.95\linewidth]{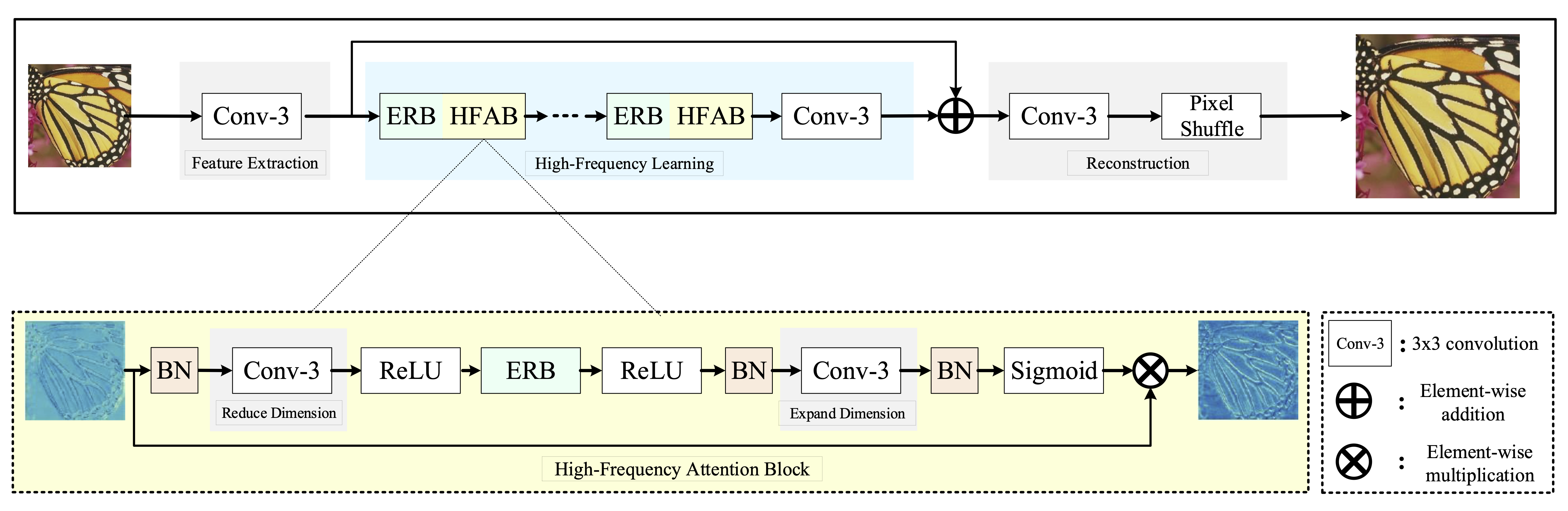} 
	    \caption{\textit{NJU\_Jet Team}: The overall architecture of the fast and memory efficient network (FMEN).}
	    \label{fig:architecture}
    \end{figure*}

\textbf{Contrastive Loss}
Some recent works~\cite{GenericPerceptualLoss, RandomStyleTransfer} find that a randomly initialized feature extractor, without any training, can be used to improve the performance of models on several dense prediction tasks. Inspired by these works, RLFN builds a two-layer network as the feature extractor. The weights of convolution kernels are randomly initialized. The contrastive loss is defined as:
\begin{equation}
    \label{eqn: cl}
    CL = \frac{\|\phi(y_{sr}) - \phi(y_{hr})\|}{\|\phi(y_{sr}) - \phi(y_{lr})\|}
\end{equation}
where $\phi$ defines the feature map generated by the feature extractor, $\|\phi(y_{sr}) - \phi(y_{hr})\|$ is the L1 distance loss between feature maps of $y_{sr}$ and $y_{hr}$ and $\|\phi(y_{sr}) - \phi(y_{lr})\|$ is the L1 distance loss between feature maps of $y_{sr}$ and $y_{lr}$.

\textbf{Implementation details.} 
The proposed RLFN has four RLFBs, in which the number of feature channels is set to 48 while the channel number of ESA is set to 16. During training, DIV2K~\cite{agustsson2017ntire} and Flickr2K datasets are used for  the whole process. The details of training steps are as follows:

\begin{itemize}[nosep]
  \item [I.] 
  At the first stage, the model is trained from scratch. HR patches of size $256\times256$ are randomly cropped from HR images, and the mini-batch size is set to 64. The RLFN model is trained by minimizing L1 loss function with Adam optimizer. The initial learning rate is set to $5 \times 10 ^{-4}$ and halved at every 200 epochs.The total number of epochs is 1000.     
  \item [II.]
  At the second stage, the model is initialized with the pretrained weights, and trained with the same settings as in the previous step. This process repeats twice.
  \item [III.]
  At the third stage, the model is initialized with the pretrained weights from Stage 2. The same training settings as Stage 1 are kept to train the model, except that the loss function is replaced by the combination of L1 loss and contrastive loss with a regularization factor $\times 255$.
  \item [IV.]
  At the fourth stage, the number of Conv-1 of RLFBs in the pretrained model from 48 to 46 using Soft Filter Pruning~\cite{SFP}. Training settings are the same as Stage 1, except that the size of HR patches changes to $512\times512$. After 1000 epochs, L2 loss is used for fine-tuning with $640\times640$ HR patches and a learning rate of $10^{-5}$. 

\end{itemize}

\subsection{NJU\_Jet}

   Runtime and memory consumption are two important aspects for efficient image super-resolution (EISR) models to be deployed on resource-constrained devices. 
   Recent advances in EISR\cite{IMDN,RFDN} exploit distillation and aggregation strategies with plenty of channel split and concatenation operations to make full use of limited hierarchical features.
   By contrast, sequential network operations avoid frequently accessing preceding states and extra nodes, and thus are beneficial to reducing the memory consumption and runtime overhead.
   Following this idea, the team designed a lightweight network backbone by mainly stacking multiple highly optimized convolution and activation layers and decreasing the usage of feature fusion. The overall network architecture is shown in \cref{fig:architecture}. The feature extraction part and reconstruction part are the same as recent works~\cite{IMDN, RFDN}, and the high-frequency learning part is composed of the proposed enhanced residual block (ERB) and high-frequency attention block (HFAB) pairs.
   %
    \begin{figure}[!ht]
	    \centering
	    \includegraphics[width=1.0\linewidth]{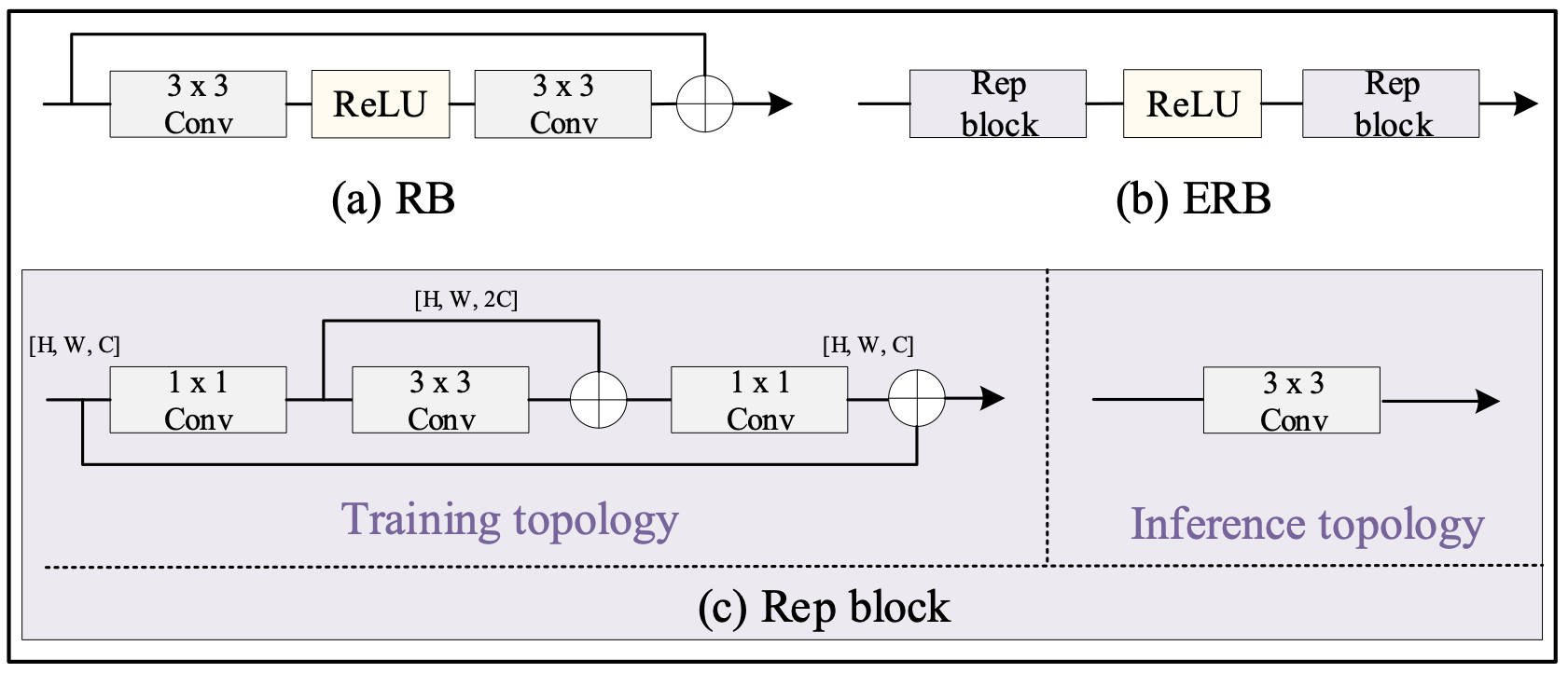} 
	    \caption{\textit{NJU\_Jet Team}: Comparison between a normal residual block (RB)~\cite{lim2017enhanced} and an enhanced residual block (ERB).}
	    \label{fig:ERB}
    \end{figure}

   \textbf{Enhanced residual block.} The team first proposed enhanced residual block (ERB) to replace normal residual block (RB) in EDSR~\cite{lim2017enhanced}, for reducing the memory access cost (MAC) introduced by skip connection. As shown in \cref{fig:ERB}, ERB is composed of two re-parameterization blocks (RepBlock) and one ReLU. During training, each RepBlock utilizes $1\times1$ convolution to expand or reduce the number of feature maps and adopts $3\times3$ convolution to extract features in higher dimensional space. Besides, two skip connections are used to ease training difficulty. During inference, all the linear transformations can be merged~\cite{ding2021repvgg}. Thus each RepBlock can be converted into a single $3\times3$ convolution. In general, ERB takes advantage of residual learning without additional MAC.
   
   \textbf{High-frequency attention block.} Recently, attention mechanism has been extensively studied in the SR community. Based on the grain-size composition, it can be divided into channel attention~\cite{SENet, IMDN}, spatial attention~\cite{CBAM}, pixel attention~\cite{PAN}, and layer attention~\cite{niu2020single}. Previous attention-based methods lack consideration of two important aspects. First, some attention schemes, such as CCA~\cite{IMDN} and ESA~\cite{RFDN}, have multi-branch topology, which introduces too much extra memory consumption. Second, some nodes in the attention branch are not computationally optimal, such as $7\times 7$ convolution used in ESA~\cite{RFANet}, which is much less efficient than $3\times3$ convolution.
   
   Considering both aspects, a sequential attention branch is designed to rescale each position based on its nearby pixels. The attention branch is inspired by edge detection, where the linear combination of nearby pixels can be used to detect edges. 
   Furthermore, the team found out that the attention branch focused on high-frequency areas and named the proposed block as high-frequency attention block (HFAB) shown in \cref{fig:architecture}. HFAB rescales every position according to the non-linear combination of its window. In HFAB, $3\times3$ convolution rather than $1\times1$ convolution is used to reduce and expand feature dimension for larger receptive field and efficiency. Batch normalization (BN) is adopted in the attention branch to introduce global statistics and to keep features within the unsaturated area of sigmoid function. During inference, BN can be merged into convolution without additional computational cost.

   \textbf{Implementation details.} DIV2K and Flickr2K are used as the training dataset. Five ERB-HFAB pairs are stacked sequentially. The number of feature maps of ERB and HFAB is set to 50 and 16, respectively. In each training batch, 64 HR RGB patches are cropped with size $256 \times 256$ and augmented by random flipping and rotation. The learning rate is initialized as $5\times 10^{-4}$ and decreases by half per $2\times10^5$ iterations. The network is trained for $10^6$ iterations in total by  minimizing L1 loss function with Adam optimizer. The team loaded the trained weights and repeated the above training setting for 4 times. After that, L2 loss is used for fine-tuning. The initial learning rate is set to $1\times10^{-5}$ for $2\times10^5$ iterations. Finally, the reconstruction part is fine-tuned with batch size 256 and HR patch size 640 for $10^5$ iterations, with L2 loss.

\begin{figure}[!t]
\centering
\includegraphics[trim={0 0 940  0},clip,width=0.6\linewidth]{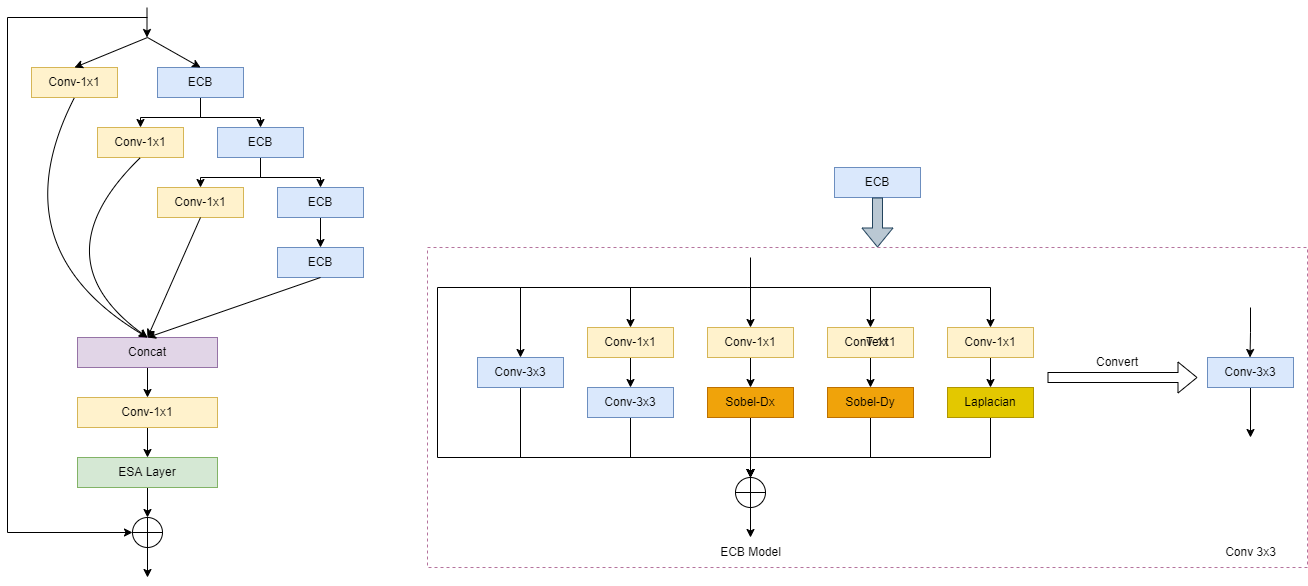}
\caption{\textit{NEESR Team:} Detailed architecture of RFDECB.}
\label{fig:neesr_RFDECB}
\end{figure}

\begin{figure}[!t]
\centering
\includegraphics[trim={420 0 0  150},clip,width=1.0\linewidth]{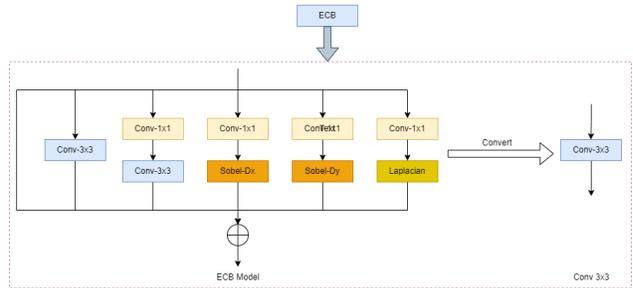}
\caption{\textit{NEESR Team:} Details of the edge-oriented convolution block (ECB).  In the inference stage, the ECB module will be converted into a single standard $3 \times 3$ convolution layer.}
\label{fig:neesr_ECB}
\end{figure}

\begin{figure*}[!ht]
  \centering
    \includegraphics[width=0.85\linewidth]{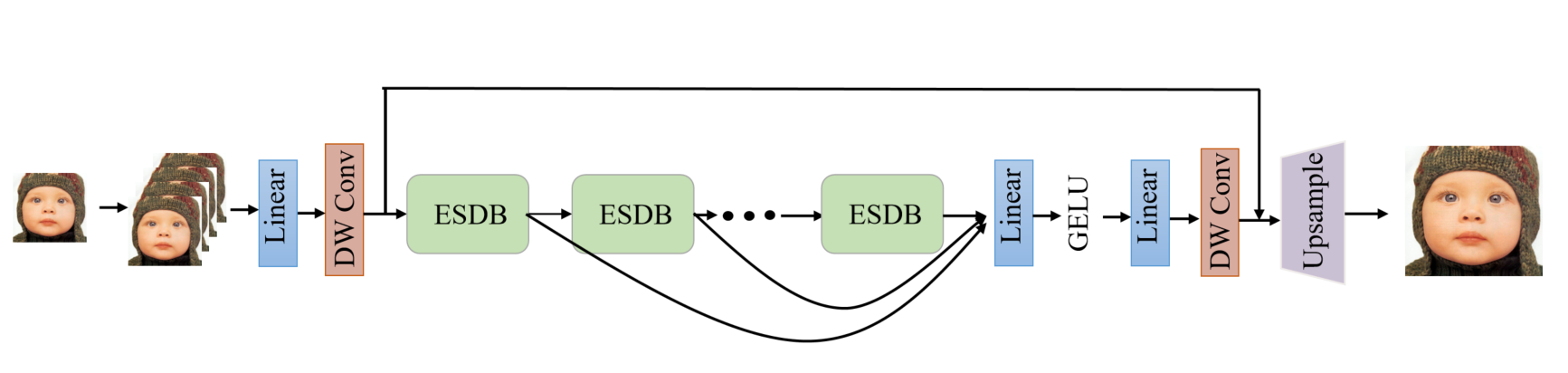}
  \caption{\textit{XPixel Team:} The architecture of blueprint separable residual network (BSRN).}
  \label{fig:xpixel_BSRN_arch}
\end{figure*}

\subsection{NEESR}

The NEESR team proposed edge-oriented feature distillation network (EFDN) for lightweight image super resolution. The proposed EFDN 
is modified from RFDN~\cite{RFDN} with some efficiency improvement considerations such as less channels and the replacement of the shallow residual block (SRB) with edge-oriented convolution block (ECB). 
Different from RFDN, EFDN only uses 42 channels to further accelerate the network. Inspired by ECBSR~\cite{zhang2021edge}, EFDN employs the re-parameterization technique to boost the SR performance while maintaining high efficiency. EFDN has four RFDECBs shown in~\cref{fig:neesr_RFDECB}.
In the training stage, RFDECB utilizes the ECB module which consists of four types of carefully designed operators including normal $3 \times 3$ convolution, channel expanding-and-squeezing convolution,
first and second order spatial derivatives from intermediate features.
Such a design can extract edge and texture information more effectively. In the inference stage, the ECB module is converted into a single standard $3 \times 3$ convolution layer. \cref{fig:neesr_ECB} illustrates the fusion
process. The training process contains two stages with three steps.

\begin{enumerate}
    \item[I.] At the first stage, the ECB module is equipped with multiple branches.

    \textbf{Pre-training on DIV2K+Flickr2K (DF2K).} HR patches of size $256 \times 256$ are randomly cropped from HR images, and the mini-batch size is set to 64. The original EFDN model is trained by minimizing L1 loss function with Adam optimizer. The initial learning rate is set to $6 \times 10 ^{-4}$ and halved at every 200 epochs. The total number of epochs is 4000.

    \textbf{Fine-tuning on DF2K.} HR patch size and the mini-batch size are set to $1024 \times 1024$ and 256, respectively. The EFDN model is fine-tuned by minimizing L2 loss function. The initial learning rate is set to $2.5 \times 10 {-4}$ and halved at every 200 epochs. The total number of epochs is 4000.

    \item[II.] At the second stage, the final plain EFDN model is obtained by converting ECB module into a single $3 \times 3$ convolution layer.
    
    \textbf{Fine-tuning on DF2K.} HR patch size is set to $1024 \times 1024$ and the mini-batch size is set to 256. The final EFDN model is fine-tuned by minimizing L2 loss function. The initial learning rate is set to $5 \times 10 ^ {-6}$ and halved at every 200 epochs. The total number of epochs is 4000.
\end{enumerate}

\begin{figure}[!ht]
  \centering
    \includegraphics[width=1\linewidth]{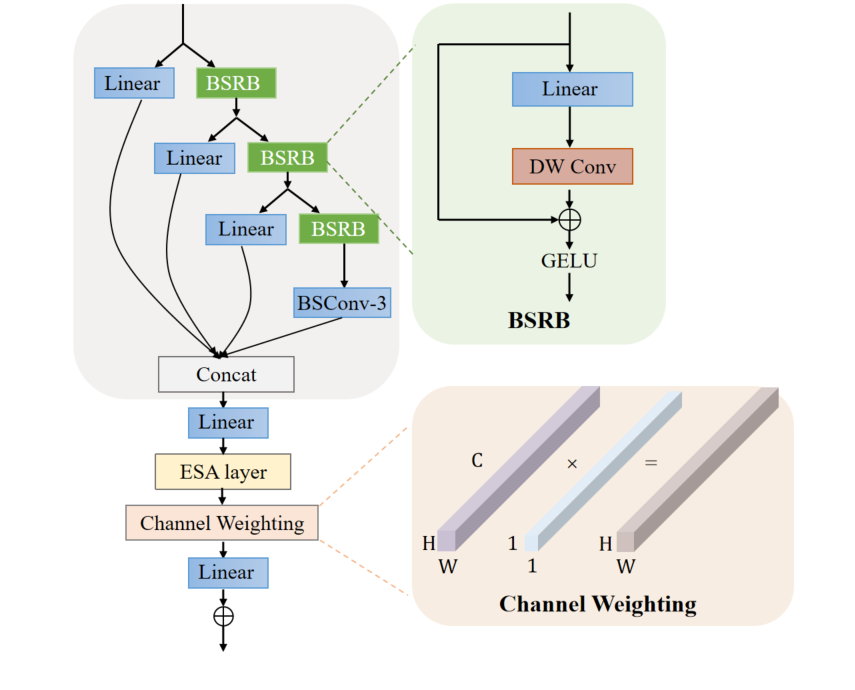}
  \caption{\textit{XPixel Team:} The architecture of the proposed efficient separable residual blocks (ESDB).}
  \label{fig:xpixel_ESDB}
\end{figure}

\subsection{XPixel}

\textbf{General method description.} 
The XPixel team proposed Blueprint Separable Residual Network (BSRN) as shown in \cref{fig:xpixel_BSRN_arch}. 
Following the overall architecture of RFDN,  BSRN consists of four stages including the shallow feature extraction, the deep feature extraction, multi-layer feature fusion, and reconstruction.
In the first stage, the shallow feature extraction contains input replication followed by a linear mapping and a depthwise convolution to map from the input image space to a higher dimensional feature space.
Then stacked efficient separable residual blocks (ESDB) build up the deep feature extraction to gradually refine the extracted features.
Features generated by each ESDB are fused along the channel dimension at the end of the trunk.
Finally, the SR image is produced by the reconstruction module, which only consists of a $3\times3$  convolution and a non-parametric sub-pixel operation.

\begin{figure*}
	\begin{center}
		\includegraphics[width=0.6\linewidth]{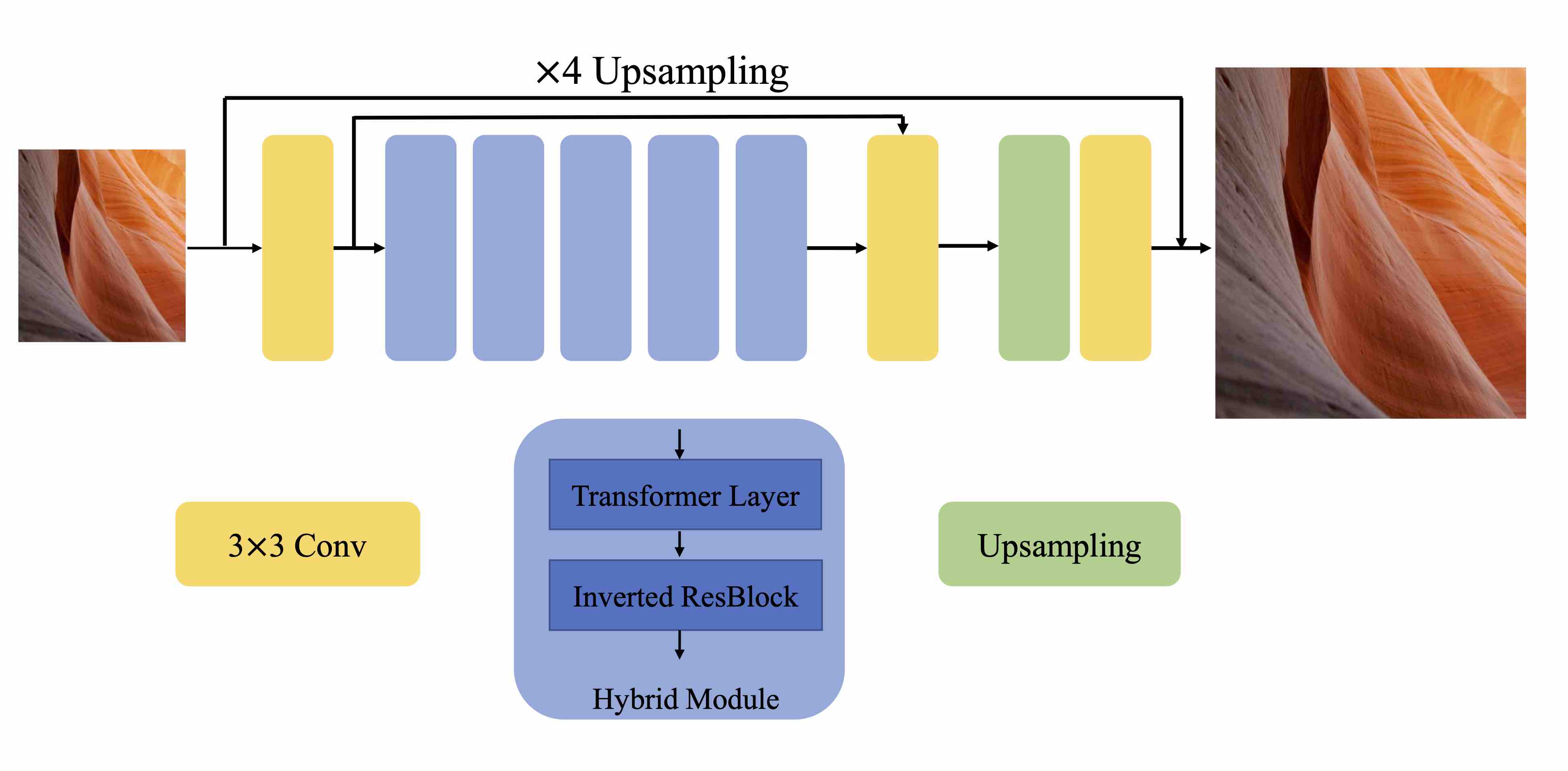}\\
		\caption{\textit{NJUST\_ESR Team}: The architecture of the proposed MobileSR model.}
		\label{fig:mobilesr}
	\end{center}
\end{figure*} 

\textbf{Building block description.}
For further optimization, the blueprint separable convolution is adopted, which is an extremely efficient decomposition of the convolution, to replace the regular convolution in the proposed blueprint separable residual block (BSRB), as shown in \cref{fig:xpixel_ESDB}.
RFDN replaces the contrast-aware channel attention (CCA) layer with the enhanced spatial attention (ESA) block for better performance. Yet, it has been found that the channel-wise feature rescaling is effective for shallow SR models to boost reconstruction accuracy.
Therefore, a channel weighting layer is involved in each ESDB for modelling channel-wise relationships to utilize inter-dependencies among channels with slightly additional cost.

\textbf{Model Details.}
The proposed BSRN model contains 5 ESDBs. the overall framework follows the pipeline of RFDN.
A global feature aggregation is employed at the end of the network body to aggregate the final features, which is set to 48.
Correspondingly, the channel weighting matrix is set to $ 1\times 1\times 48$ to match the dimension, which is initialized by normal distribution with $\sigma=0.9$, $\mu=1$.

\textbf{Training strategy.}
Data augmentation including random rotation by 90$^{\circ}$, 180$^{\circ}$, 270$^{\circ}$ and horizontal flipping  is performed on the DIV2K and Flickr2K training images.
In each training batch, 72 LR color patches with the size of $64\times64$ are extracted as inputs per GPU. 
The model is trained by ADAM optimizor with $\beta_1=0.9, \beta_2=0.999$.
The initial leaning rate is set to $5\times10^{-4}$ equipped with cosine learning rate decay.
Different from the recent SR models, $L2$ loss is used for training from scratch for $1\times10^{6}$ iterations.
The model is implemented by Pytorch 1.9.1 and trained with 4 GeForce RTX 2080ti GPUs.

\subsection{NJUST\_ESR}

The NJUST\_ESR introduced a vision transformer (ViT) based method for efficient SR, which combines the merits of convolution and multi-head self-attention. Specifically, a hybrid module containing a ViT block~\cite{li2022uniformer} and a inverted residual block~\cite{sandler2018mobilenetv2} is employed to simultaneously extract local and global information. This module is stacked multiple times to learn discriminative feature representation. 

The network consists of three stages as detailed in \cref{fig:mobilesr}. First, a convolution layer maps the input image to feature space. Then five hybrid blocks are stacked to learn discriminative feature representation. At last, there are several convolution layers and pixel shuffle layers to generate the HR image.

The proposed MobileSR model is trained on DIV2K dataset. The input patches of size  64$\times$64 are randomly cropped from LR images, and the mini-batch size is set to 64. MobileSR is trained by minimizing L1 loss and the frequency loss\cite{MIMO} with Adam optimizer. The initial learning rate is set to $5\times 10^{-4}$ and halved at every 100 epochs for total 450 epochs. 

\subsection{HiImageTeam}

\begin{figure*}[!ht]
\flushleft 
\includegraphics[width=1\textwidth]{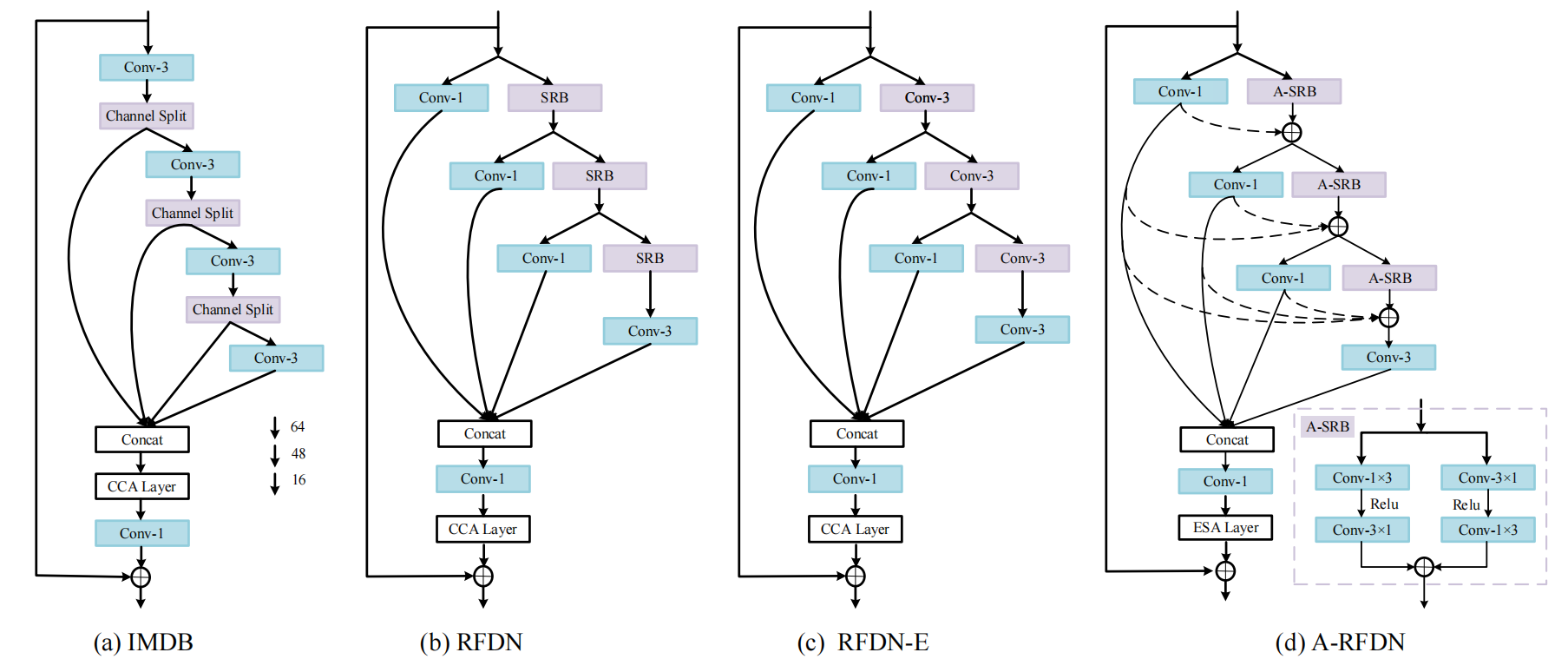}
\caption{\textit{HiImageTeam Team}: (a) The original information multi-distillation block (IMDB). (b)  Residual feature distillation block (RFDB). (c) The equivalent architecture of the RFDB (RFDB-E). (d) Asymmetric residual feature distillation block (A-RFDB).}
\label{fig:HiImageTeam_ARFDN}
\end{figure*}

The HiImageTeam team proposed Asymmetric Residual Feature Distillation Network (ARFDN) inspired by the IMDN~\cite{IMDN} and RFDN~\cite{RFDN} for efficient SR. IMDN~\cite{IMDN} is an efficient network architecture for image SR. 
Yet, there are still many redundant calculation and inefficient operators as shown in \cref{fig:HiImageTeam_ARFDN}a.  
Compared with IMDB, RFDB has a significant efficiency improvement in the calculation as shown in \cref{fig:HiImageTeam_ARFDN}b. 
A shallow residual block (SRB) in RFDN is equivalent to a normal convolution with sufficient training. The equivalent architecture of RFDB is shown in \cref{fig:HiImageTeam_ARFDN}c.
That is to say, there is also redundant calculation in RFDN. An efficient attention module, namely ESA module is used in the enhancement stage of distillation information of RFANet~\cite{RFANet}. Therefore, this module is also used in the proposed network. 
As shown in \cref{fig:HiImageTeam_ARFDN}d, an asymmetric residual feature distillation block is designed, which consists of both the asymmetric information distillation and the information recombination and utilization operation.

There are 4 ARFDBs in the proposed ARFDN, the overall framework follows the pipeline of RFDN~\cite{RFDN}, where global feature aggregation is used to augment the final feature and the number of feature channels is set to 50. 
HR patches are set to 256 $\times$ 256 and randomly cropped from HR images during the training of ARFDN. 
The mini-batch size is set to 36. 
The overall training process is divided into two stages. 
In the first stage, the ARFDN model is trained for 1000 epochs by minimizing L1 loss function with Adam optimizer and the initial learning rate is set to $2 \times$ $10^{-4}$  and halved at every 30 epochs. L2 loss function is used to fine-tune the network with learning rate of $1\times$ $10^{-4}$ in the second stage. Div2K, OST and Flickr2K datasets are used to train the ARFDN model.

\begin{figure}[htpb]
	\centering
	\includegraphics[width=0.48\textwidth]{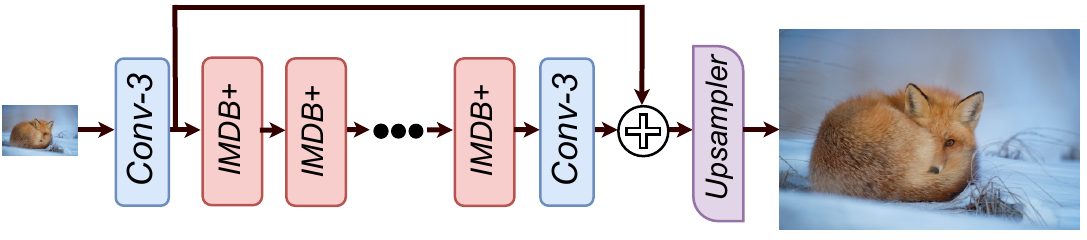}
	\caption{\textit{rainbow Team:} The architecture of improved information multi-distillation network (IMDN+). The number of IMDB+ is $8$.}
	\label{fig:IMDN+}
\end{figure}

\begin{figure}[htpb]
	\centering
	\includegraphics[width=0.48\textwidth]{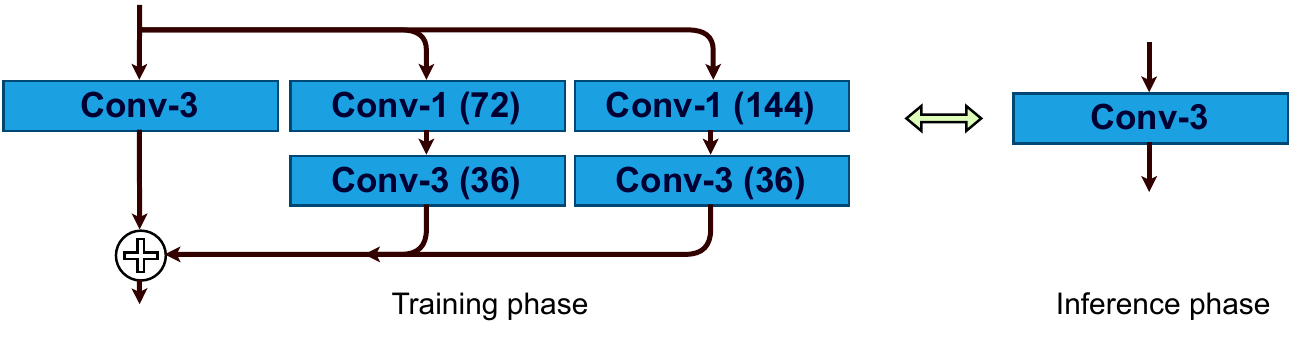}
	\caption{\textit{rainbow Team:} A schematic diagram of structural re-parameterization strategy.}
	\label{fig:reparam}
\end{figure}

\begin{figure}[htpb]
	\centering
	\includegraphics[width=0.4\textwidth]{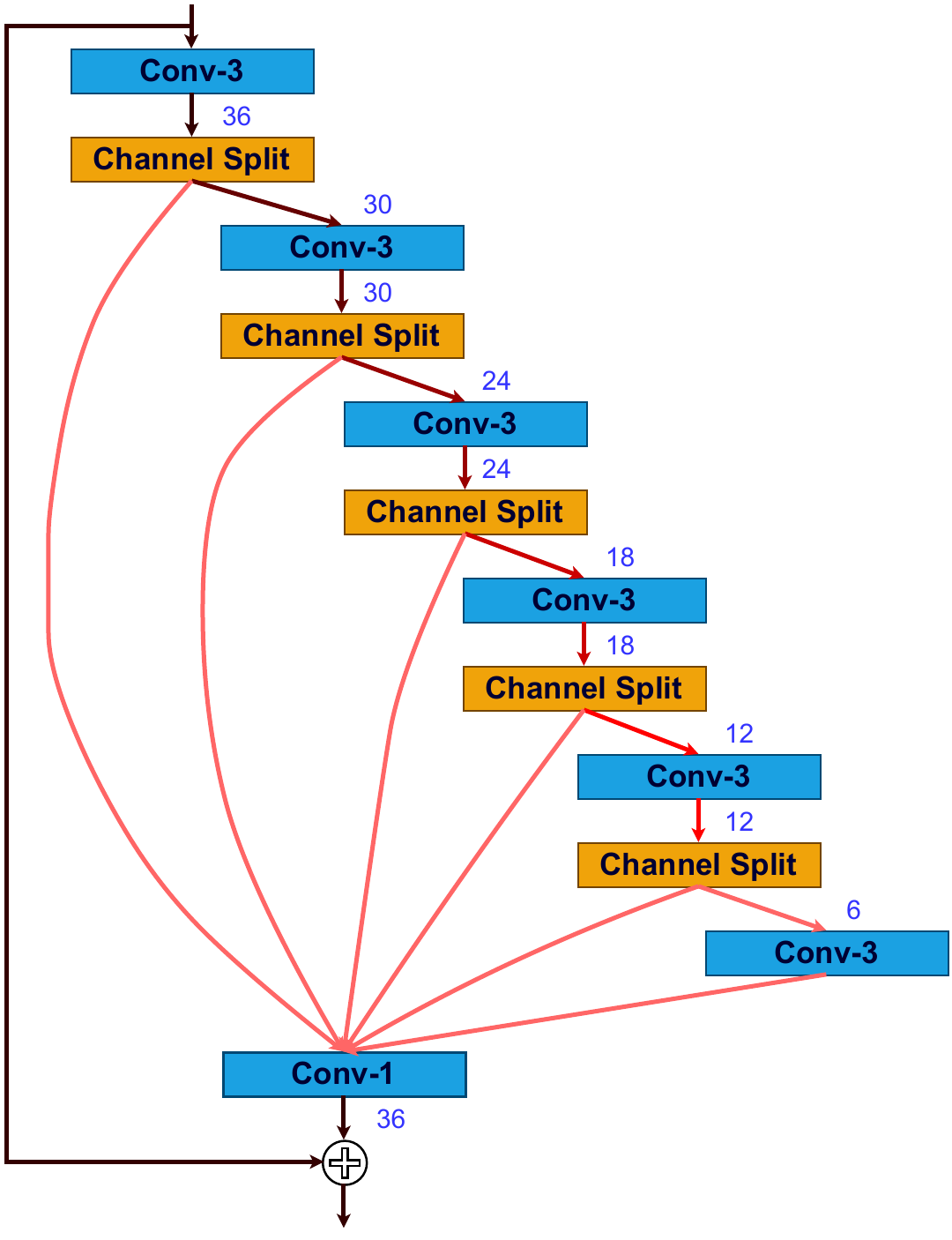}
	\caption{\textit{rainbow Team:} The architecture of the proposed improved information multi-distillation block (IMDB+). Here, $36$, $30$, $24$, $18$, $12$, and $6$ all represent the output channels of the convolution layer. ``Conv-3'' denotes the $3 \times 3$ convolutional layer. Each convolution followed by a SiLU activation function except for the last $1 \times 1$ convolution.}
	\label{fig:IMDB+}
\end{figure}

\begin{figure*}
    \centering
    \includegraphics[width=0.75\linewidth]{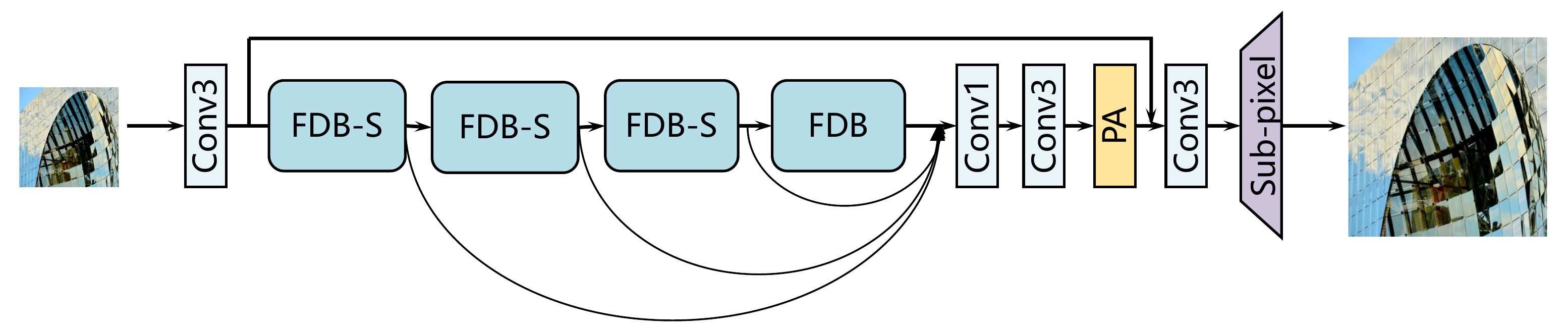}
    \caption{\textit{Super Team:} The overall network framework. }
    \label{fig:super_model}
\end{figure*}

\begin{figure}[t]
\centering
\subfloat[FDB]{\includegraphics[width=0.4\linewidth]{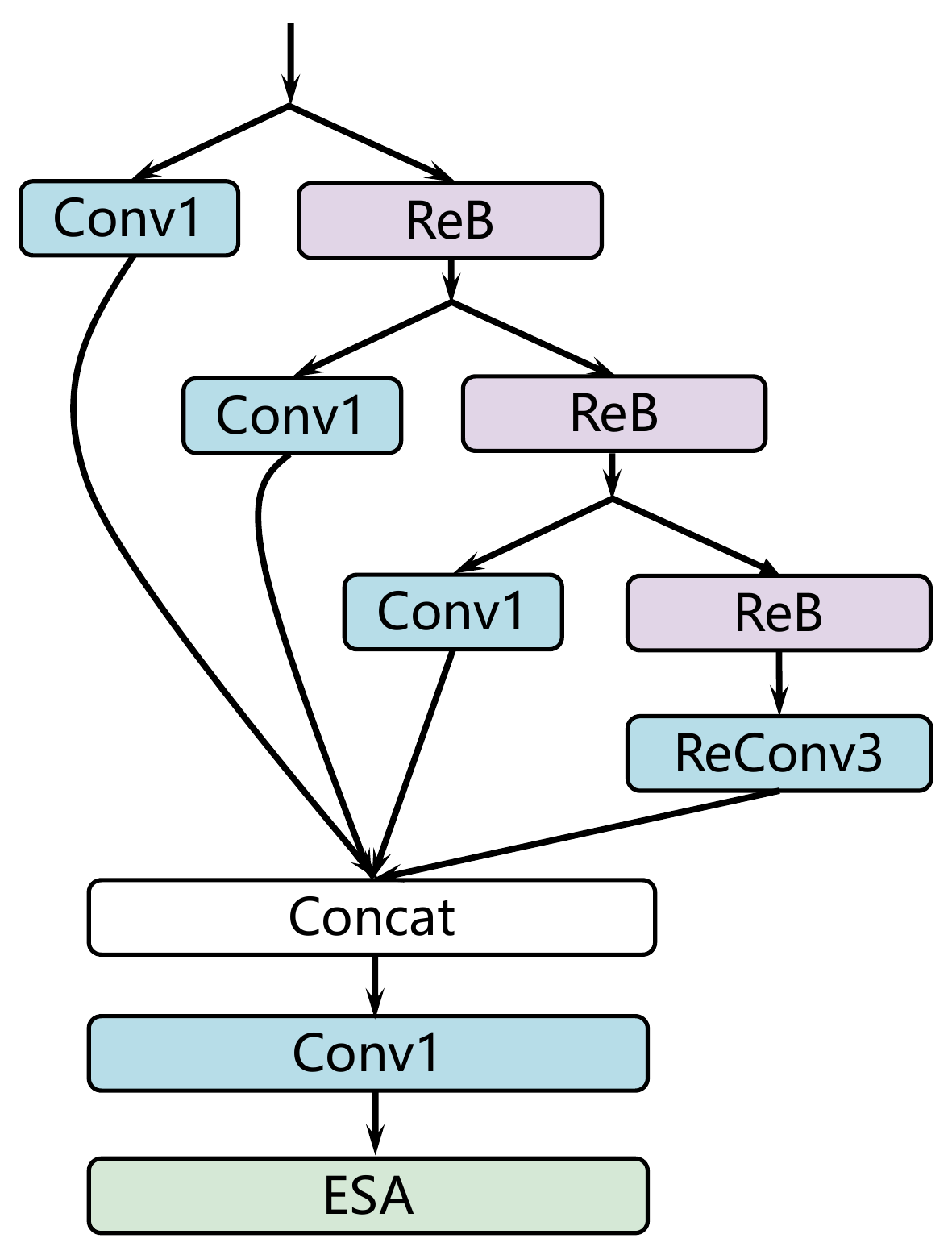}
\label{fig:super_FDB}}
\hfil
\subfloat[FDB-S]{\includegraphics[width=0.34\linewidth]{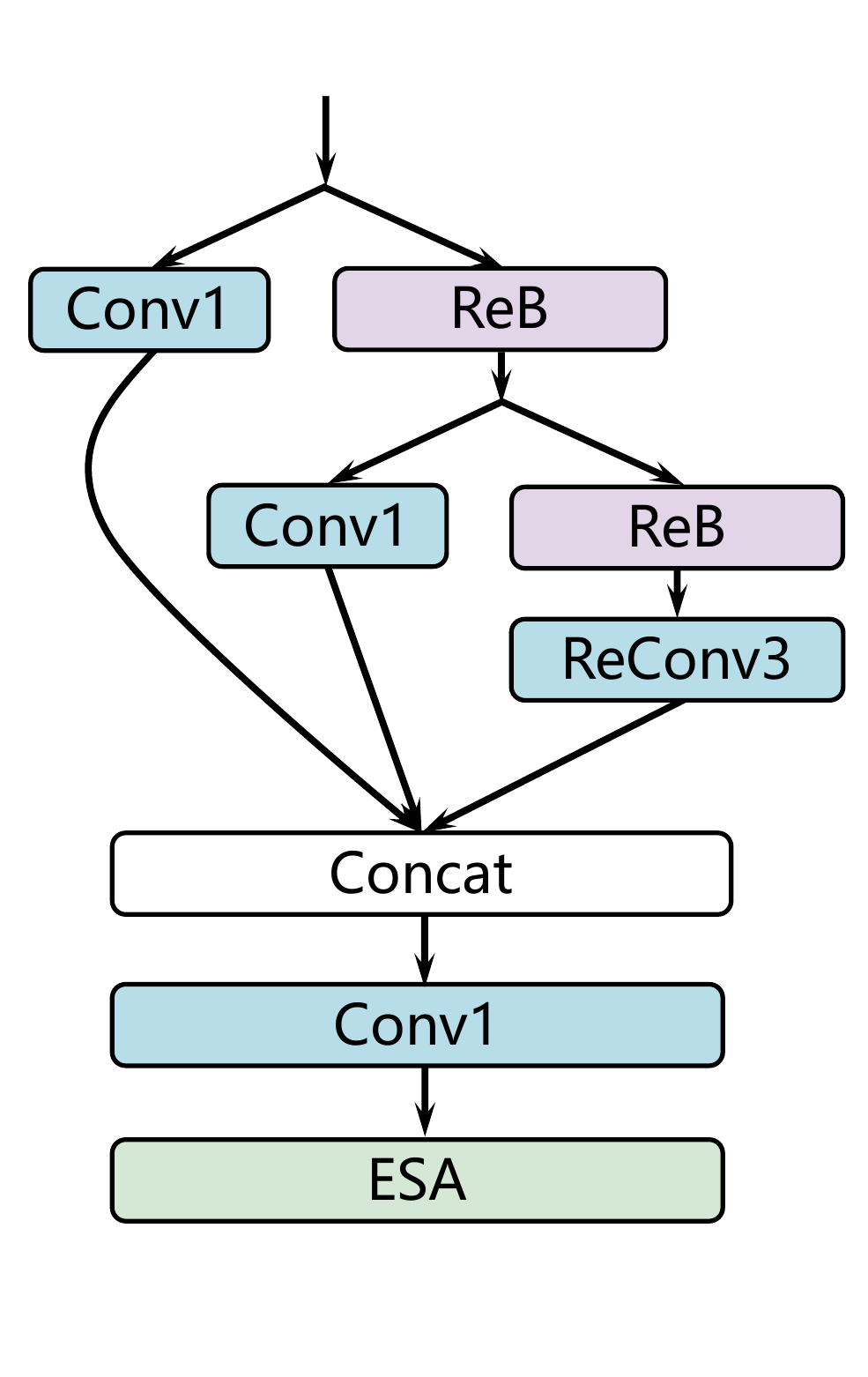}%
\label{fig:super_FDBS}}
\hfil
\subfloat[ReB]{\includegraphics[width=0.16\linewidth]{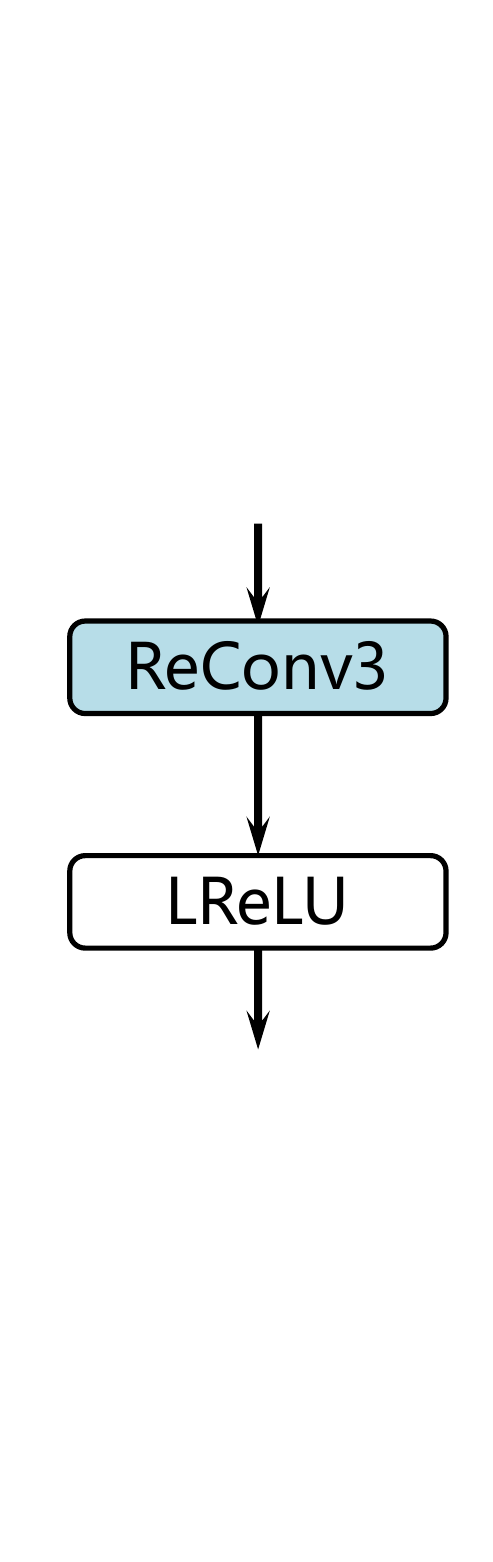}%
\label{fig:super_MB}}
\caption{\textit{Super Team:} (a) Feature Distillation Block (FDB). (b) Feature Distillation Block-Small (FDB-S). (c)  Re-parameterized Block (ReB).}
\label{fig:super_block}
\end{figure}

\subsection{rainbow}

The rainbow team proposed Improved Information Distillation Network for efficient SR shown in \cref{fig:IMDN+}.
This solution mainly concentrates on improving the effectiveness of the information multi-distillation block (IMDB)~\cite{IMDN}. Different from the original IMDB, as illustrated in \cref{fig:IMDB+}, the improved IMDB (IMDB+) uses $5$ channel split operations. The number of input channels is set to $36$. In order to improve the performance of IMDB+, structural re-parameterization methods are used~\cite{ding2021diverse,zhang2021edge} to replace ``Conv-3'' during the training phase as shown in \cref{fig:reparam}. Although re-parameterization can improve performance (about 0.03dB on DIV2K validation set), it will increase the training time. Different from the ECB proposed in ECBSR~\cite{zhang2021edge}, Conv1$ \times$1-Sobel and Conv1$\times$1-Laplacian are removed for efficient training. This is because Sobel and Laplacian filters are implemented by depth-wise convolution.

\subsection{Super}

The Super team proposed a solution mainly based on RFDN~\cite{RFDN}, where the channel splitting operation in IMDB~\cite{IMDN} is replaced by $1 \times 1$ convolution for feature distillation. The method differs from RFDN in three aspects: 1) Pixel Attention~\cite{PAN} is introduced in the network to effectively improve the feature representation capacity. 
2) Model re-parameterization technique is adopted to expand the capacity of the network during training, while keeping the computations during inference.
3) Further compression of the model is accomplished by reducing the size of the first three blocks.

\textbf{Framework.}
The team used a similar framework as RFDN~\cite{RFDN}, as shown in \cref{fig:super_model}. The Pixel Attention Feature Distillation Network (RePAFDN) consists of four parts: the feature extraction convolution, the stacked feature distillation blocks with different sizes (FDB-S and FDB), the feature fusion part and the reconstruction block. 

Given the input $x$, coarse features are first extracted as:
\begin{equation}
    F_0 = h(x),
\end{equation}
where $h$ denotes the feature extraction function, implemented by a $3 \times 3$ convolution, and $F_0$ is the extracted features. Next, three FDB-S and one FDB are stacked to gradually refine the extracted features, formulated as:
\begin{equation}
    F_k = H_k(F_{k-1}), k=1, \cdots, 4,
\end{equation}
where $H_k$ denotes the $k$-th feature distillation block, $F_{k-1}$ and $F_k$ represent the input feature and output feature of the $k$-th feature distillation block, respectively. All the intermediate features are fused by a $1\times 1$ convolution and a $3 \times 3$ convolution. The fused features are then fed to the pixel attention layer as:
\begin{equation}
    F_{fused} = H_{PA}(H_f(Concat(F1, \cdots, F_4)),
\end{equation}
where $Concat$ is the concatenation operation along the channel dimension, $H_f$ denots the $1\times 1$ convolution followed by a $3\times 3$ convolution, $H_{PA}$ is the pixel attention layer, and $F_{fused}$ is the fused features. Finally, the output is generated as:
\begin{equation}
    y = Up(F_{fused}+F_0),
\end{equation}
where $Up$ is the reconstruction function (\emph{i.e.} a $3\times 3$ convolution and a sub-pixel operation) and $y$ is the output SR image. 

\textbf{Feature Distillation Block.}
Two variants of Feature Distillation Block (FDB) are designed. The primitive FDB (\cref{fig:super_FDB}) is similar to RFDB except that the residual connection is removed and Re-parameterized Block (ReB) is used to replace the Shallow Residual Block (SRB). As shown in \cref{fig:super_MB}, ReB contains a re-parameterized $3\times 3$ convolution (ReConv3) and LReLU function. The details of the ReConv3 will be explained in the next paragraph. The whole structure of FDB can be described as:
\begin{equation}
    \begin{aligned}
        F_{distilled1}, F_{remain1} &= DL_1(F_{in}), RL_1(F_{in}), \\
        F_{distilled2}, F_{remain2} &= DL_2(F_{remain1}), RL_2(F_{remain1}), \\
        F_{distilled3}, F_{remain3} &= DL_3(F_{remain2}), RL_3(F_{remain2}), \\
        F_{distilled4} &= ReConv3(F_{remain3}),
    \end{aligned}
\end{equation} 
where $DL_i$ is the $i$-th $1 \times 1$ convolution and $RL_i$ is the $i$-th ReB. The distilled features $F_{distilled1}, \cdots, F_{distilled4}$ are concatenated and fed to a $1\times 1$ convlution and the ESA block~\cite{RFANet} for further enhancement. 
As for the more lightweight FDB (FDB-S), a feature distillation layer (\emph{i.e.} a $1\times 1$ convolution and a ReB) is removed as shown in \cref{fig:super_FDBS}.

\begin{figure}
    \centering
    \includegraphics[width=0.85\linewidth]{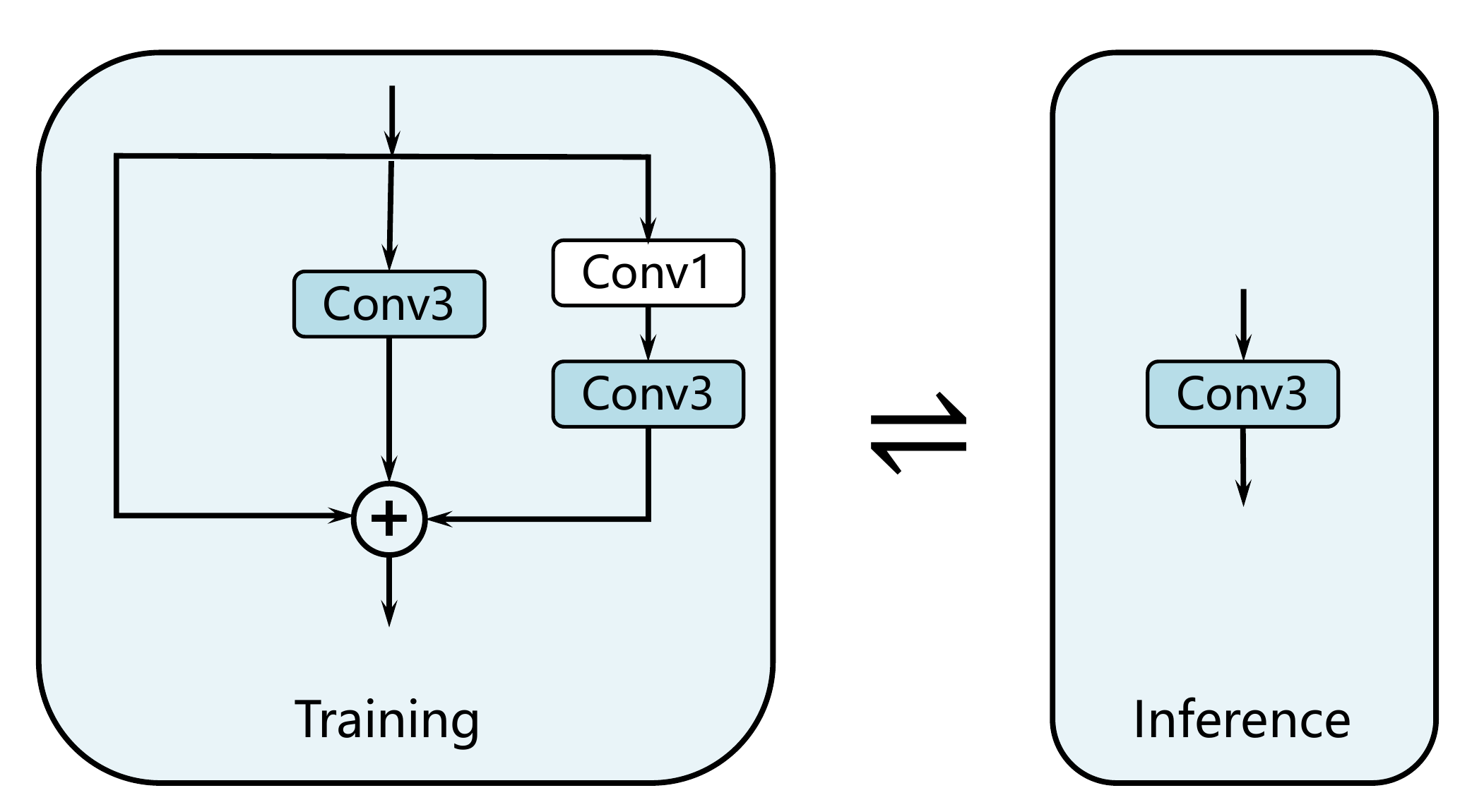}
    \caption{\textit{Super Team:} Re-parameterized $3\times 3$ Convolution (ReConv3). }
    \label{fig:super_Reconv3}
\end{figure}

\begin{figure}
    \centering
    \includegraphics[width=0.85\linewidth]{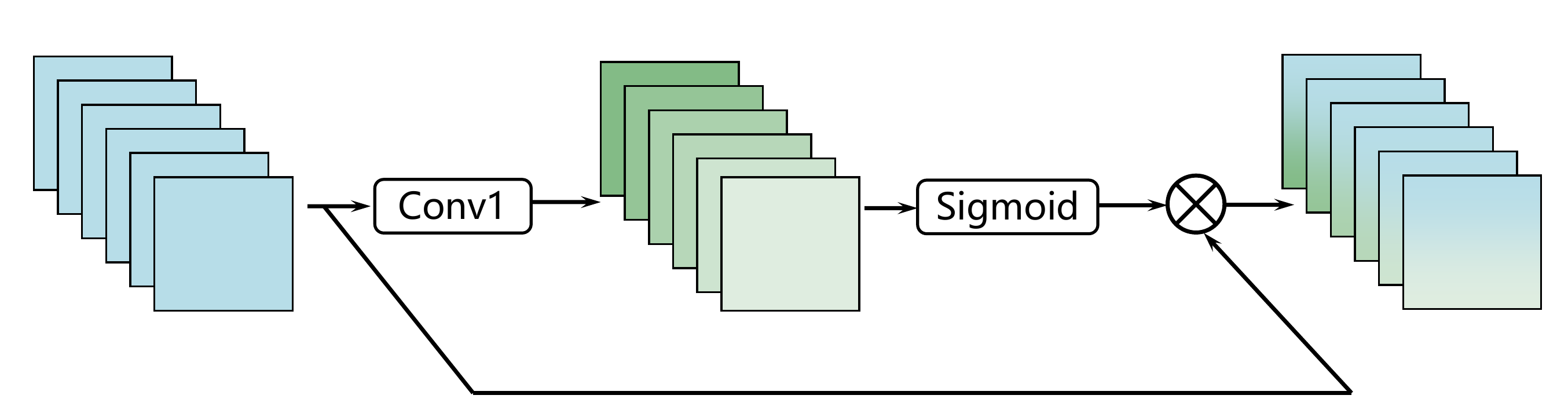}
    \caption{\textit{Super Team:} Pixel Attention (PA).}
    \label{fig:super_PA}
\end{figure}

\begin{figure*}[!ht]
    \centering
    \includegraphics[width=0.8\linewidth]{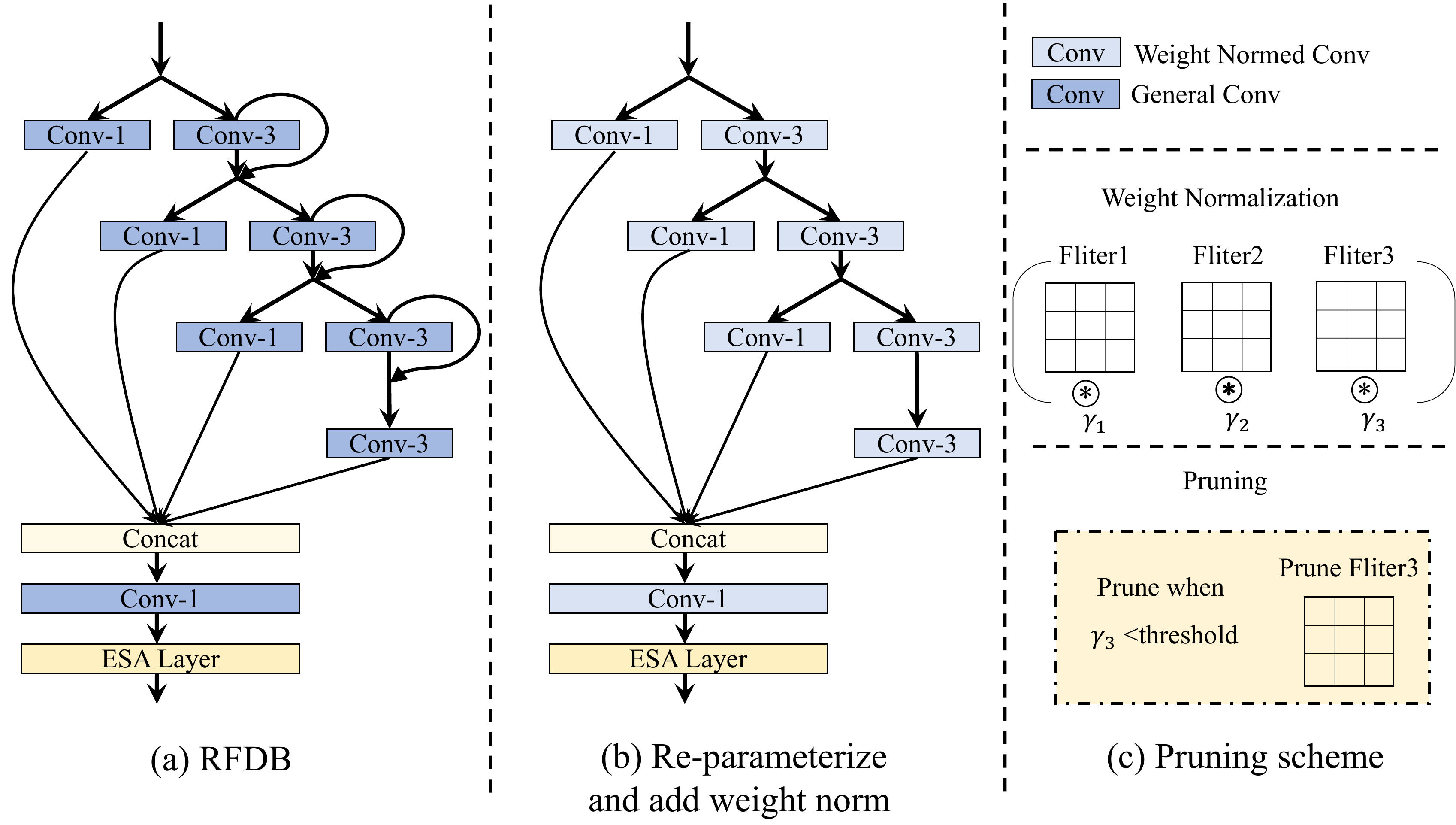}
    \caption{\textit{MegSR Team:} (a) The basic block of RFDN. (b) Reparameterization for identity connection and applying weight normalization on convolutional layers. (c) Pruning the unimportant weight.}
    \label{fig:ppl}
\end{figure*}

\textbf{Model Re-parameterization.}
Inspired by recent model re-parameterization works~\cite{bhardwaj2021collapsible, zhang2021edge}, similar techniques are adopted to improve model performance. Specifically, the SRB block inside RFDB is redesigned by introducing multi-branch convolution during training. As shown in \cref{fig:super_Reconv3}, there are two extra branches along with the original convolution of size $3\times3$, which consists of an identity shortcut and two cascaded convolution layers with size of $1\times1$ and $3\times3$ respectively. The outputs of the three branches are added before being fed into the activation layer, which can be formulated as:
\begin{equation}
\begin{aligned}
    F_{ReConv3}^{training} = ~&~ F_{in}+
    Conv_{3\times3}^1(F_{in})+\\
    ~&~Conv_{3\times3}^2(Conv_{1\times1}(F_{in})),\\
    F_{ReConv3}^{inference} = ~&~ Conv_{3\times3}^{rep}(F_{in}),
\end{aligned}
\end{equation}
where $F_{in}$ represents the input feature, and $Conv_{3\times3}^{rep}$ represents re-parameterized $3\times3$ convolution.
Since the operations of the three branches are completely linear, the re-parameterized architecture can be equally converted to a single convolution of size $3\times3$ for inference. In the experiments, the re-parameterization technique helps improve the PSNR of small models by 0.02dB.

\textbf{Pixel Attention.} 
Inspired by PAN~\cite{PAN}, pixel attention is used to more effectively generates features for the final reconstruction block. Specifically, a $1\times 1$ convolution followed by a Sigmoid function is responsible for generating a 3D attention coefficients map for all pixels of the feature map (shown in \cref{fig:super_PA}). The PA layer can be formulated as:
\begin{equation}
    F_{PA} = PA(F_{in})\cdot F_{in},
\end{equation}
Unlike PAN, PA is not introduced into FDBs since PA is not runtime friendly. For similar consideration, the PA is conducted in low-resolution space to save computations while PA in UPA of PAN is conducted in higher resolution. 

The channel number used in the model is $48$. For $DL_1, \cdots, DL_3$ in FDB, the number of output channels is $12$. For $DL_1$ and $DL_2$ in FDB-S, the number of output channels is $24$. 
DIV2K and Flickr2K are used as the training set. For the first training stage, patches of size $128\times128$ are cropped from the LR images as inputs. For the second training stage, patches of size $160 \times 160$ are cropped from the LR images as inputs. 

\subsection{MegSR}

The MegSR team proposed PFDNet, a light-weight network for efficient super-resolution. Previous works such as IMDN~\cite{IMDN} and RFDN~\cite{RFDN} introduce novel network blocks, which are variants of feature distillation block, and achieve favorable performance. Unlike these works, the team proposed to tackle this problem based on pruning strategies. Albeit the techniques of network pruning are widely used in high-level tasks, such as image classification and segmentation, its applications on low-level tasks are rare. A recent work ASSL~\cite{zhang2021aligned} propose a pruning scheme for residual network in the SR task, showing the network pruning technique is effective. Inspired by RFDN and ASSL, the team explored how to combine pruning and feature distillation network.

Specifically, the method contains two stages: \textit{training stage} and \textit{fine-tuning stage}. 
\textit{Training stage}: In this stage, the original architecture of RFDN is first trained to obtain a pretrained model. Then, the model is reparameterized to reduce the residual addition operators as many as possible. When pruning the features of a network, the indices of features retained in different layers may be different. Thus, it is not reasonable to add up the features with different indices. To solve this problem, except for the ESA ~\cite{RFANet} layers, weight normalization (WN) is applied to all convolutions. The learnable parameters $\gamma$ of WN indicate the importance of features. Finally, the new model is trained with $\ell_1$ loss to maintain the performance, while a regularized term is used to force the unimportant weights to converge to zero.  
\textit{Fine-tuning stage}: After the training stage, the weights in the model is pruned according to the values of parameters $\gamma$. Note that, the remaining convolutional layers where are WN is not applied are pruned according to $\gamma$ of the previous layers. The parameter $\gamma$ can be fused into network weight during inference. Thus, using WN does not increase the computational cost.  After pruning, the pruned model is fine-tuned with $\ell_1$ loss in the first $300, 000$ iterations and with $\ell_2$ in the last $100, 000$ iterations.

\textbf{Reparameterization.} Denote a feature as $X$, and a weight of convolution as $W$, this following equation holds:
\begin{equation}
    WX + X = (W+I)X,
\end{equation}
where $I$ is the identity matrix. As depicted in \cref{fig:ppl}(b), all the skip-connections are removed from RFDB without degradation.

\textbf{Weight Normalization.} Weight Normalization includes learnable parameters which can tell the importance of weights, as shown in \cref{fig:ppl} (c):
\begin{equation}
\hat{\mathbf{W}}_{i}=\frac{\mathbf{W}_{i}}{\left\|\mathbf{W}_{i}\right\|_{2}}, \mathbf{W}_{i}=\boldsymbol{\gamma}_{i} \hat{\mathbf{W}}_{i}, \text { for } i \in\{1,2, \cdots, N\}
\end{equation}
where $\mathbf{W} \in \mathbb{R}^{N \times C \times H \times W}$ $W$ represents the 4 dimensional convolutional kernel, and $\boldsymbol{\gamma} \in \mathbb{R}^{N}$ stands for the
1 dimensional trainable scale parameters in WN.

\textbf{Loss Function.} During training, the model is trained with $\ell_1$ loss and the following penalty term:
\begin{equation}
\mathcal{L}_{S I}=\alpha \sum_{l=1}^{L} \sum_{i \in S^{(l)}}, \gamma_{i}^{2}
\end{equation}
where $\alpha$ is the scalar loss weight, $\gamma_{i}$ denotes the i-th element of $\gamma$, and $S^{(l)}$ represents the unimportant filter index set in the l-th layer.

\subsection{VMCL\_Taobao}

\begin{figure}[!ht]
	\centering
	\includegraphics[width=1.0\linewidth]{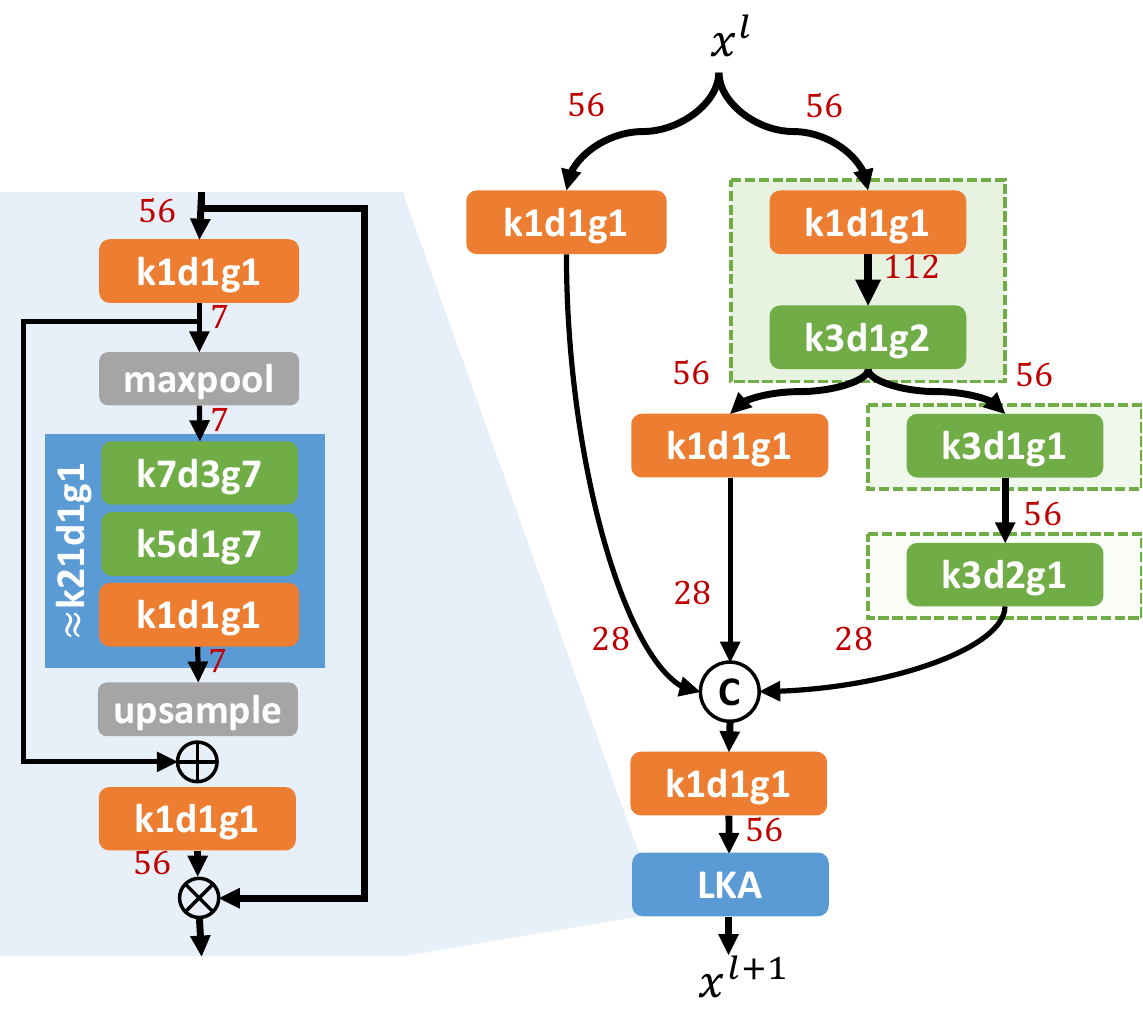}
	\caption{\textit{VMCL\_Taobao Team:} The architecture of the proposed multi-scale information distillation block (MSDB).}
	\label{fig:vmcl_taobal_framework}
\end{figure}

\begin{figure*}[!httbp]
	\centering
	\includegraphics[scale=0.25]{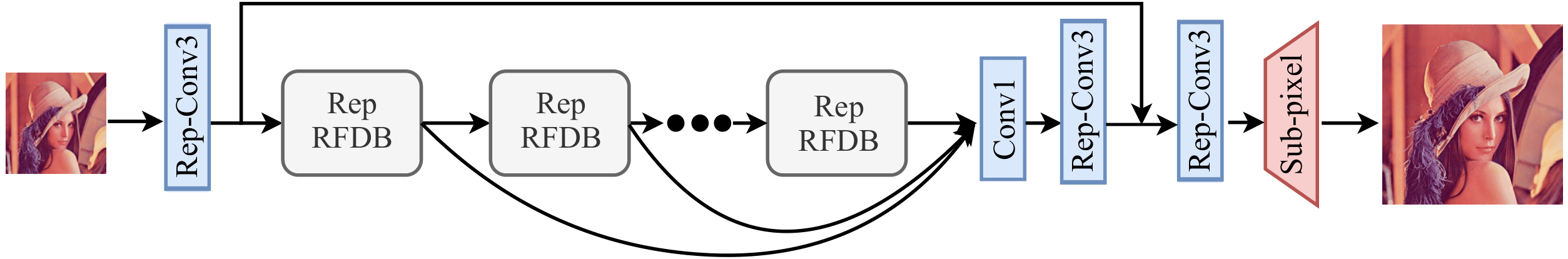}\\
	\caption{\textit{Bilibili AI Team}: The architecture of Re-parameterized Residual Feature Distillation Network (Rep-RFDN). }
	\label{fig:bilibili_fig1}
\end{figure*}
\begin{figure}[!httb]
	\centering
	\includegraphics[width=1.0\linewidth]{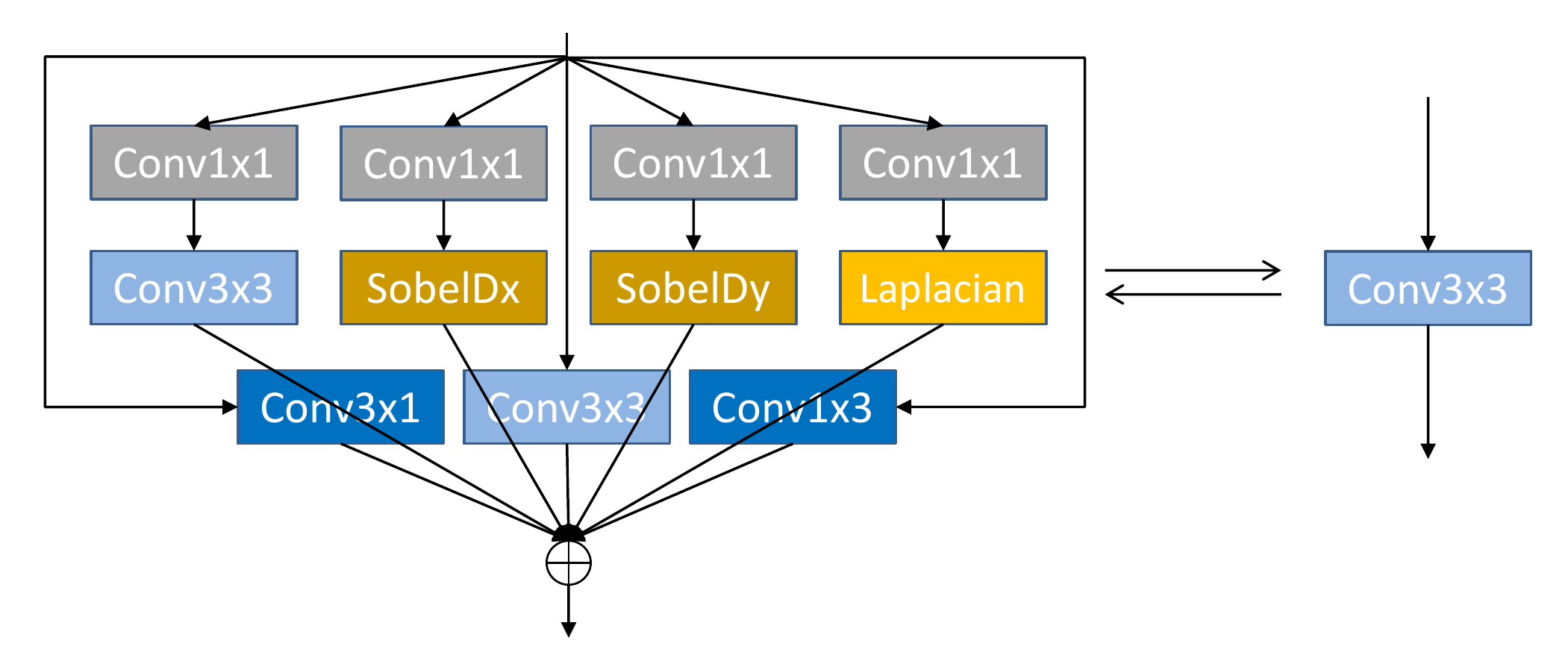}\\
	\caption{\textit{Bilibili AI Team}: The architecture of the proposed RepBlock (RB). }
	\label{fig:bilibili_fig2}
\end{figure}

To improve the representation capacity of information flows in IMDN, the VMCL\_Taobao team proposed a Multi-Scale information Distillation Network (MSDN) for efficient super-resolution, which stacks a group of multi-scale information distillation blocks (MSDB). Particularly, inspired by RFDN, a $1\times1$ convolution is used for information distillation and a $3\times3$ convolution is used for feature refinement to alleviate the limitation of channel splitting operation in IMDB. 
As shown in \cref{fig:vmcl_taobal_framework}, in $l\text{-th}$ MSDB, a multi-scale feature refinement module (marked with green dotted boxes) is used to replace the $3\times3$ convolution of RFDB.
In \cref{fig:vmcl_taobal_framework}, an upsampling refinement module with the scale factor of 2 is designed. A $1\times1$ convolution is used for channel expansion and a $3\times3$ convolution with two groups is used for feature refinement, which has $\sqrt{s}h\times\sqrt{s}h$ receptive field to capture a larger region of neighbors and acts equivalently on an upsampled feature.
Then, a single $3\times3$ convolution is used for identical refinement as done in RFDB. 
Last, a dilated $3\times3$ convolution with dilation rate of 2 is employed for downsampling refinement, which has $(h/\sqrt{s})\times(h/\sqrt{s})$ receptive field and acts equivalently on a downsampled feature. 
By applying the multi-scale feature refinement, multi-scale information of the input features can be captured with fewer computations.
Moreover, the Large Kernel Attention (LKA)~\cite{guo2022visual} is introduced to enhance the features by capturing a larger receptive field.

\subsection{Bilibili AI}

The Bilibili AI team used Re-parameterized Residual Feature Distillation Network (Rep-RFDN) as shown in \cref{fig:bilibili_fig1}. Different from the original RFDN~\cite{RFDN}, all $3\times3$ convolutional layers except those in the ESA block \cite{RFDN}) are replaced by RepBlocks (RB) in the training stage. During inference stage, the RepBlocks are converted into single $3\times3$ convolutional layers.
Inspried by ECB~\cite{zhang2021edge} and ACB~\cite{ding2019acnet}, $3\times1$ Conv and $1\times3$ Conv sub-branches are added into the original ECB (\cref{fig:bilibili_fig2}).
The number of feature channels is set to 40, while in the original RFDN50 version it is set to 50.

\subsection{NKU-ESR}
Generally, the team proposed an edge-enhanced feature distillation network, named EFDN, to preserve the high-frequency information under the synergy of network and loss devising. 
In detail, an edge-enhanced convolution block is built by revisiting the existing reparameterization methods. The backbone of the EFDN is searched by neural architecture search (NAS) to improve the basis performance. Meanwhile, an edge-enhanced gradient loss is proposed to calibrate the reparameterized block training.

\begin{figure}
  \centering
  \begin{subfigure}{1.0\linewidth}
  \includegraphics[width=1.0\linewidth]{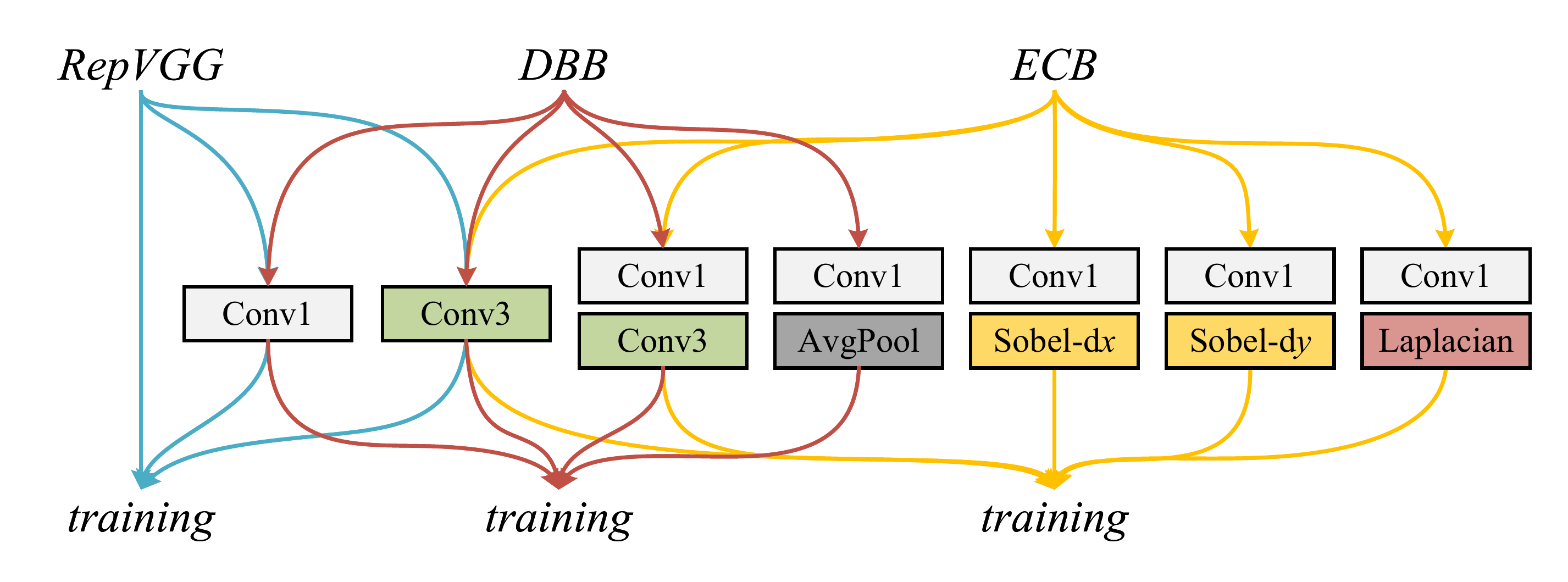}
    \caption{Revisiting re-parameterizable typology.}
    \label{fig:NKU_ESR_Block-a}
  \end{subfigure}
  \\
  \begin{subfigure}{1.0\linewidth}
  \includegraphics[width=1.0\linewidth]{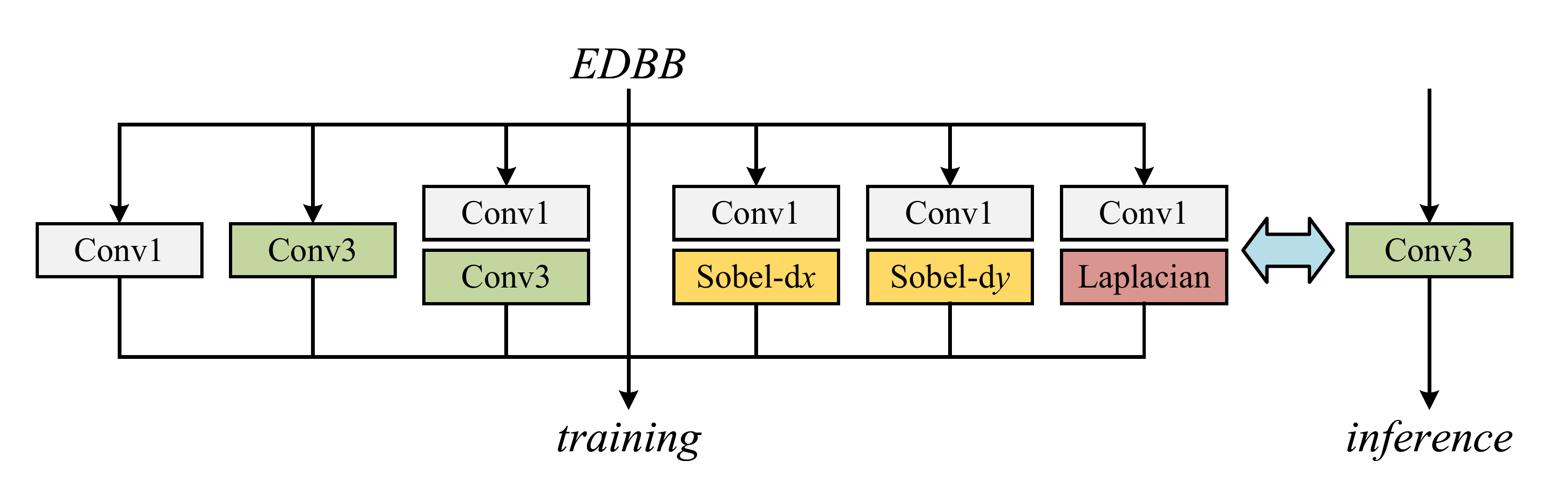}
    \caption{Proposed edge-enhanced diverse branch block (EDBB).}
    \label{fig:NKU_ESR_Block-b}
  \end{subfigure}
  \caption{\textit{NKU-ESR Team:} Illustration of re-parameterization method.}
  \label{fig:NKU_ESR_short}
\end{figure}

\begin{figure*}[t]
  \centering
   \includegraphics[width=1.0\linewidth]{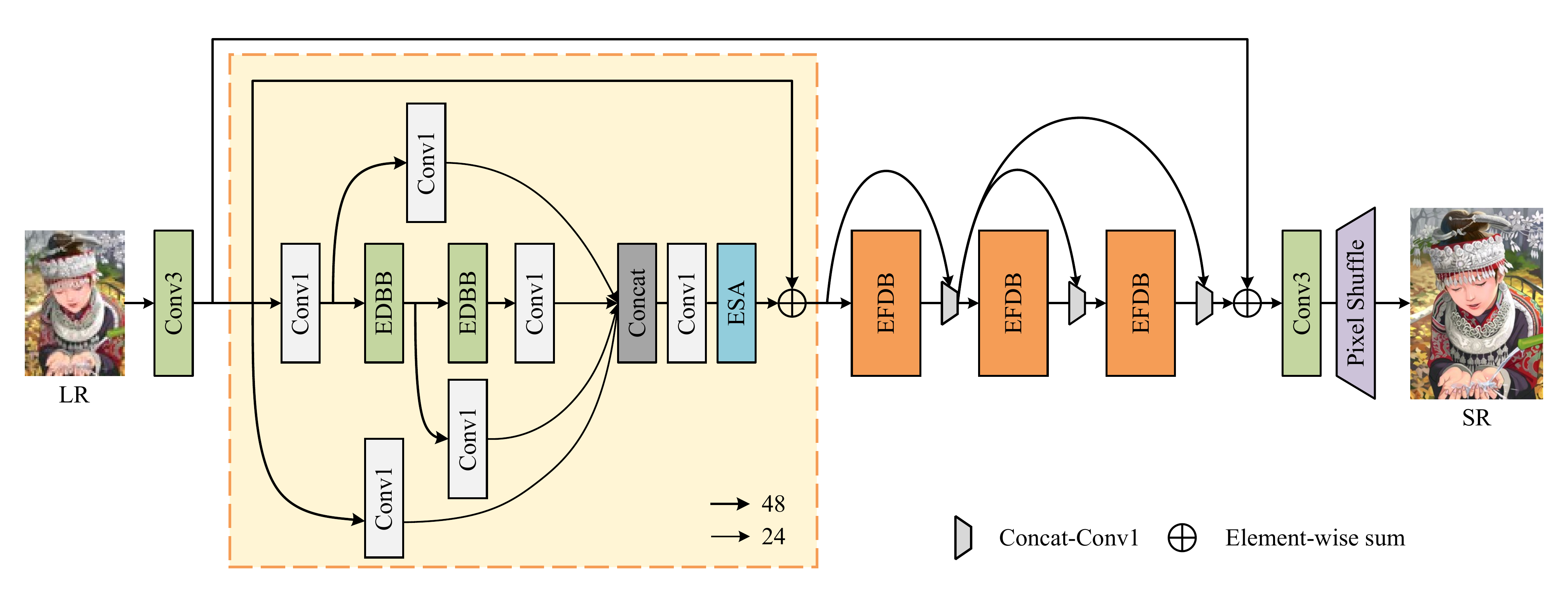}

   \caption{\textit{NKU-ESR Team:} Network architecture of the proposed EFDN.}
   \label{fig:NKU_ESR_Network}
\end{figure*}

\textbf{Edge-enhanced diverse branch block.}
As shown in \cref{fig:NKU_ESR_Block-a}, the detail of RepVGG Block, DBB, and ECB is presented. A total of eight different structures have been designed to improve the feature extraction ability of the vanilla convolution in different scenarios. Although the performance may be higher with more re-parameterizable branches, the expensive training cost is unaffordable for straightly integrating these paths. Meanwhile, another problem is that edge and structure information may be attenuated during the merging of parallel branches. 

To address the above concerns, a more delicate and effective reparameterization block is built, namely Edge-enhanced Diverse Branch Block (EDBB), which can extract and preserve high-level structural information for the low-level task. As illustrated in \cref{fig:NKU_ESR_Block-b}, the EDBB consists of seven  branches of single convolutions and sequential convolutions.

\textbf{Network architecture.}
Following IMDN\cite{IMDN} and RFDN\cite{RFDN}, an EFDN is devised to reconstruct high-quality SR images with sharp edges and clear structure under restricted resources. As illustrated in \cref{fig:NKU_ESR_Network}, the EFDN consists of an shallow feature extraction module, multiple edge-enhanced feature distillation blocks (EFDBs), and upscaling module. Specifically, a single vanilla convolution is leveraged to generate the initial feature maps. 

This coarse feature is then sent to stacked EFDBs for further information refining. In detail, the shallow residual block in~\cite{RFDN} is replaced by the proposed EDBB to construct the EFDB. Different from IMDN and RFDN utilizing global distillation connections to process input features progressively, neural architecture search (NAS)~\cite{DLSR} is adopted to decide the feature connection paths. The searched structure is shown in the orange dashed box. 
Finally, the SR images are generated by upscaling module.

\textbf{Edge-enhanced gradient-variance loss.}
In previous work\cite{lim2017enhanced}, $\mathcal{L}_{1}$ and $\mathcal{L}_{2}$ loss have been in common usage to obtain higher evaluation indicators. The network trained with these loss functions often leads to the loss of structural information. Although the edge-oriented components are added into the EDBB, it is hard to ensure their effectiveness during the complex training procedure of seven parallel branches.
Inspired by the gradient variance (GV) loss~\cite{GV}, an edge-enhanced gradient-variance (EG) loss is proposed, which utilizes the filters of the EDBB to monitor the optimization of the model. In detail, the HR image $I^{HR}$ and SR image $I^{SR}$ are transfered to gray-scale images $G^{HR}$ and $G^{SR}$. The Sobel and Laplacian filters are leveraged to compute the gradient maps and then unfold gradient maps into $\frac{HW}{n^2}\times n^2$ patches $G_{x}$, $G_{y}$, $G_{l}$. The $i$-th variance maps can be formulated as:
\begin{equation}
  v_i = \frac{\sum^{n^2}_{j=1}(G_{i,j}-\bar{G}_i)}{n^2-1}
  \label{var}
\end{equation}
where $\bar{G}_i$ is the mean value of the $i$-th patch. Thus, the variance metrics $v_{x}$,$v_{y}$,$v_{l}$ of HR and SR images can be calculated, respectively. Referring to GV-loss, the gradient variance loss of different filter can be obtained by:
\begin{equation} 
  \begin{aligned}
  \mathcal{L}_{x} &= \mathds{E}_{I^{SR}}\|v^{HR}_{x}-v^{SR}_{x}\|_2\\
  \mathcal{L}_{y} &= \mathds{E}_{I^{SR}}\|v^{HR}_{y}-v^{SR}_{y}\|_2\\
  \mathcal{L}_{l} &= \mathds{E}_{I^{SR}}\|v^{HR}_{l}-v^{SR}_{l}\|_2
\label{GV}
\end{aligned}
\end{equation}

Besides, $\mathcal{L}_{1}$ is added to accelerate convergence and improve the restoration performance. In order to better optimize the edge-oriented branches of EDBBs and preserve sharp edges for visual effects, coefficients $\lambda_x$, $\lambda_y$, and $\lambda_l$ are traded off, which are related to the scaled parameters of corresponding branches. The sum of the loss function can be expressed by:
\begin{equation}
   \mathcal{L} = \mathcal{L}_{1} + \lambda_x\mathcal{L}_{x} + \lambda_y\mathcal{L}_{y} + \lambda_l\mathcal{L}_{l} 
  \label{EG}
\end{equation}

\subsection{NJUST\_RESTORARION}

\begin{figure}[!t]
\centering
\includegraphics[scale=0.3]{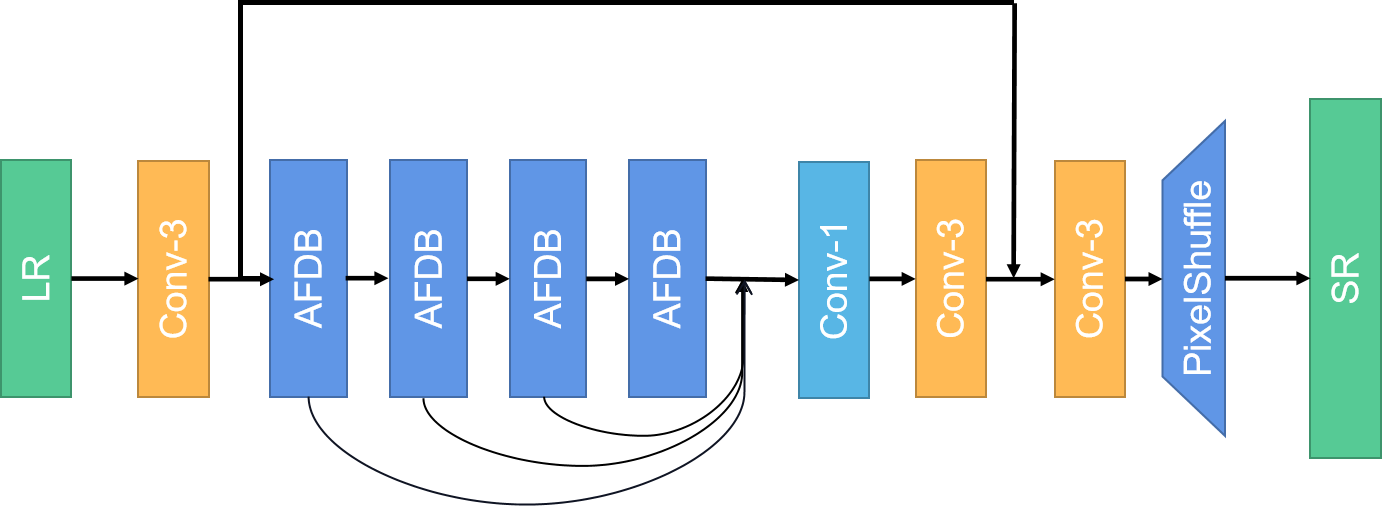}
\caption{\textit{NJUST\_RESTORARION Team:} The overall architecture of the AFDN.}
\label{fig:NJUST_RESTORARION_AFDN}
\end{figure}

\begin{figure}[!t]
	\centering
	\subfloat[AFDB]{\includegraphics[scale=0.3]{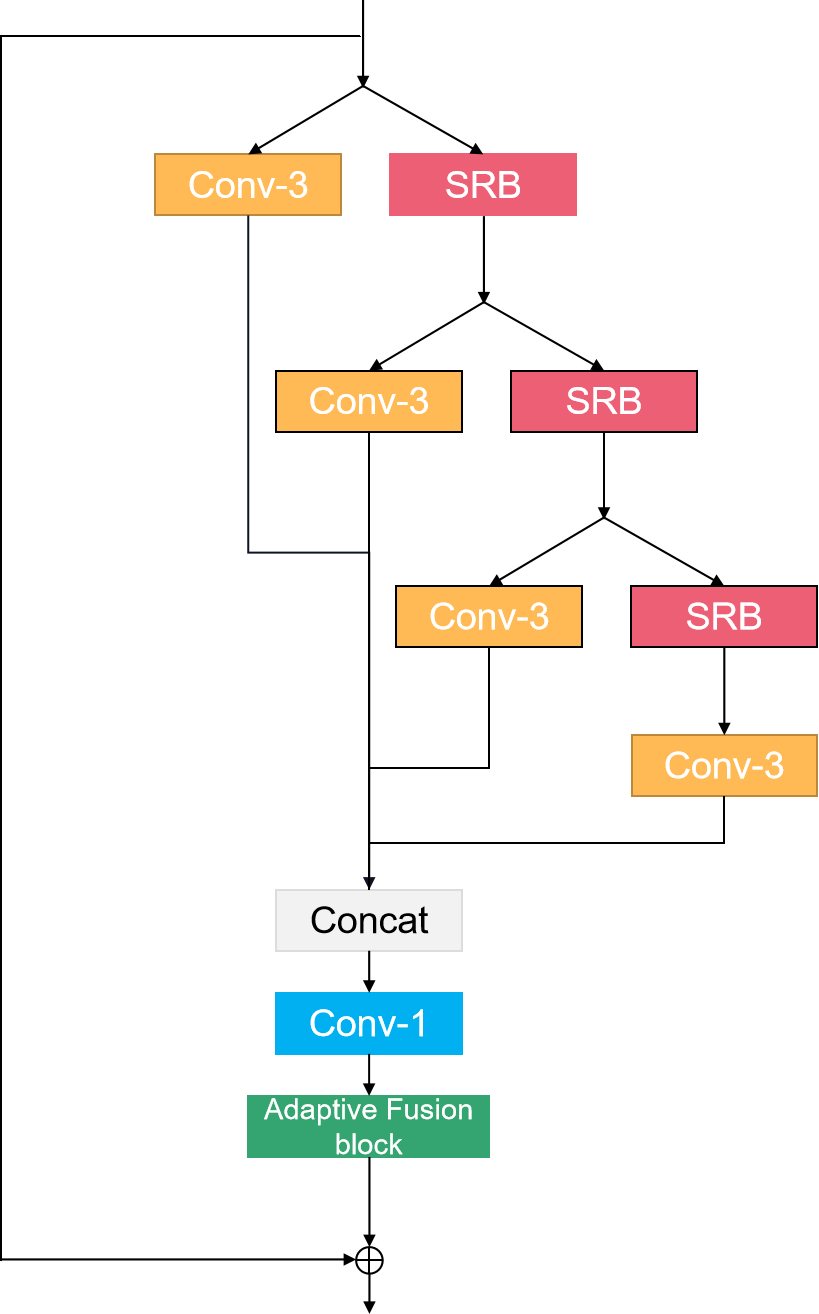}%
		\label{fig_first_case}}
	\hfil
	\subfloat[Adaptive Fusion Block]{\includegraphics[scale=0.3]{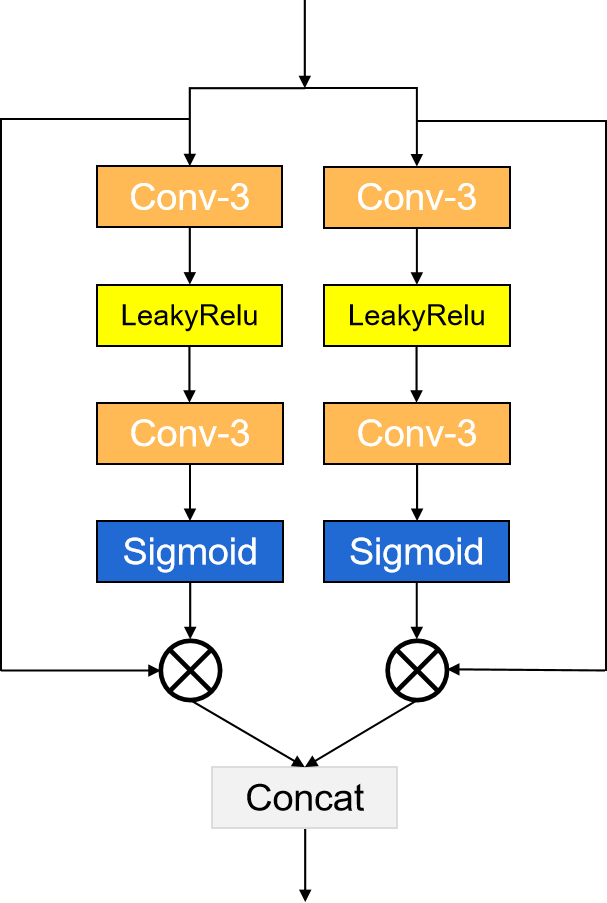}%
		\label{fig_second_case}}
	\caption{\textit{NJUST\_RESTORARION Team:} Adaptive Fusion Distillation Block}
	\label{fig:NJUST_RESTORARION}
\end{figure}

The NJUST\_RESTORARION team proposed Adaptive Feature Distillation Network(AFDN) for lightweight image SR. The proposed AFDN shown in \cref{fig:NJUST_RESTORARION_AFDN} is modified from RFDN~\cite{RFDN} with minor improvements. AFDN uses 4 AFDBs as the building blocks, and the overall framework follows the pipeline of RFDN.

As illustrated in \cref{fig:NJUST_RESTORARION_AFDN}, AFDN uses Adaptive Fusion Block (AFB) which is more efficient to fuse features. AFB splits the feature in half. Each branch uses ``Conv\_3-LeakyRelu-Conv\_3'' to learn the adaptive attention matrix. Then AFB multiplies the feature with the attention matrix. Finally, it concatenates the features of two branches.\\

\subsection{TOVBU}

\begin{figure}[!tb]
    \centering
    \includegraphics[width=1.0\linewidth]{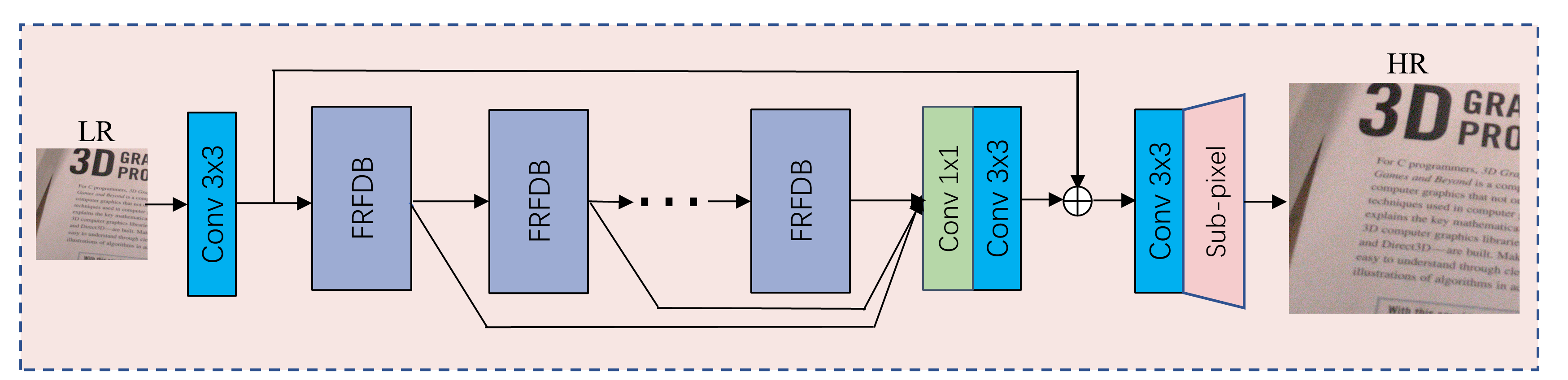}
    \caption{\textit{TOVBU Team:} Overall framework of of faster feature distillation network (FasterRFDN).}
    \label{fig-faster-rdfn}
\end{figure}

\begin{figure}[!tb]
    \centering
    \includegraphics[width=1.0\linewidth]{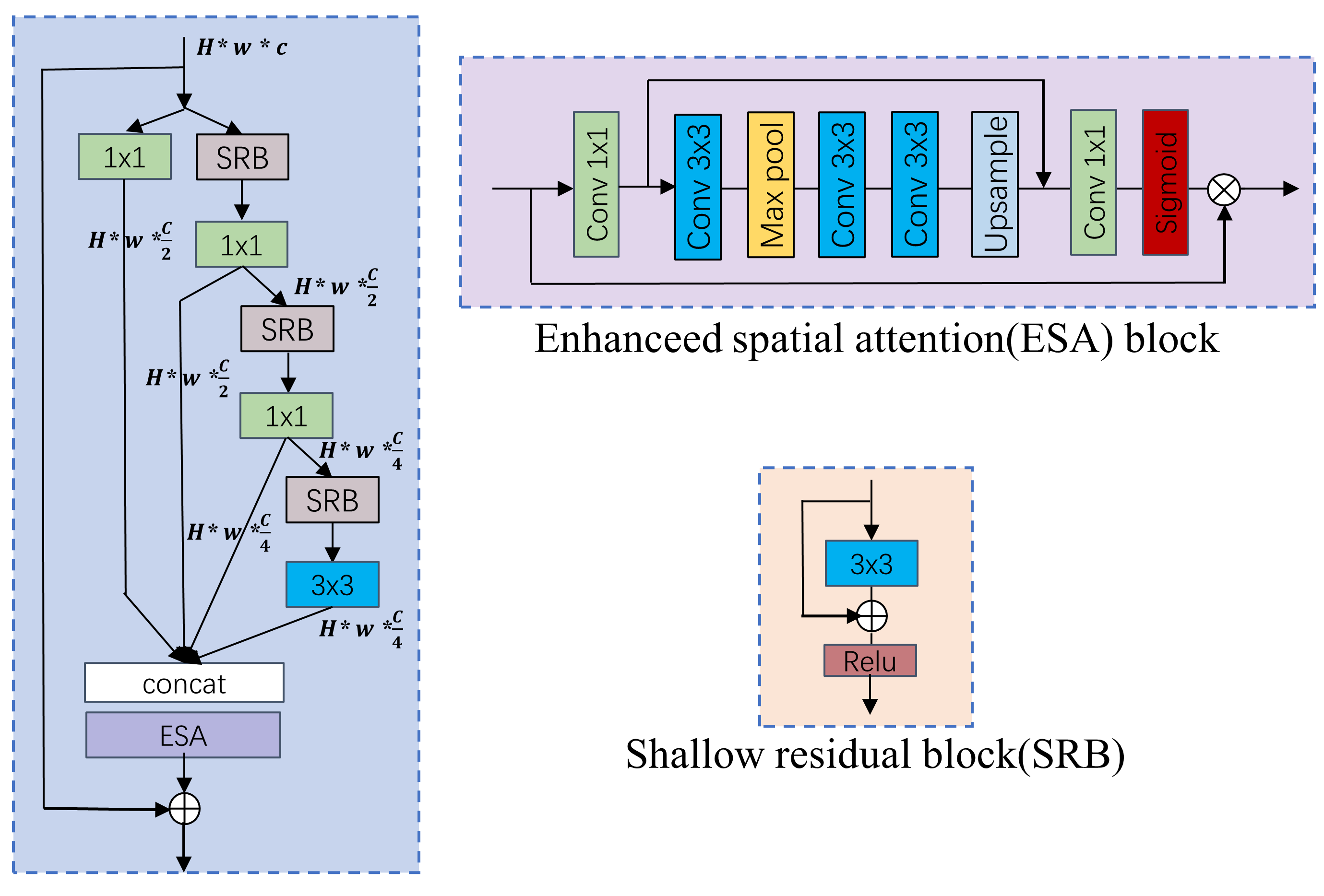}
    \caption{\textit{TOVBU Team:} Faster feature distillation block (FRFDB).}
    \label{fig-faster-rdfb}
\end{figure}

\textbf{Method details.}
On the basis of Residual Feature Distillation Network, the team proposed a novel efficient Faster Residual Feature Distillation Network (FasterRFDN) for single image super resolution. The overall framework of the proposed method is shown in \cref{fig-faster-rdfn} and \cref{fig-faster-rdfb}. The overall framework contains 4 faster residual feature distillation blocks (FRFDB). First, to further reduce the parameters and computational complexity of the FRFDB module, the number of channels of layered distillation is effectively compressed. The number of channels in each layer from top to bottom is 64, 32, 16, 16, respectively. These distillation features are extracted by three $1\times1$ and one $3\times3$ convolutional filters. Then, these features are fed to enhanced spatial attention (ESA) by concatenation along the channel dimension. Furthermore, in order to enhance the model's representation power, the number of channels of the model is increased to 64. 


\textbf{Training strategy.}
The training procedure can be divided into three stages.
\begin{enumerate}
    \item Pretraining on DIV2K and Flickr2K (DF2K). HR patches of size $256 \times 256$ are randomly cropped from HR images, and the mini-batch size is set to 64. The  model is trained by minimizing L1 loss function with Adam optimizer. The initial learning rate is set to $5\times 10^{-4}$ and halved at every 200k iterations. The total number of iterations is 1,600k.

    \item Finetuning on DF2K. HR patch size is $512 \times 512$, and the mini-batch size are set to 64, respectively. The model is fine-tuned by minimizing PSNR loss function. The initial learning rate is set to $5 \times 10 ^{-5}$ and halved at every 80k iterations. The total number of iterations is 480k.

    \item Fine-tuning on DF2K again. HR patch size and the mini-batch size are set to $640 \times 640$ and 16, respectively. The model is fine-tuned by minimizing L2 loss function. The initial learning rate is set to $1 \times 10 ^{-5}$ and cosine learning rate is used.
\end{enumerate}

\subsection{Alpan}

\begin{figure}
    \centering
    \includegraphics[width=0.9\linewidth]{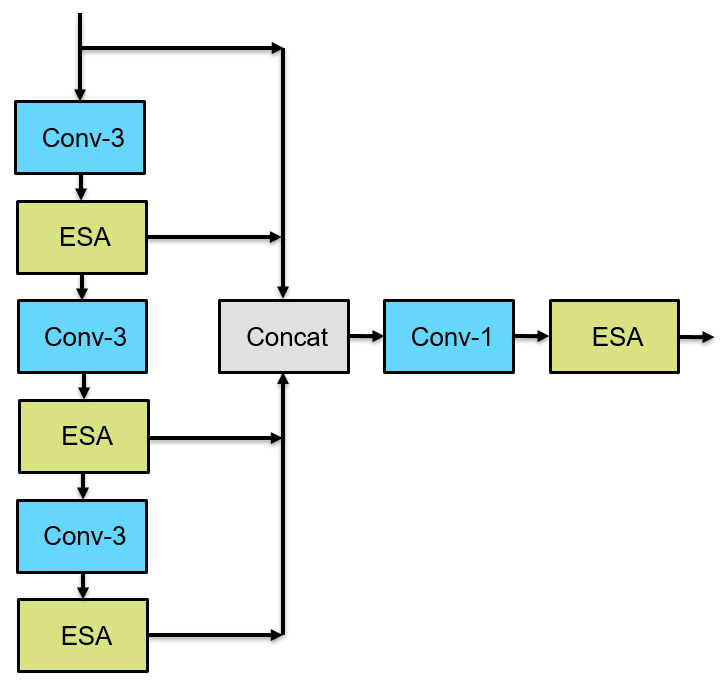}
    \caption{\textit{Alpan Team:} The building block for SR\_model consists of three $3\times3$ convolutions and three ESA blocks with $16$ channels (one ESA block after one convolution) followed by concatenation of input and $3$ outputs of each ESA block. Then $1\times1$ convolution and ESA block - exactly the same as in RFDB.}
    \label{fig:Alpan_block}
\end{figure}

The Alpan team proposed the method based on RFDN~\cite{RFDN} according to the following steps: 1) Rethinking of RFDB\cite{RFDN}. 2)Efficiency and PSNR trade-off for ESA\cite{RFDN} block and convolution. 3) Fine-tuning width and depth.

The team's first observation is that ESA~\cite{RFDN} block was efficient and significantly improves the results. Thus, the team placed ESA block after each $3 \times 3$ convolution in RFDB~\cite{RFDN}. All distillation convolutions from RFDB~\cite{RFDN} (three $1\times1$ convolutions and one $3\times3$ convolution) are removed and the number of RFDB\cite{RFDN} blocks is reduced from $4$ to $3$ to keep the same inference time. All these changes have the following effect:
1) PSNR goes up from $29.04$ to $29.05$ on DIV2K validation set.
2) The number of parameters is reduced from $0.433$M to $0.366$M.
3) FLOPS is reduced from $1.69$G to $1.256$G.

The team's next observation was that in the modified RFDB $75$\% parameters and more than $90$\% FLOPS belonged to convolutions outside of ESA~\cite{RFDN} blocks. So the team decided to re-balance the number of channels in ESA block. Specifically, the overall number of channels in the model is reduced from $50$ to $44$ but the number of channels in ESA blocks is increased from $12$ to $16$. All these changes have the following effect:
1) PSNR is almost the same on DIV2K validation set.
2) The number of parameters is reduced from $0.366$M to $0.356$M.
3) FLOPS is reduced from $1.256$G to $1.034$G.

In most of the team's experiments deeper models are better than wider models with the same efficiency. So the team decided to reduce the overall number of channels from $44$ to $32$ while keeping $16$ channels in ESA blocks and to increase the number of modified RFDB blocks from $3$ to $4$. This leads to significant reduction in FLOPS and small reduction in parameters in the final model:
1) Number of parameters: $0.326$M.
2) FLOPS: $0.767$G

The final model SR\_model consists of $4$ modified RFDB blocks with $32$ channels. All the other unmentioned parts of the model are the same as in RFDN. The modified RFDB block is shown in \cref{fig:Alpan_block}.

\subsection{xilinxSR}

\begin{figure}[t]
\begin{center}
   \includegraphics[width=1.0\linewidth]{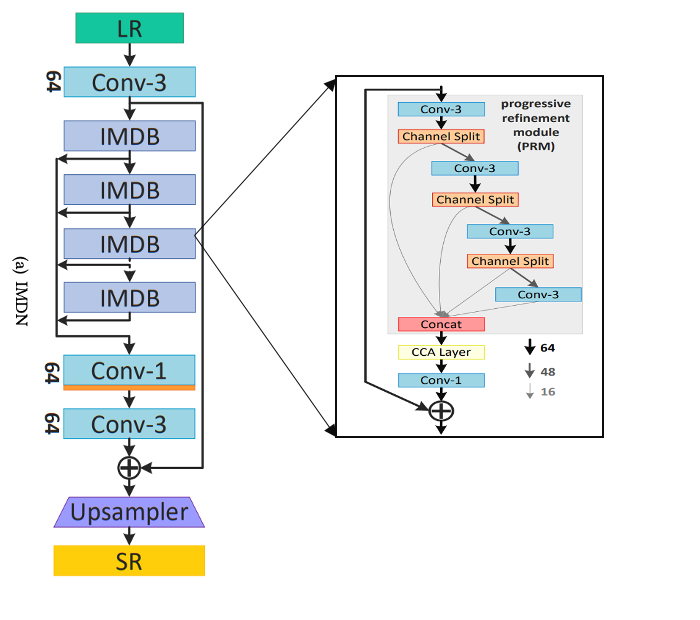}
\end{center}
    \caption{\textit{xilinxSR Team:} An overview of the basic IMDN architecture.}
\label{fig:xilinxSR_imdn}
\end{figure}
\begin{figure}[t]
\begin{center}
   \includegraphics[width=1.0\linewidth]{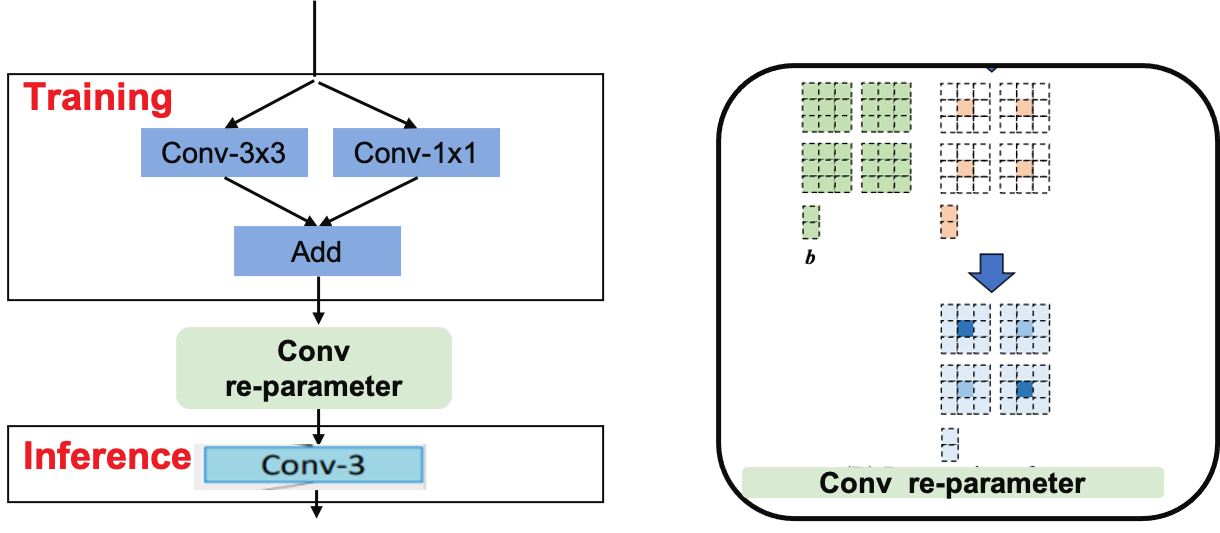}
\end{center}
    \caption{\textit{xilinxSR Team:} Structural re-parameterization of a collapsible block.}
\label{fig:xilinxSR_rep}
\end{figure}

\begin{figure*}[!t]
\centering
\includegraphics[width=0.8\linewidth]{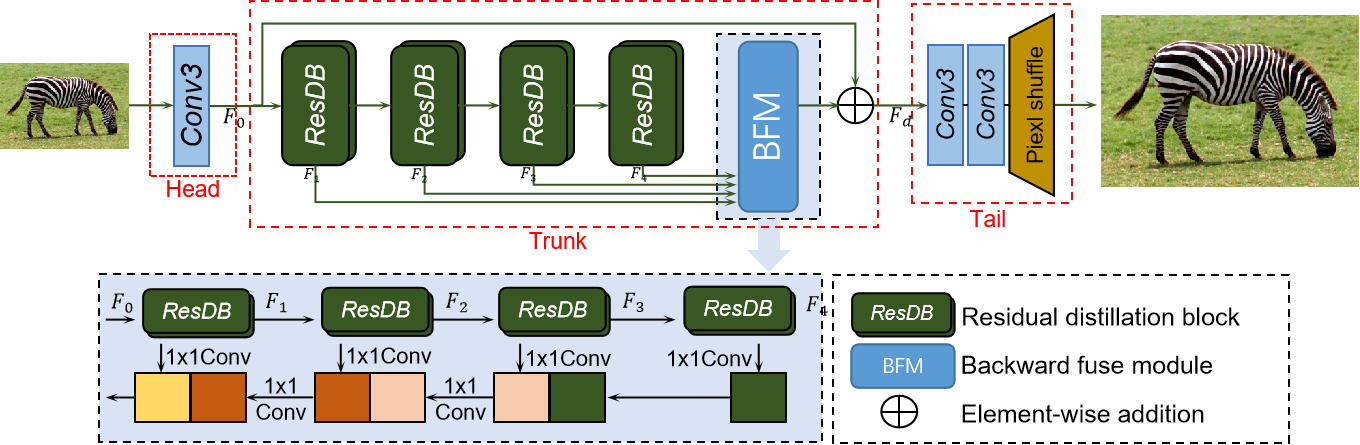}
\caption{\textit{cipher Team:} The architecture of the residual distillation network (ResDN).}
\label{fig:cipher_Fig1}
\end{figure*}

\begin{figure*}[htp]
\centering
\includegraphics[width=0.8\linewidth]{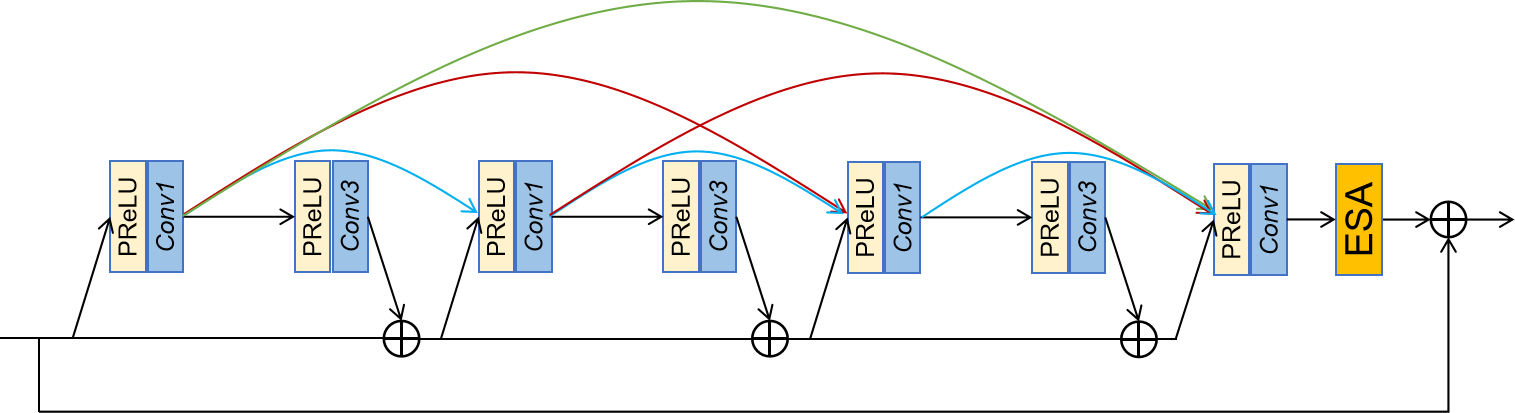}
\caption{\textit{cipher Team:} The architecture of residual distillation block (ResDB). 1$\times$1 convolution is used to expand the channels for information distillation. The color branch transmits the distilled features  to later residual blocks. The number of distillation channels is 16.}
\label{fig:cipher_Fig2}
\end{figure*}

The overview of IMDN is illustrated in \cref{fig:xilinxSR_imdn}. It is a lightweight information multi-distillation network composed of the cascaded information multi-distillation blocks (IMDB). Specifically, it adopts a series of IMDB blocks (default 8) and a traditional upsampling layer (pixelshuffle) for high-resolution image restoration. 

\textbf{Network Pruning.} Based on IMDN, network pruning is firstly performed. \cref{tab:imdn_model} provides models of different sizes and the corresponding accuracy. To achieve a trade-off between accuracy and runtime, the number of IMDB blocks is reduced from 8 to 7 as the baseline. 

\begin{table}[htbp]
\begin{center}
\begin{tabular}{l|c|c|c}
\toprule
Method  & \#IMDB & Params & PSNR  \\
\midrule
target  &     &           &29.00dB \\
\hline 
IMDN & 8     & 0.8939M    &  29.13dB  \\
IMDN & 7     & 0.7905M    &  28.97dB  \\
IMDN & 6     & 0.6871M    &  28.93dB  \\
IMDN & 5     & 0.5836M    &  28.91dB  \\
IMDN & 4     & 0.4802M    &  28.85dB  \\
\bottomrule
\end{tabular}
\end{center}
\caption{Comparison of IMDN with different IMDB numbers and corresponding accuracies on DIV2K valition.}
\label{tab:imdn_model}
\end{table}

\textbf{Collapsible Block.}
Inspired by SESR~\cite{bhardwaj2021collapsible}, the team applied a collapsible block to improve the pruned IMDN accuracy. Specifically, as shown in \cref{fig:xilinxSR_rep}, a dense block is adopted to enhance the representation during training. Each $3\times3$ convolution in IMDB is replaced by a $3\times3$ convolution and a $1\times1$ convolution. These two convolutions are conducted in parallel and the outputs are summed. During inference, the two parallel convolutions are converted to one $3\times3$ convolution.

\textbf{Training strategy.} The network was trained on DIV2K with Flick2K as the extra dataset. The training patch size is progressively increased from $64 \times 64$ to $128 \times 128 $ to improve the performance. The batch size is 32 and the number of epochs is 500. 
The network is trained by minimizing L1 loss with Adam optimizer and
a dynamic learning rate ranging from 
$2 \times 10^{-4}$ to 
$1 \times 10^{-5}$.
Data augmentation, like rotation and horizontal flip, is applied.

\subsection{cipher}

The cipher team proposed an end-to-end residual distillation network (ResDN) for lightweight image SR. As shown in \cref{fig:cipher_Fig1}, the proposed ResDN consists of three parts: the head, trunk and tail parts.

The trunk part consists of four ResDBs and one BFM. After the coarse features $F_0$ is obtained, the four ResDBs will extract intermediate features in turn, namely
\begin{gather}
F_i=\mathcal{H}_{ResDB}^i\left(F_{i-1}\right), i=1, 2, 3, 4,
\end{gather}
where $\mathcal{H}_{ResDB}^i\left(\cdot\right)$ denotes the function of the $i$-th ResDB, and $F_i$ represents the intermediate features extracted by the $i$-th ResDB. 

As shown in \cref{fig:cipher_Fig1}, the $F_i \left(i=1, 2, 3, 4\right)$ will be aggregated into BFM and the feature dimensions is first halved by $1\times1$ convolution followed by the ReLU activation function (omitted in \cref{fig:cipher_Fig1}), and then sequential concatenations are utilized. This can be formulated as
\begin{gather}
T_i=
\begin{cases}
F_i ,i=4, \\
Concat(Conv1(F_i) ,Conv1(T_{i+1})),i=3,2,1,
\end{cases}
\end{gather}
where $Concat\left(\cdot\right)$ and $Conv1\left(\cdot\right)$ denote the concatenation operation along the channel dimension and $1\times1$ convolution, respectively. By the sequential fusion, different hierarchical features can be used more fully. Finally, the coarse features $F_0$ will be transmitted by a residual connection to generate the deep features $F_d$.

As shown in \cref{fig:cipher_Fig2}, the body of ResDB is stacked by several residual blocks (RBs). Here, there is a PReLU activation function in front of each convolution layer, and a learnable parameter is set for each channel in PReLU. In RB, $1\times1$ convolution is first used to expand the channel dimension for the convenience of distillation. Suppose there are $K$ RBs in total and the input  of the $k$-th RB is $F_{res}^k$ with $c$ channels, and the number of  distilled feature channel is $d$. In $k$ RB, the intermediate feature obtained by $1\times1$ convolution can be expressed as
\begin{gather}
F_{inter}^k=\mathcal{H}_{Conv1}^{c+(K-k)*d} (\delta(F_{res}^{k-1}))
\end{gather}
where $\delta$ denotes the PReLU activation function. $\mathcal{H}_{Conv1}^{(K-k)*d+c}$ denotes the $1\times1$ convolution with $c+(K-k)*d$ convolutional kernels. Then, the intermediate features are split along the channel axis, and each distilled features with $d=16$ channels flow to the latter RBs. 
And the retained features with $c=48$ channels flow to the $3\times 3$ convolution for further refinement.
Moreover, at the beginning of each residual branch, the distilled features are concatenated on the previous residual branches of RBs and the input feature of current RB. Finally, ESA and a skip connection are used to generate the output features of ResDB.

\subsection{NJU\_MCG}

\begin{figure}[!ht]
    \centering
    \begin{subfigure}[t]{\linewidth}
		\centering
        \includegraphics[width=0.6\linewidth]{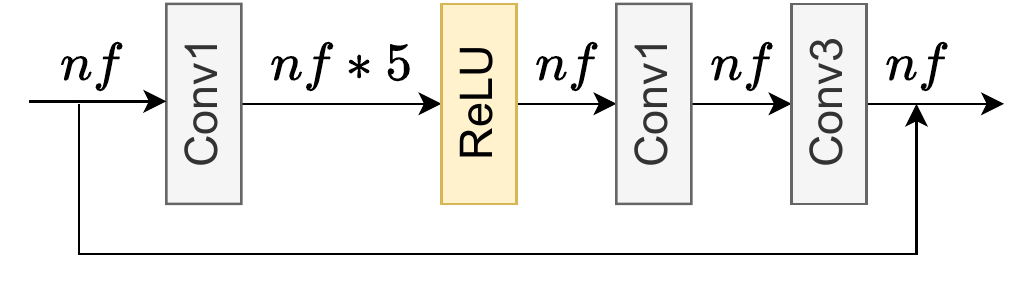}
        \caption{\textit{NJU\_MCG Team:} Residual Feature Exansion Block (RFEB)}\label{fig:RFEB}		
	\end{subfigure}
    \begin{subfigure}[t]{\linewidth}
		\centering
        \includegraphics[width=\linewidth]{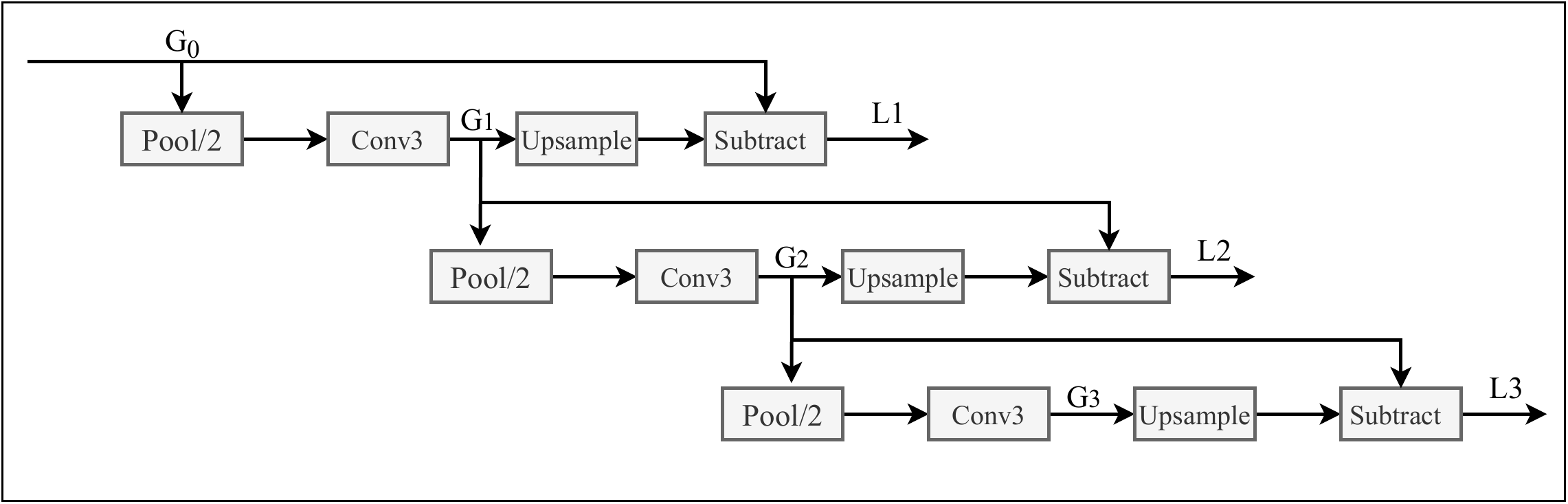}
        \caption{\textit{NJU\_MCG Team:} Laplacian Pyramid (LapPyra)}\label{fig:LapPyra}		
	\end{subfigure}
    \begin{subfigure}[t]{\linewidth}
		\centering
        \includegraphics[width=\linewidth]{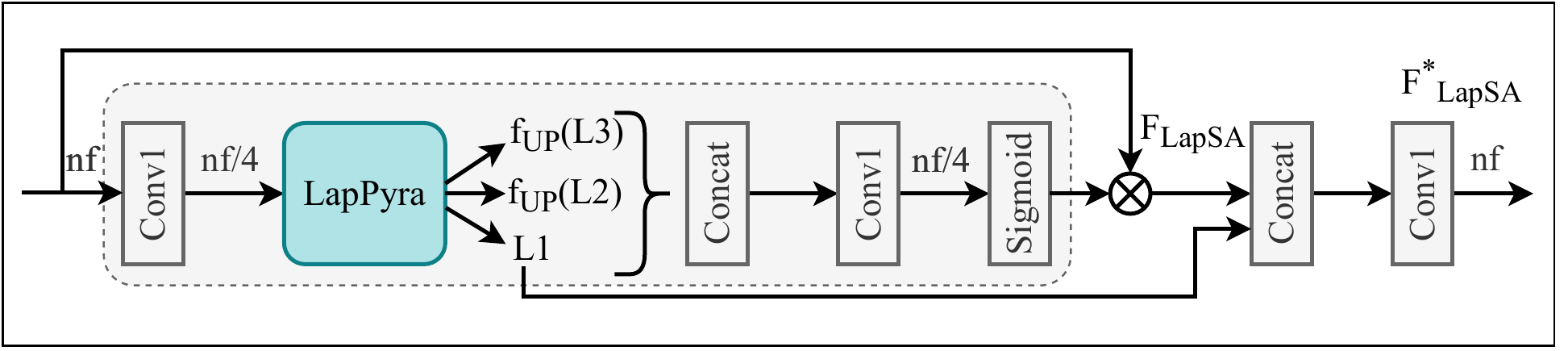}
        \caption{\textit{NJU\_MCG Team:} Laplacian Attention (LapSA)}\label{fig:LapSA}		
	\end{subfigure}
	\caption{\textit{NJU\_MCG Team:} The proposed solution.}
\end{figure}

\begin{figure*}[!t]
\centering
\includegraphics[scale=0.3]{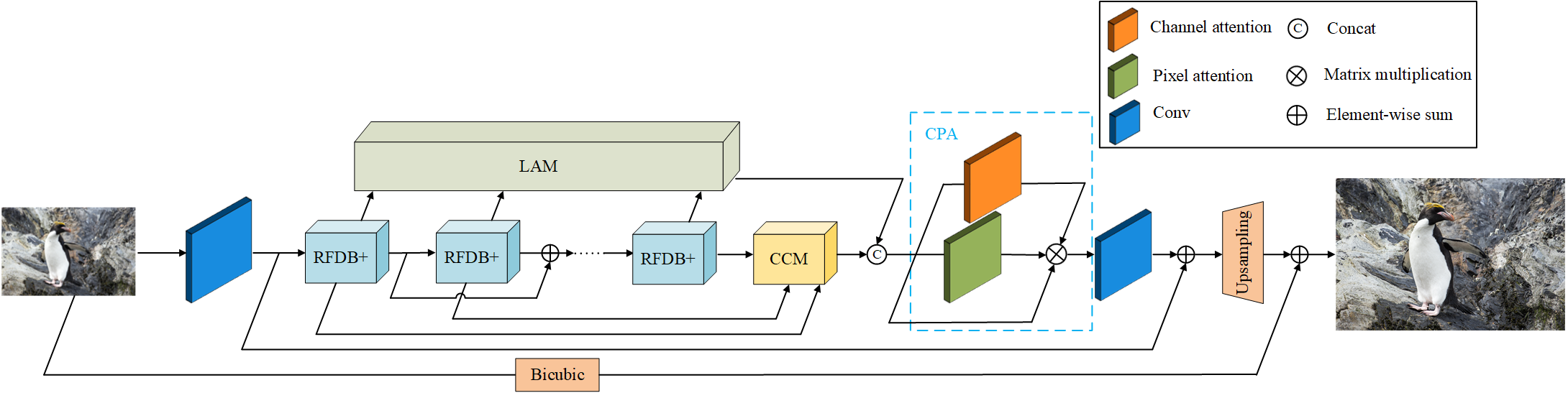} 
\caption{\textit{IMGWLH Team:} The architecture of RLCSR network.}
\label{fig:IMGWLH1}
\end{figure*}

\begin{figure*}[!t]
\centering
\includegraphics[scale=0.4]{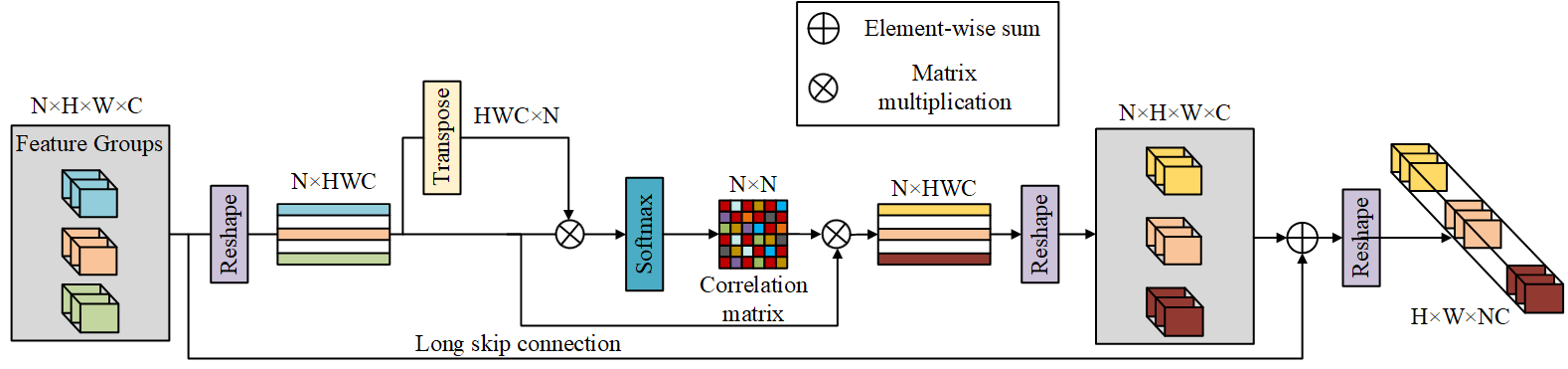} 
\caption{\textit{IMGWLH Team:} The architecture of LAM module.}
\label{fig:IMGWLH3}
\end{figure*}

Inspired by WDSR~\cite{yu2018wide}, the residual feature expansion block (RFEB) in \cref{fig:RFEB} is used as the basic unit of the feature distillation and expansion network (FDEN). 
FDEN adopts the architecture of RFDN~\cite{RFDN} except that the basic unit is replaced by RFEB and the attention mechanism 
is replaced by the LapSA module in \cref{fig:LapSA}.
The LapSA module is responsible for applying the scale transformation for each spatial position of the input features. To achieve this goal, it needs to have a global assessment to assign different scaling factors for different positions according to the spatial importance.
Specifically, for the task of image super-resolution, the network should concentrate more on the high-frequency regions that are usually difficult to recover because of the complex details. As a consequence, the LapSA module is implemented to contain a Laplacian 
pyramid (\cref{fig:LapPyra}) which 
has a large receptive field and can extract the high-frequency details as well. This process can be formulated as
\begin{align}
    \begin{split}
        \bm{G}_1 = f_{G}(\bm{G}_0;\bm{\theta}_{G_1}), \bm{L}_1 = \bm{G}_0 - f_{UP}(\bm{G}_1),\\
        \bm{G}_2 = f_{G}(\bm{G}_1;\bm{\theta}_{G_2}), \bm{L}_2 = \bm{G}_1 - f_{UP}(\bm{G}_2),\\
        \bm{G}_3 = f_{G}(\bm{G}_2;\bm{\theta}_{G_3}), \bm{L}_3 = \bm{G}_2 - f_{UP}(\bm{G}_3).
    \end{split}
\end{align}
As depicted in \cref{fig:LapPyra}, $f_{G}$ denotes the downsampling function that consists of a pooling layer and a $3\times3$
convolutional layer. $f_{UP}$ is the interpolation function that upsamples the input feature. $\bm{L}_1$, $\bm{L}_2$ and $\bm{L}_3$
are the output features of the three pyramid levels, respectively. The extracted feature $\bm{L}_{j\in[1,2,3]}$ contains high-frequency
information and is used in the LapSA module (\cref{fig:LapSA}) to generate the scaling factors.

As shown in \cref{fig:LapSA}, the first $1\times1$ convolution is used to reduce the channel dimension and the last $1\times1$ convolution
is used to recover the channel dimension. The middle $1\times1$ convolution is used to aggregate the pyramid features and then the sigmoid  function is applied to generate the final scaling factors. $\bm{L}_1$ is concatenated with the scaled features to augment the output features $\bm{F}_{LapSA}^*$ as
\begin{align}
    \begin{split}
        \bm{F}_{LapSA}^* = f_{Conv1}([\bm{F}_{LapSA}, \bm{L}_1]).
    \end{split}
\end{align}

The number of filters ($nf$) in FDEN is set to 29. The proposed FDEN is trained with the same training setting as RFDN.
DIV2K and Flickr2K datasets are adopted as the training dataset. 

\begin{figure*}[!t]
\centering
\includegraphics[scale=0.5]{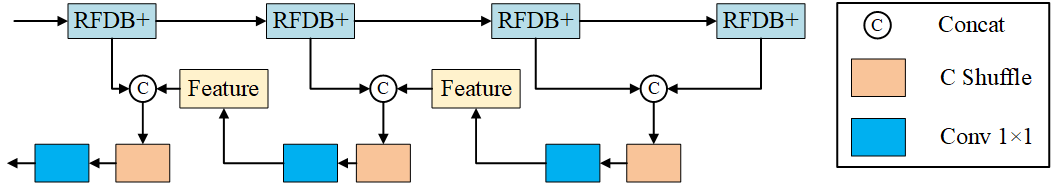} 
\caption{\textit{IMGWLH Team:} The architecture of CCM module.}
\label{fig:IMGWLH2}
\end{figure*}

\subsection{IMGWLH}

The IMGWLH team proposed RLCSR network. The overall network structure is shown in \cref{fig:IMGWLH1}.
A $3\times3$ convolutional layer is firstly used to extract shallow features from low-resolution images.
Then six Local residual feature fusion block (RFDB+) modules are then stacked to perform deep feature extraction on the shallow features. RFDB+ is an improved version of RFDB \cite{RFDN}. Without increasing the number of parameters of the ESA module, more skip connections are included to ensure better retention of useful information and use dilated convolution to expand the receptive field for preserving more texture details.

In order to produce compact features, a CCM model is introduced to fuse the intermediate features from several RFDB+. The module is developed based on MBFF module~\cite{muqeet2020multi} and the backward fusion model~\cite{luo2020latticenet}. The detailed structure of CCM is illustrated in \cref{fig:IMGWLH2}. It can be observed that different levels of features from several RFDB+ are gradually fused to a single feature map using our proposed CCM. The channel shuffle operation and $1\times 1$ convolution kernel can integrate the features of all basic residual blocks, which helps to extract more contextual information in a compact manner.

To further enhance the intermediate features produced by RFDB+,  LAM~\cite{niu2020single} module is used to introduce an attention mechanism for adaptively selecting representative features from multiple intermediate features. One can refer to \cref{fig:IMGWLH3} for the structure of LAM.

However, the collected features could still be full of redundancy. To address the issue, a weight is assigned to the feature map through the Hadamard multiplication of channel attention and pixel attention. To produce features for high-resolution reconstruction, a long-term skip connection is included, with which the deep features are added to the shallow features that are extracted at the beginning of the network. 

\begin{figure*}[!t]
    \centering
    \includegraphics[width=15cm]{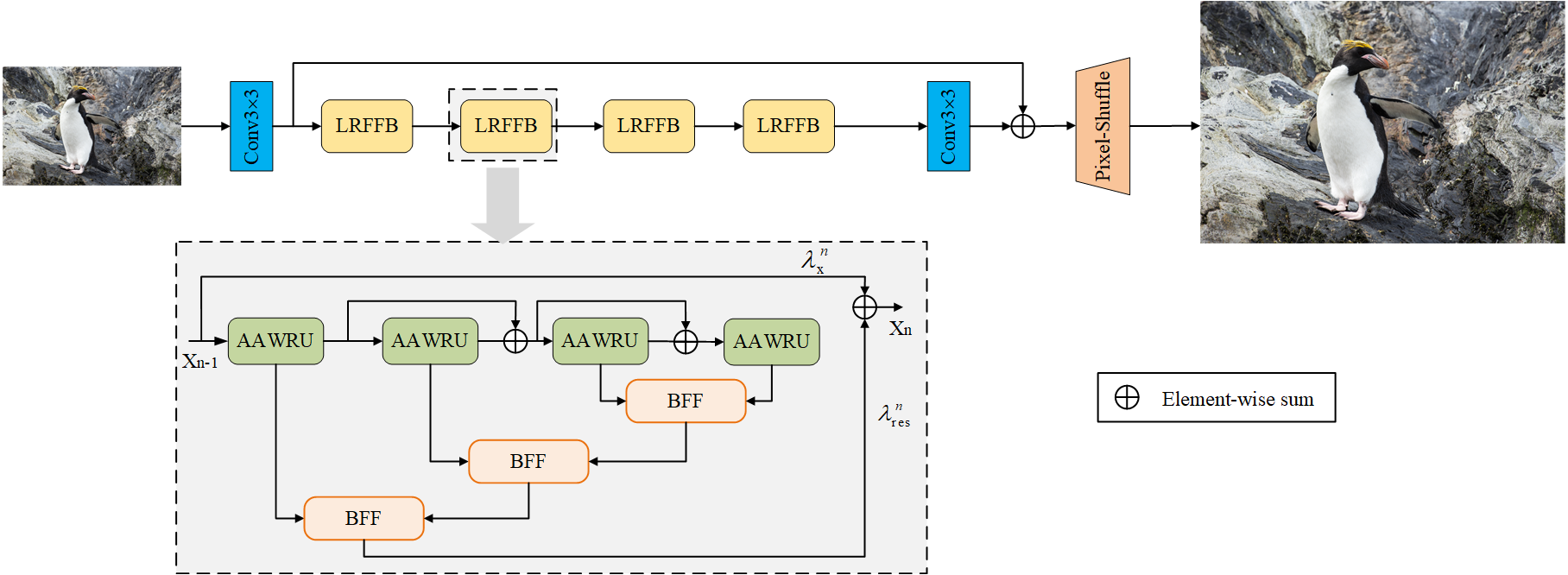}
    \caption{\textit{imgwhl team:} The network structure of the proposed method.}
    \label{fig:imgwhl1}
\end{figure*}

\begin{figure*}[!t]
    \centering
    \includegraphics[width=15cm]{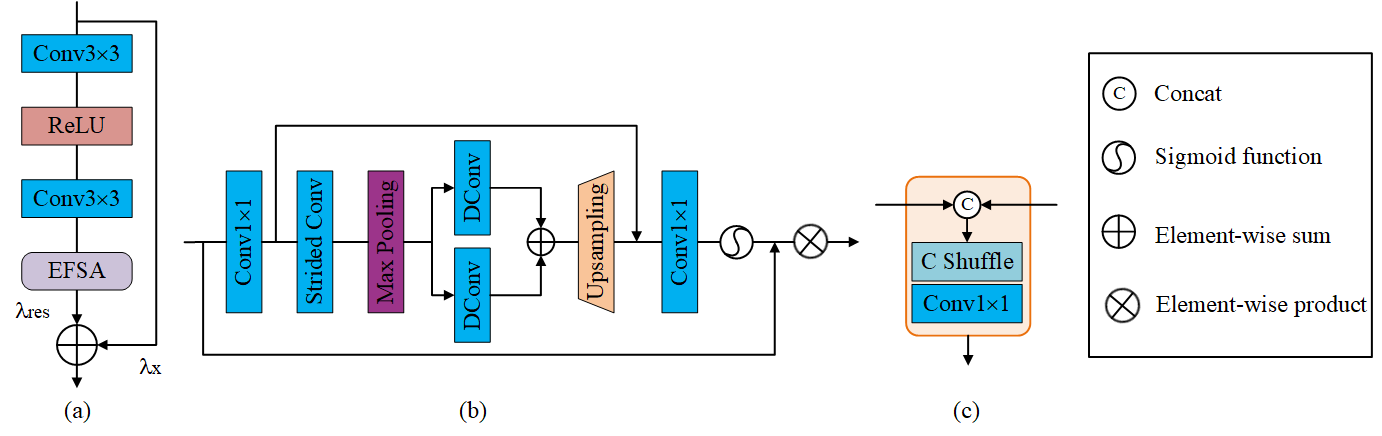}
    \caption{\textit{imgwhl team:} (a) Structure of AAWRU. (b) Structure of EFSA. (c) Structure of BFF.}
    \label{fig:imgwhl2}
\end{figure*}

\subsection{imgwhl}

\begin{figure*}[htp]
  \centering
  \includegraphics[width=.9\linewidth]{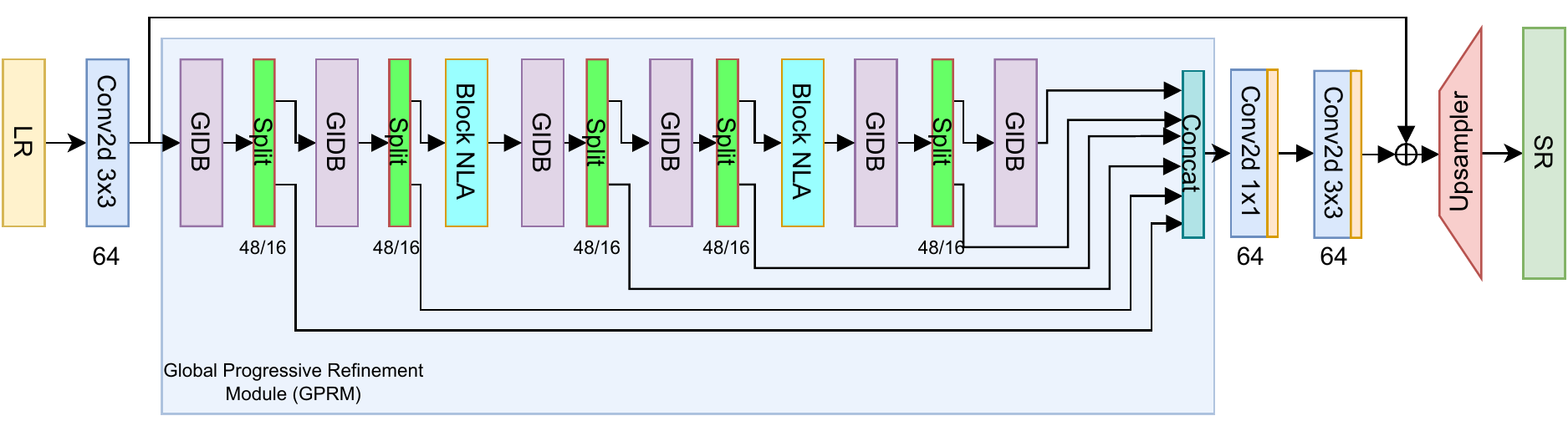}
  \caption{\textit{Aselsan Research Team:} The proposed architecture of IMDeception.}
  \label{fig:IMDeceptionNetwork}
\end{figure*}

The imgwhl team proposed a lightweight SR network named RFESR to achieve a compact network design and fast inference speed. To be specific, the work is based on the structure of IMDB~\cite{IMDN} and inspired by several advanced techniques \cite{muqeet2020multi, RFANet}. 

The proposed network is shown in \cref{fig:imgwhl1}. A $3\times3$ convolution is first used to extract shallow features from inputs. Then, four Local residual feature fusion block (LRFFB) modules are stacked to perform deep feature extraction on the shallow features. After gradual feature refinement by the LRFFBs, another $3\times3$ convolution is used to extract final deep features from the output of the last LRFFN module. The final deep and shallow features are element-wise added through a long-range skip connection. Finally, the high-resolution images can be reconstructed through a pixel Shuffle block that consists of a $3\times3$ convolution and a non-parametric sub-pixel operation.

The two building blocks are presented as follows.

\textbf{LRFFB.}
Each LRFFB module contains four basic residual units, \ie, Attention-guided Adaptive Weighted Residual Unit (AAWRU). Inspired by the MBFF module \cite{muqeet2020multi}, the backward feature fusion (BFF) module is introduced to fuse multi-level features acquired from AAWRUs.  The feature extracted by $j_{th}$ AAWRU of $i_{th}$ LRFFB is denoted by $F_{ij}$. For example, the feature extracted by the second AAWRU in the first LRFFB is $F_{12}$. Specifically, in the $i_{th}$ LRFFB, the last two features ($F_{i3}$ and $F_{i4}$) are aggregated by a BFF module.
The structure of the BFF module is shown in \cref{fig:imgwhl2}(c). It first concatenates the two input features and then processes the aggregated features by a channel shuffle operation \cite{zhang2018shufflenet} and $1\times1$ convolution kernel. The BFF module is repeated three times until all level features are fused in an LRFFB. The input features are then added to the output fused feature in an element-wise manner. As residue may contain redundant information, the results are multiplied with trainable parameters to select useful information.

\textbf{AAWRU.}
The detailed structure of the AAWRU module is shown in \cref{fig:imgwhl2}(a). Inspired by the residual block proposed in the RFANet \cite{RFANet}, MAFFSRN \cite{muqeet2020multi} introduced an enhanced fast spatial attention module (EFSA). It aims to realize spatial attention weighting to make the features more concentrated in some desired regions, so that more representative features can be obtained. The two branches of the residual structure in AAWRU are assigned by adaptive weights, which help more shallow-level features be activated without increasing parameters. The design of the EFSA module is shown in \cref{fig:imgwhl2}(b). Using the blocks above, the proposed model can better extract and integrate compact contextual information with fewer parameters, which helps produce more delicate SR images.

\subsection{whu\_sigma}

The team designed their method based on RFDN~\cite{RFDN}. They simply used dilated convolution to replace the convolution part in the RFDB module and adjusted the number of channel of RFDN to 64.

\begin{figure}
  \centering
  \includegraphics[width=.60\linewidth]{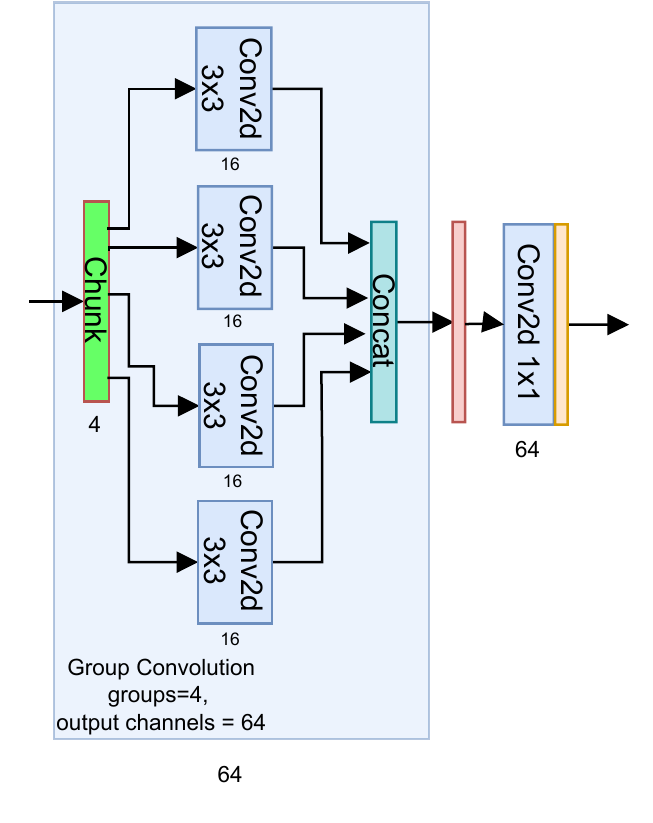}
  \caption{\textit{Aselsan Research Team:} Gblock. Red and orange stripes stands for ReLU and LeakyReLU activation.}
  \label{fig:Gblock}
\end{figure}

\begin{figure}
  \centering
  \includegraphics[width=1.0\linewidth]{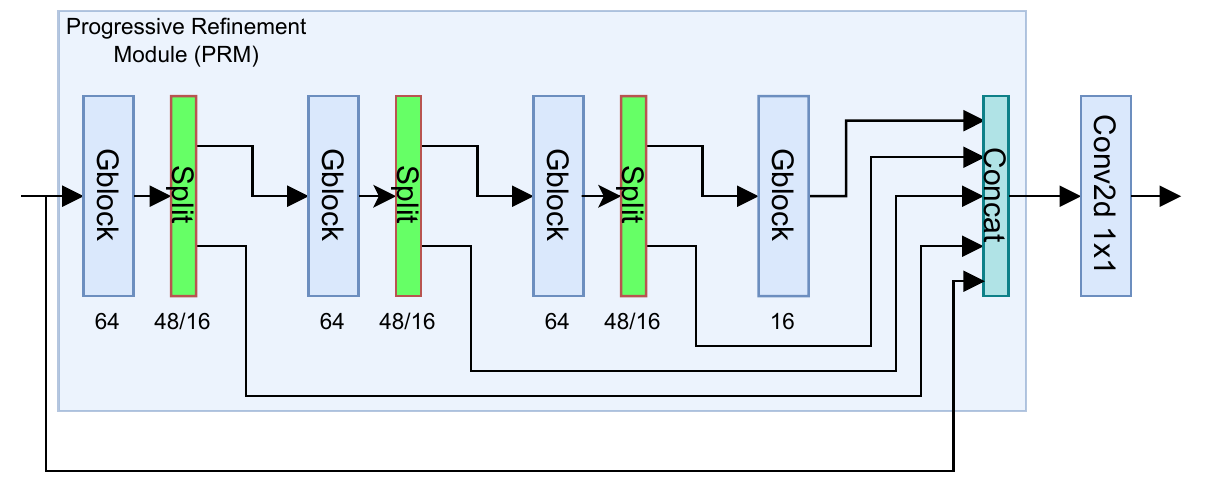}
  \caption{\textit{Aselsan Research Team:} GIDB Block}
  \label{fig:GIDB}
\end{figure}

\begin{figure*}[!ht]
	\centering
	\includegraphics[width=0.8\linewidth]{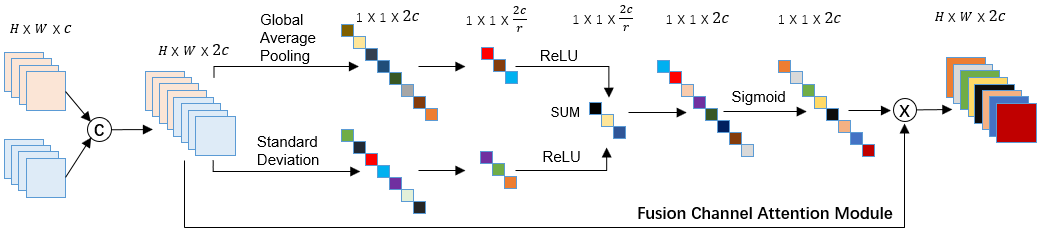}
	\caption{\textit{Drinktea Team:} Details of the fusion channel attention module.}
	\label{fig:drinktea_fca}
\end{figure*} 

\begin{figure*}[!ht]
	\centering
	\includegraphics[width=0.6\linewidth]{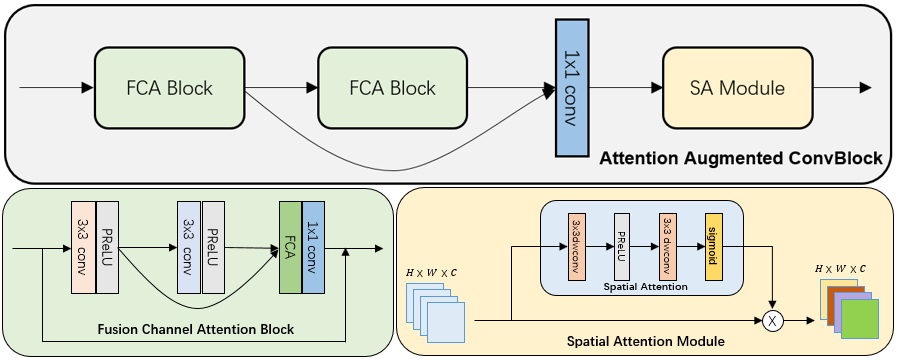}
	\caption{\textit{Drinktea Team:} Details of the fusion channel attention block.}
	\label{fig:drinktea_basicblock}
\end{figure*} 

\begin{figure}[!ht]
	\centering
	\includegraphics[width=\linewidth]{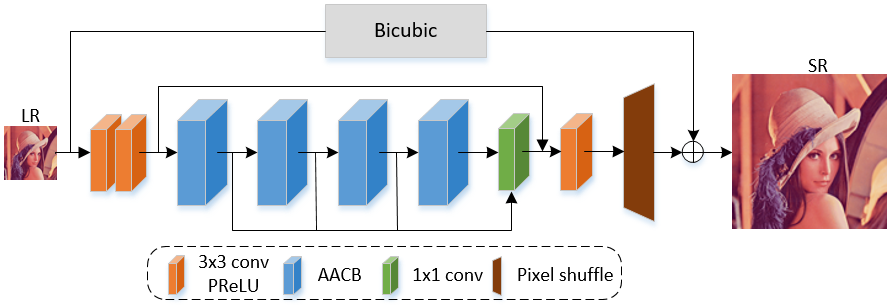}
	\caption{\textit{Drinktea Team:} The architecture of attention augmented lightweight network (AALN).}
	\label{fig:drinktea_aaln}
\end{figure} 

\subsection{Aselsan Research}

The team created a network structure where progressive refinement module (PRM) is repeated locally in the blocks and globally among the blocks to reduce the number of parameters. 
This is done in a way that intermediate information collection (IIC) modules in the global setting is replaced with proposed Global PRM.  
Furthermore, block based non-local attention block~\cite{wang2018non} is employed in the main path of the network while is avoided in the individual IMDB blocks. 
To further reduce the number of parameters and number of operations of the network, every single convolution operation inside the IMDB is replaced with Gblocks (\cref{fig:Gblock}) as in XLSR~\cite{ayazoglu2021extremely} which is based on group convolution.
The group convolution based structures is referred to as Grouped Information Distilling Blocks (GIDB). Yet, grouped convolutions are unfortunately not well optimized in PyTorch framework~\cite{gibson2020optimizing}. If implemented properly within an inference oriented framework, group convolutions can lead to speedups~\cite{ayazoglu2021extremely,gibson2020optimizing} especially in mobile devices where efficient network structures are usually employed. 

\subsection{Drinktea}

Inspired by IMDN and LatticeNet~\cite{luo2020latticenet}, the Drinktea team proposed a method to obtain channel attention which can effectively utilize the mean and the standard deviation of feature maps. First, as shown in \cref{fig:drinktea_fca}, the mean value is calculated by global average pooling and the standard deviation of each feature map is also computed. Second, the statistic vector in each branch is passed to a $1 \times 1$ convolution layer which performs channel-downscaling with reduction ratio $ r $ and then activated by ReLU. 
Third, the two vectors are added up to fuse the information extracted by each vector. Then the fusion vector is restored to the original number of channels. Finally, the sigmoid activation is utilized to weight the vector to generate channel attention. FCAB utilizes the features from different hierarchical and augments them with channel attention~\cite{xiong2021attention}.

Spatial attention can effectively improve the performance of the model, but it is difficult to achieve a balance between complexity and performance in lightweight network. Inspired by ULSAM~\cite{saini2020ulsam}, the team designed a spatial attention module for super-resolution task. As shown in \cref{fig:drinktea_basicblock}, the features on each channel is first extracted with depthwise convolution and activated with PReLU function. Then another depthwise convolution and sigmoid function are applied to redistribute the weights. As shown in \cref{fig:drinktea_basicblock}, the basic the Attention Augmented ConvBlock (AACB) is composed of two FCABs, a $1 \times 1$ convolution and a spatial attention module. The architecture of attention augmented lightweight network (AALN) for efficient image SR is shown in \cref{fig:drinktea_aaln}.

\subsection{GDUT\_SR}

\begin{figure*}[htbp]
    \centering
    \includegraphics[width=0.95\linewidth]{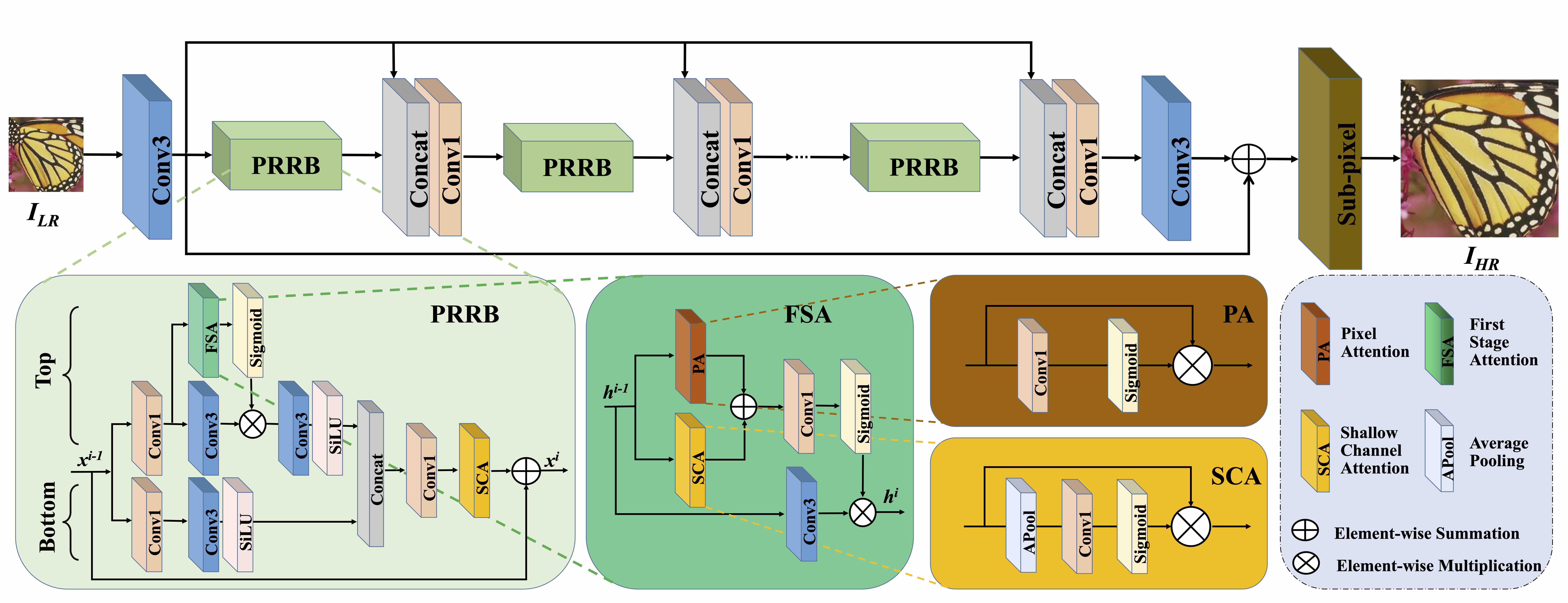}
    \caption{\textit{GDUT\_SR Team:} The overall architecture of the progressive representation re-calibration network (PRRN).}
    \label{fig:PRRN}
\end{figure*}

The GDUT\_SR proposed \textit{Progressive Representation Re-Calibration Network (PRRN)} for lightweight SR. The proposed PRRN shown in \cref{fig:PRRN} is modified from PAN~\cite{PAN} but achieves better performance and runtime efficiency than PAN with limited increase of parameters. The main contribution of PRRN is to adjust the receptive field of CNN by using the pixel and channel information in a two-stage manner. A shallow channel attention (SCA) mechanism is proposed to build the correspondences between channels in a simpler yet more efficient way. The architecture of PRRN can be divided into three components: shallow feature extractor, deep feature extractor, and reconstruction. The shallow feature is extracted by using $3 \times 3$ convolution layer, while the deep feature extractor is stacked by the proposed Progressive Representation Re-calibration Blocks (PRRBs). Finally, the pixel shuffle layer is used to reconstruct the HR image.

The deep feature extractor consists of 16 PRRBs and multiple long skip connections are applied to propagate the initial features to the intermediate layers. PRRB precisely explores the discriminative information in a two-stage manner. In the first stage, the First Stage Attention (FSA) uses pixel attention (PA)~\cite{PAN} to capture important pixel information and the proposed SCA mechanism is applied to learn useful channel information. Therefore, the first stage of PRRB can explore the spatial and channel information simultaneously. In the second stage, an SCA modified from squeeze-and-excitation (SE)~\cite{hu2018squeeze} is used to further rescale the importance of the output feature channels. SCA uses average pooling to collect channel information and then uses $1 \times 1$ convolution and sigmoid activation function to process the information. Moreover, inspired by the recent work~\cite{lin2022revisiting}, \textbf{SiLU} activation is used at the end of both the top and bottom branches of FSA.

\subsection*{Giantpandacv}

\begin{figure*}
	\centering
	\includegraphics[width=0.85\textwidth]{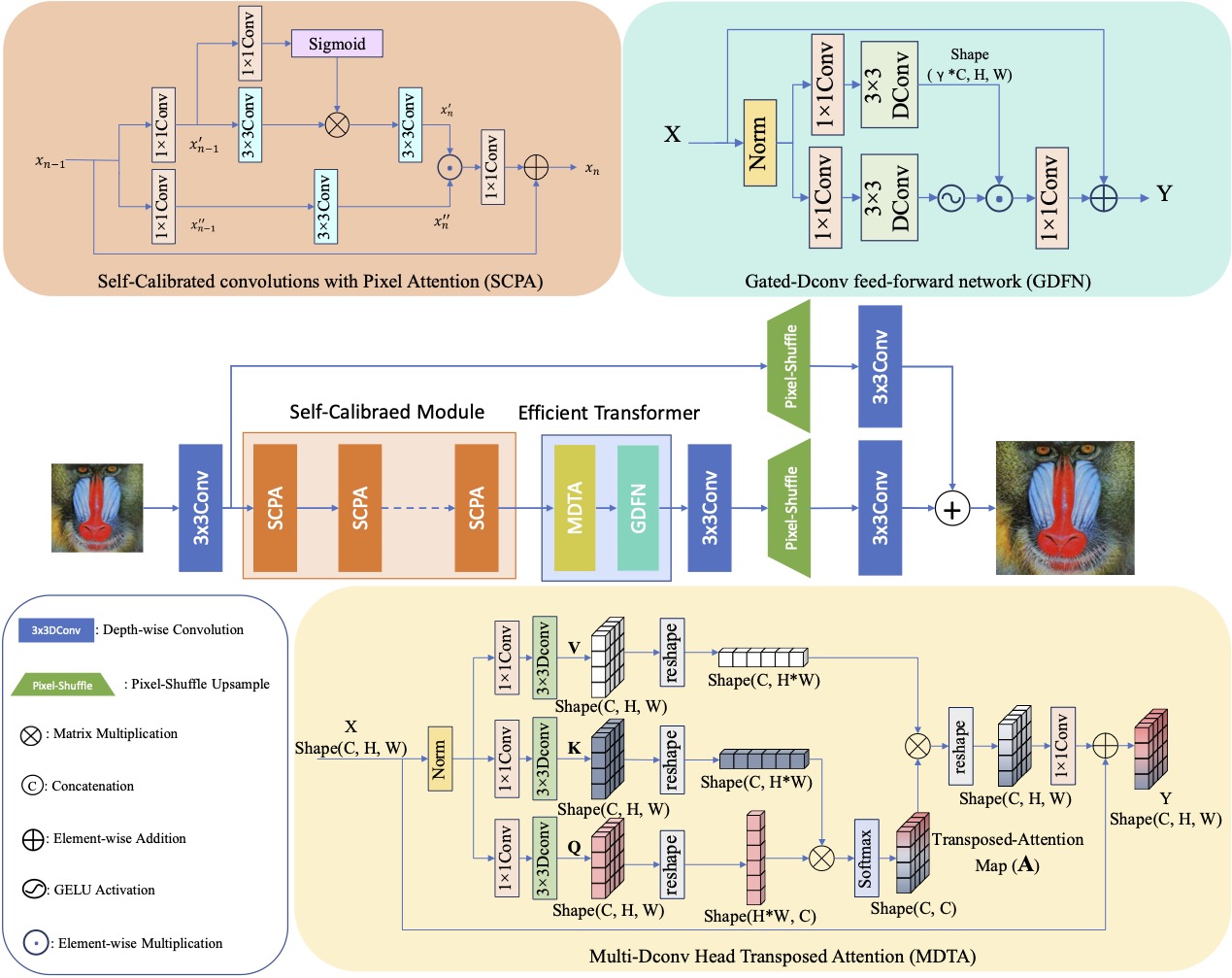}\\
	\caption{\textit{Giantpandacv Team:} Self-Calibrated Efficient Transformer (SCET) Network.}
	\label{fig:scet}
\end{figure*}

\begin{figure}[t]
\begin{center}
\includegraphics[width=\linewidth]{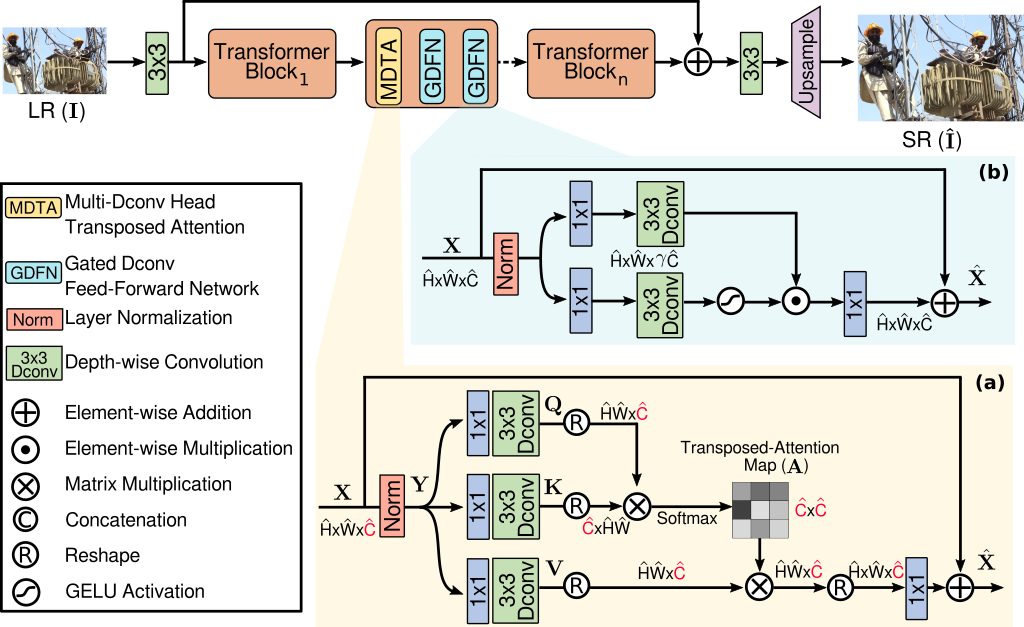}
\end{center}
\caption{\textit{TeamInception:} Overall framework of Restormer~\cite{Zamir2021Restormer}.}
\label{fig:Restormer_framework}
\end{figure}

The Giantpandacv team proposed a lightweight Self-Calibrated Efficient Transformer (SCET) network, which is inspired PAN~\cite{PAN} and Restormer~\cite{Zamir2021Restormer}. The architecture of SCET mainly consists of the Self-Calibrated module and Efficient Transformer block, where the Self-Calibrated module adopts the pixel attention mechanism to extract image features effectively. To further exploit the contextual information from features, an efficient Transformer module is employed to help the network obtain similar features over long distances and thus recover sufficient texture details. 

The main architecture of the network is shown in \cref{fig:scet}, which consists of the Self-Calibrated module and the Efficient Transformer block. The details of these modules of SCET are described as follows.

\textbf{Self-Calibrated module.} In this module, 16 cascaded  Self-Calibrated convolutions with Pixel Attention (SCPA) blocks are utilized for a larger receptive field. The SCPA block~\cite{PAN} consists of two branches. One of the branches is equipped with a pixel attention module to perform the attention mechanism in the spatial dimension while the other branch is used to retain the original high-frequency feature information. Furthermore, skip connection is utilize to facilitate network training.

\textbf{Efficient Transformer.} The efficient transformer block is utilized to further exploit the contextual information from features to obtain useful contextual information. In the efficient transformer block, Multi-Dconv Head Transposed Attention (MDTA) is used to avoid the vast computational complexity of the traditional self-attention mechanism. And a feed-forward network is further employed with a gating mechanism to recover precise texture details.

\subsection{TeamInception}

The proposed solution is based on the Transformer-based architecture \emph{Restormer} that is recently introduced in \cite{Zamir2021Restormer}. Specifically, an isotropic version of Restormer is built, which operates at the original resolution and does not contain any downsampling operation.

\textbf{Overall pipeline.} The overall pipeline of the Restormer architecture is presented in \cref{fig:Restormer_framework}. Given a low-resolution image $\mathbf{I}$~$\in$~$\mathbb{R}^{H\times W \times 3}$, Restormer first applies a convolution to obtain low-level feature embeddings $\mathbf{F_0}$~$\in$~$\mathbb{R}^{H\times W \times C}$; where $H\times W$ denotes the spatial dimension and $C$ is the number of channels. Next, these shallow features $\mathbf{F_0}$ pass through multiple transformer blocks (six in this work) and transformed into deep features $\mathbf{F_d}$~$\in$~$\mathbb{R}^{H\times W \times C}$, to which shallow features $\mathbf{F_0}$ are added via skip connection.
Finally, a convolution layer followed by pixel shuffle layer is applied to the deep features $\mathbf{F_d}$ to generate residual high-resolution image $\mathbf{R}$~$\in$~$\mathbb{R}^{sH\times sW \times 3}$, where $s$ denotes the scaling factor. To obtain the final super-resolved image, the residual image is added the bilinearly upsampled input image as: $\mathbf{\hat{I}} = \text{bilinear-up}(\mathbf{I}) + \mathbf{R}$.

In the proposed Transformer block, the core components are: \textbf{(a)} multi-Dconv head transposed attention (MDTA) and \textbf{(b)} gated-Dconv feed-forward network (GDFN).

\begin{figure*}[t]
\begin{center}
{\centering
	\vbox{\centerline{\includegraphics[width=17cm]{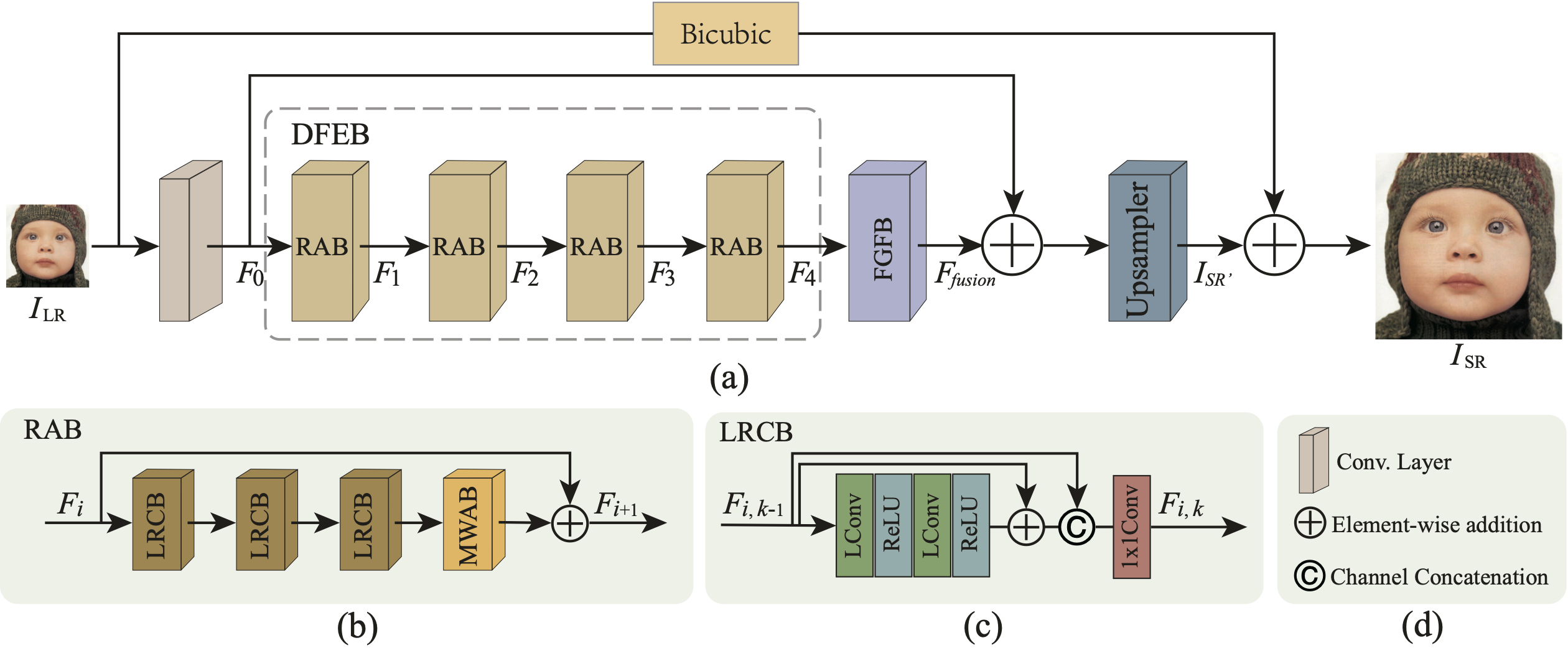}}\vskip0.5mm
	\caption{\textit{cceNBgdd Team:} (a) The architecture of our proposed very lightweight and efficient image super-resolution network (VLESR);  (b) Residual
attention block (RAB). (c) Lightweight residual concatenation block (LRCB); and (d) Sign description.}\label{VLESR}}}
\end{center}
\vskip-5mm
\end{figure*}

\begin{figure*}[t]
\begin{center}
{\centering
	\vbox{\centerline{\includegraphics[width=15cm]{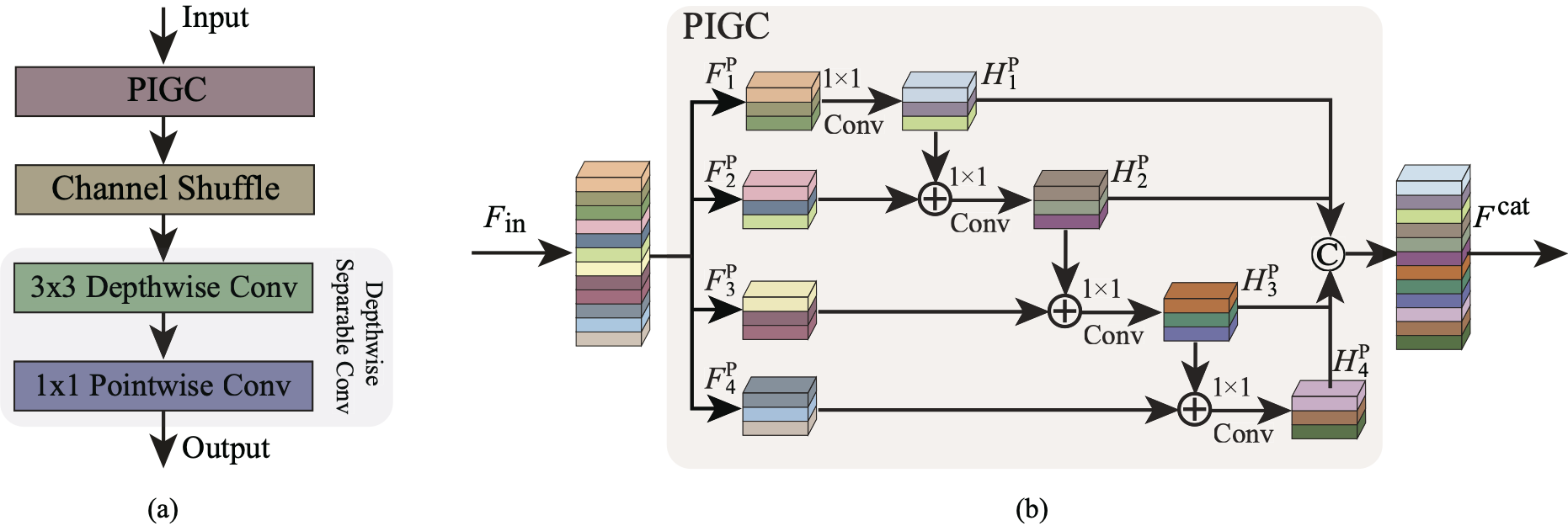}}\vskip0.5mm
	\caption{\textit{cceNBgdd Team:} (a) The structure of our proposed lightweight convolution block (LConv); and (b) Progressive interactive group convolution (PIGC).}\label{LConv}}}	
\end{center}
\vskip-5mm
\end{figure*}

\textbf{Multi-Dconv Head Transposed Attention.}
The major computational overhead in Transformers comes from the self-attention layer, which has quadratic time and memory complexity. Therefore, it is infeasible to apply SA on most image restoration tasks that often involve high-resolution images. To alleviate this issue, MDTA is proposed. The key ingredient is to apply SA across channels rather than the spatial dimension, \ie, to compute cross-covariance across channels to generate {an} attention map encoding the non-local context implicitly.
As another essential component in MDTA, {depth-wise convolutions} is introduced to emphasize on {the} local context before computing feature covariance to produce the global attention map~\cite{li2021localvit}.

\textbf{Gated-Dconv Feed-Forward Network.}
A feed-forward network (FN) is the other building block of the Transformer model~\cite{vision_transformer}, which consists of two fully connected layers with a non-linearity in between. As shown in~\cref{fig:Restormer_framework}(b), the first linear transformation layer of the regular FN~\cite{vision_transformer} is reformulated with a gating mechanism to improve the information flow through the network.
This gating layer is designed as the element-wise product of two linear projection layers, one of which is activated with the GELU non-linearity.
Our GDFN is also based on local content mixing similar to {the} MDTA module to equally emphasize on the spatial context, which is useful for learning local image structure for effective restoration.
The gating mechanism in GDFN controls which complementary features should flow forward and allows subsequent layers in the network to specifically focus on more refined image attributes, thus leading to high-quality outputs. Progressive learning is performed where the network is trained on smaller image patches in the early epochs and on gradually larger patches in the later training epochs. The model trained on mixed-size patches via progressive learning shows enhanced performance at test time where images can be of different resolutions (a common case in image restoration).

\subsection{cceNBgdd}

The VLESR network architecture  shown in \cref{VLESR} (a), mainly consists of a 3$\times$3 convolutional layer, a deep feature extraction block (DFEB), a frequency grouping fusion block (FGFB), and an Upsampler. DFEB contains four residual attention blocks (RABs), and Upsampler uses subpixel convolution.

\begin{figure}[!ht]
	\begin{center}			
		\includegraphics[width=1\linewidth]{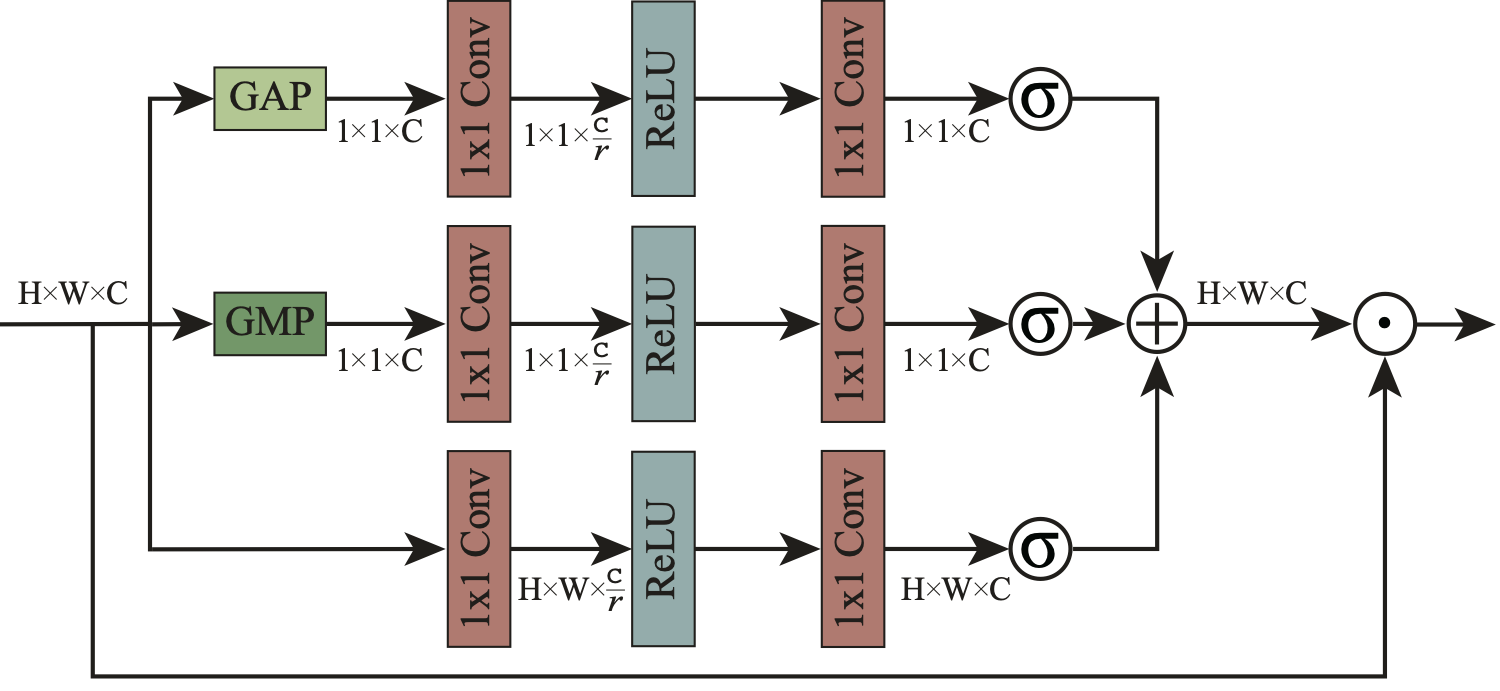}
		\caption{\textit{cceNBgdd Team:} Multi-way attention block (MWAB).}\label{MWAB}	
	\end{center}
	\vskip-5mm
\end{figure}

Each RAB contains three lightweight residual concatenation blocks (LRCBs), a multi-way attention block (MWAB), and a skip connection, as shown in \cref{VLESR} (b).
The LRCB consists of two parts, as shown in \cref{VLESR} (c). The first part contains two lightweight convolutional blocks (LConv) (see Section {\color{red}3.3}), two ReLU nonlinear activation layers immediately following each LConv and a skip connection to learn the local residual feature information. The learned residual features are concatenated with the original feature to enhance the utilization and propagation of the feature information. In the second part, a 1$\times$1 convolutional layer is used to compress the concatenated feature.
The multi-way attention block (MWAB) is shown in \cref{MWAB}. The MWAB contains three branches, where the first and the second branches focus on the global information, and the third branch focuses on the local information. The three branches explore the clues of different feature information respectively and sum the calculated importance (\ie, weights) of each channel.

\begin{figure}[!ht]
	\begin{center}			
		\includegraphics[width=0.9\linewidth]{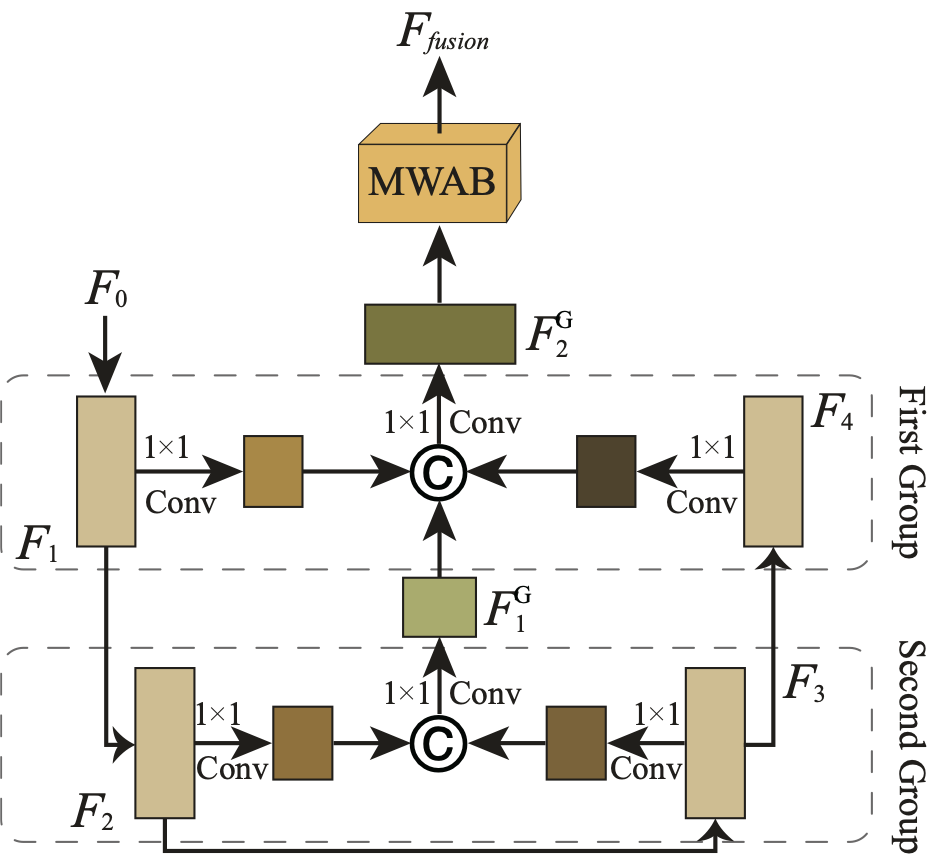}
		\caption{\textit{cceNBgdd Team:} Schematic diagram of frequency grouping fusion block (FGFB).}\label{FGFB}	
	\end{center}
\end{figure}

Based on the ShuffleNet~\cite{zhang2018shufflenet},  a very lightweight building block is designed for the SISR task, called the lightweight convolutional block (LConv). The important improvements are twofold: (1) Remove the batch normalization layers from the ShuffleNet unit, which have been shown to deteriorate the accuracy of SISR; (2) In ShuffleNet unit, the first 1$\times$1 group convolution is replaced with the progressive interactive group convolution (PIGC), and the second 1$\times$1 group convolution is replaced with the 1$\times$1 point-wise convolution to enhance the interaction between the group features. The structure of the LConv is shown in \cref{LConv} (a), which consists of a PIGC, a channel shuffle layer, a 3$\times$3 depth-wise convolution, and a 1$\times$1 point-wise convolution. The structure of the PIGC in the LConv is shown in \cref{LConv} (b).

The frequency grouping fusion block (FGFB) is shown in \cref{FGFB}. The features with the highest difference between low-frequencies and high-frequencies are divided into the first group, the features with the next highest difference are divided into the second group, and so on. Then, starting from the feature group with the smallest frequency difference, the features of each group are gradually fused until the feature group with the largest frequency difference. If the number of the RABs is odd, only the output feature of the middle RAB is used as the last feature group. The output feature by grouping fusion is then fed into the MWAB for the further fusion. When the number of the RABs is 4, there are only two feature groups.

\subsection{ZLZ}
\begin{figure}[!ht]
    \centering
    \includegraphics[width=0.9\linewidth]{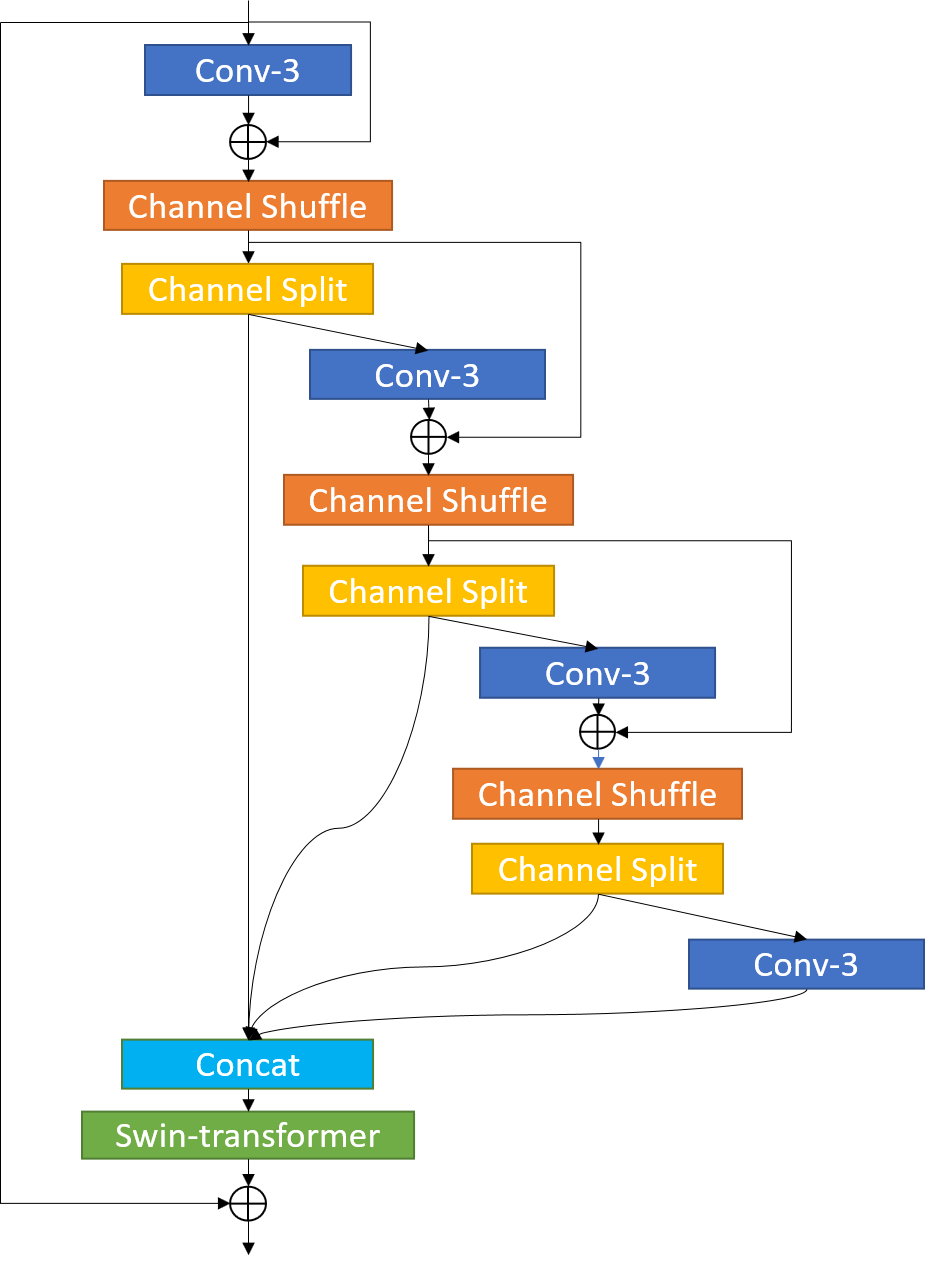}
    \caption{\textit{ZLZ Team:} IMDTB architecture diagram.}
    \label{fig:zlz_block}
\end{figure}

The team proposed to use Information Multi-distillation Transformer Block (IMDTB) in \cref{fig:zlz_block} as the basic block, where the convolution in IMDB\cite{IMDN,hui2018fast} was converted to grouped convolution and the number of groups is 4. The channel shuffling operation is used~\cite{zhang2018shufflenet} to increase the information interaction between channels. The attention mechanism is replaced with a Swin-Transformer\cite{liu2021swin,liang2021swinir} to better deal with images spatial relations with attention mechanism. 

\subsection{Express}

\begin{figure}[!ht]
	\centering
	\newlength\fsdurthreeD
	\setlength{\fsdurthreeD}{-0mm}
		\scriptsize
		\begin{tabular}{ll}
			\includegraphics[width=0.1\textwidth]{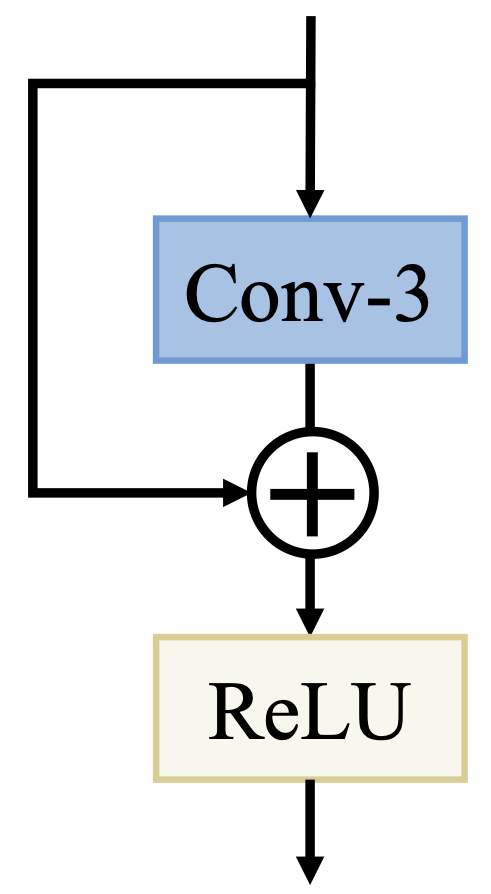} \hspace{\fsdurthreeD} &
            \includegraphics[width=0.24\textwidth]{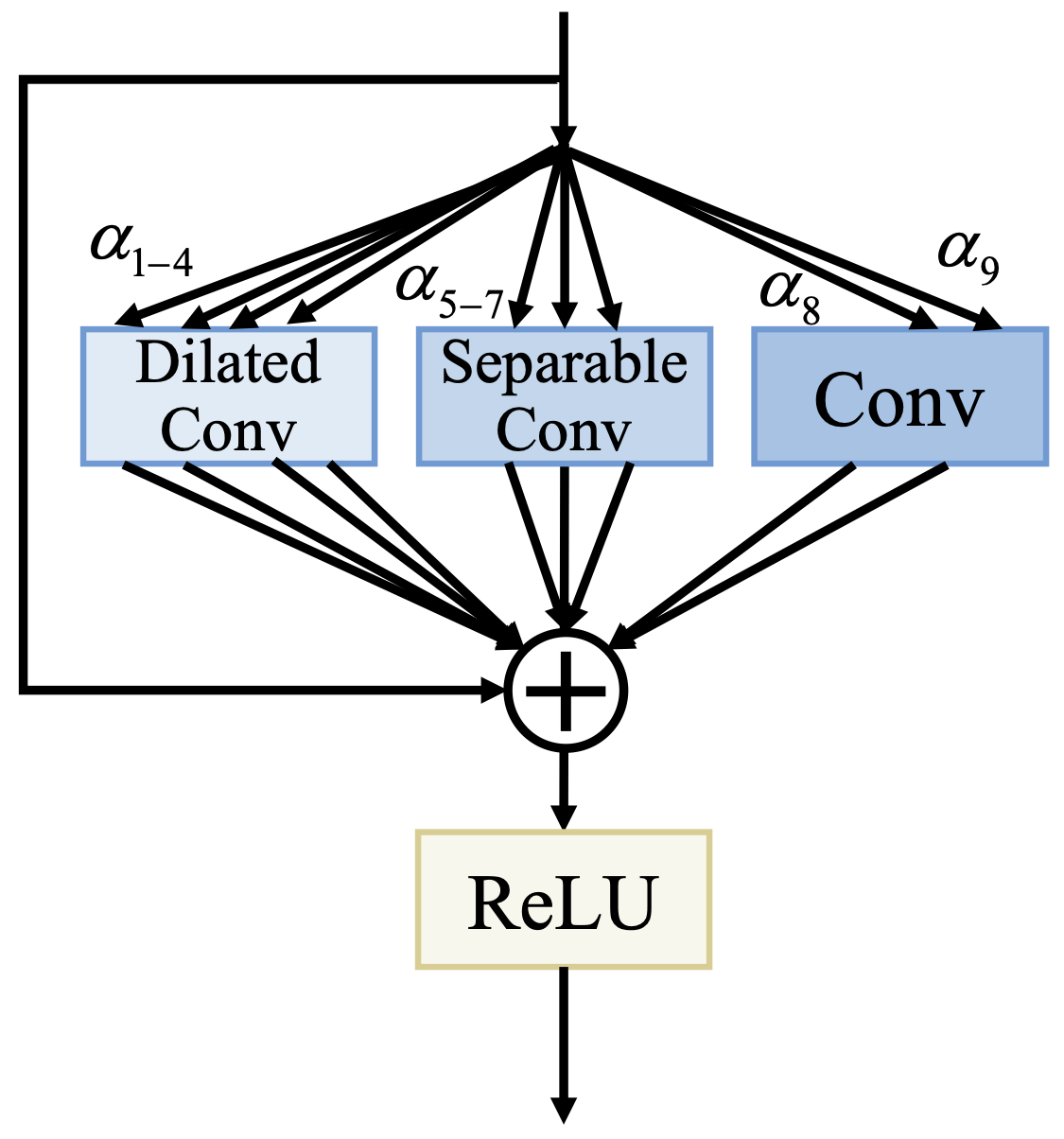} \hspace{\fsdurthreeD}
            \\
            (a) Shallow residual block (SRB) &
			(b) Mixed operations block (Mixed OP)
		\end{tabular}
	\caption{\textit{Express Team:} The shallow residual block in RFDB and mixed residual block that replace a $3\times3$ convolution layer with a mixed layer.}
	\label{fig:mixed_block}
\end{figure}

Existing lightweight SR methods such as IMDN \cite{IMDN} and RFDN~\cite{RFDN} have achieved promising performance with a great equilibrium between performance and parameters or inference speed. However, there is still room for improvement in their network architectures. For instance, the $3\times3$ convolution kernels have been widely adopted by IMDN \cite{IMDN} and RFDN \cite{RFDN}, while its optimality is still questionable. Blocks based on the $3\times3$ convolution kernels may be suboptimal for lightweight SISR tasks. Neural network architecture search (NAS) may be served as an ideal approach. Inspired by DARTS \cite{liu2019darts} and DLSR \cite{DLSR}, the proposed solution is based on DARTS \cite{liu2019darts} and DLSR \cite{DLSR}, which is a fully differentiable NAS method for lightweight SR model. The aim is to find the lightweight network for efficient SR by searching the best replacement of $3 \times 3$ convolution kernels of shallow residual block in RFDB.  Next, the search space, search strategy and the searched network are introduced in sequence.

\begin{table}[t]
\begin{small}
\begin{center}
\begin{tabular}{c|c|c|c}
\toprule
\multirow{1}{5em}{Operation}& Kernel Size  & Params (K) & Muti-Adds (G) \\
\midrule
\multirow{4}{5em}{convolution}&$1 \times 1$&2.5&0.576 \\
& $3 \times 3$&22.5&5.184\\
& $5 \times 5$&62.5&14.400\\
& $7 \times 7$&122.5&28.224\\
\hline
\multirow{3}{5em}{Separable convolution}\!\!& $3 \times 3$&5.9&1.359\\
&$5 \times 5$&7.5&1.728\\
&$7 \times 7$&9.9&2.281\\
\hline
\multirow{2}{5em}{Dilated convolution}\!\!& $3 \times 3$&2.95&0.680\\
&$5 \times 5$&3.75&0.864\\
\bottomrule
\end{tabular}
\end{center}
\caption{\textit{Express Team:} Operations and their complexities in mixed layer. Dilated convolution \cite{yu2017dilated} is applied jointly with group convolution. Muti-Adds are calculated in $\times 2$ SR task with 50 channels on a $1280 \times 720$ image.}
\label{tab:Operations}
\end{small}
\end{table}

\begin{figure*}[t]
	\centering
		\begin{tabular}{ccc}
			\includegraphics[width=0.09\textwidth]{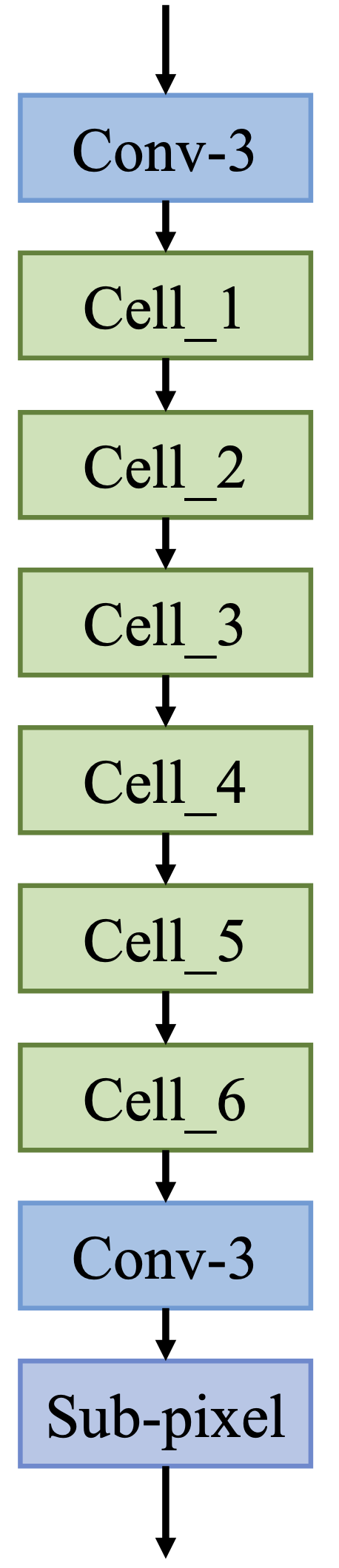}&
			\includegraphics[width=0.3\textwidth]{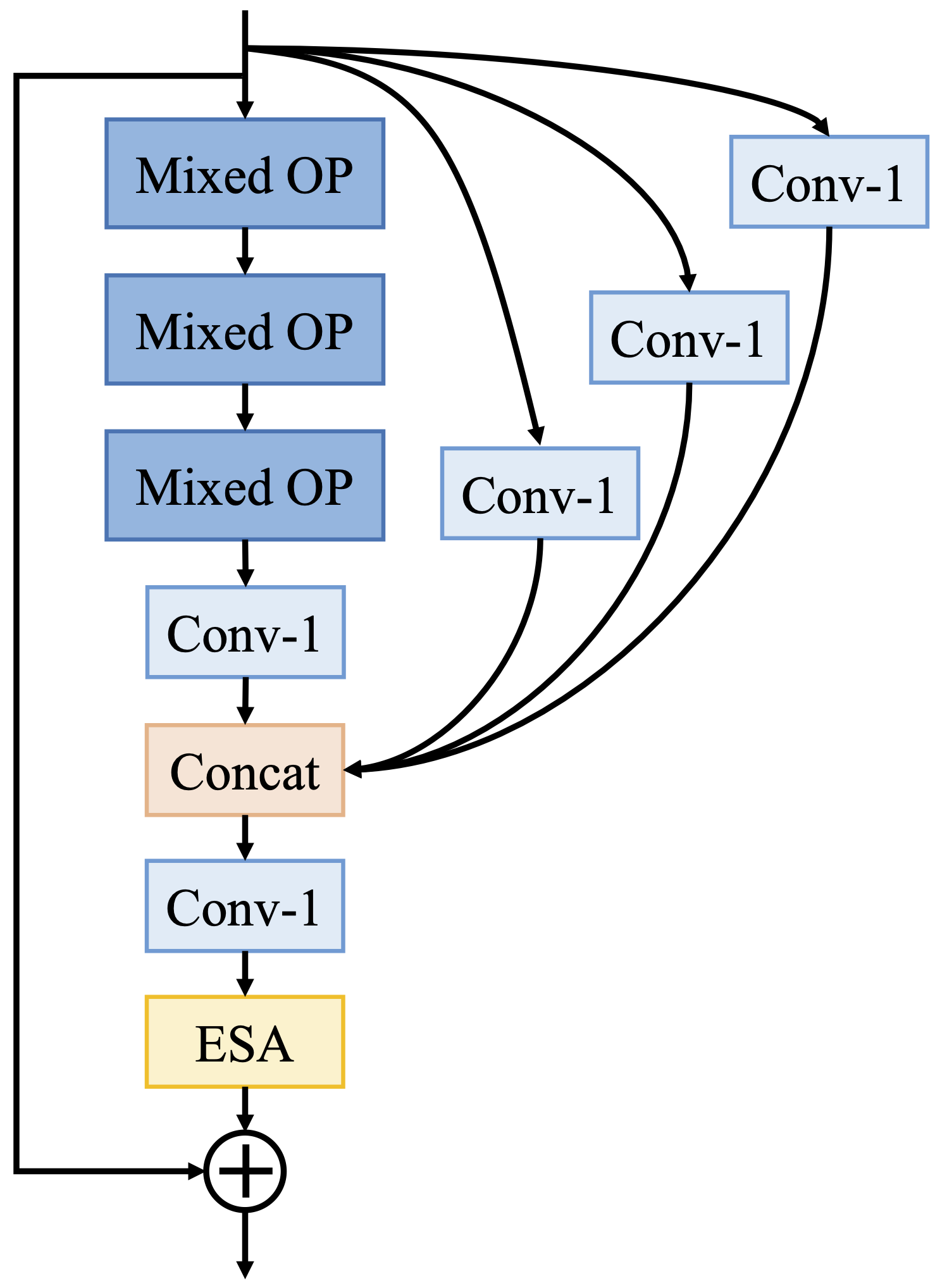}&
            \includegraphics[width=0.3\textwidth]{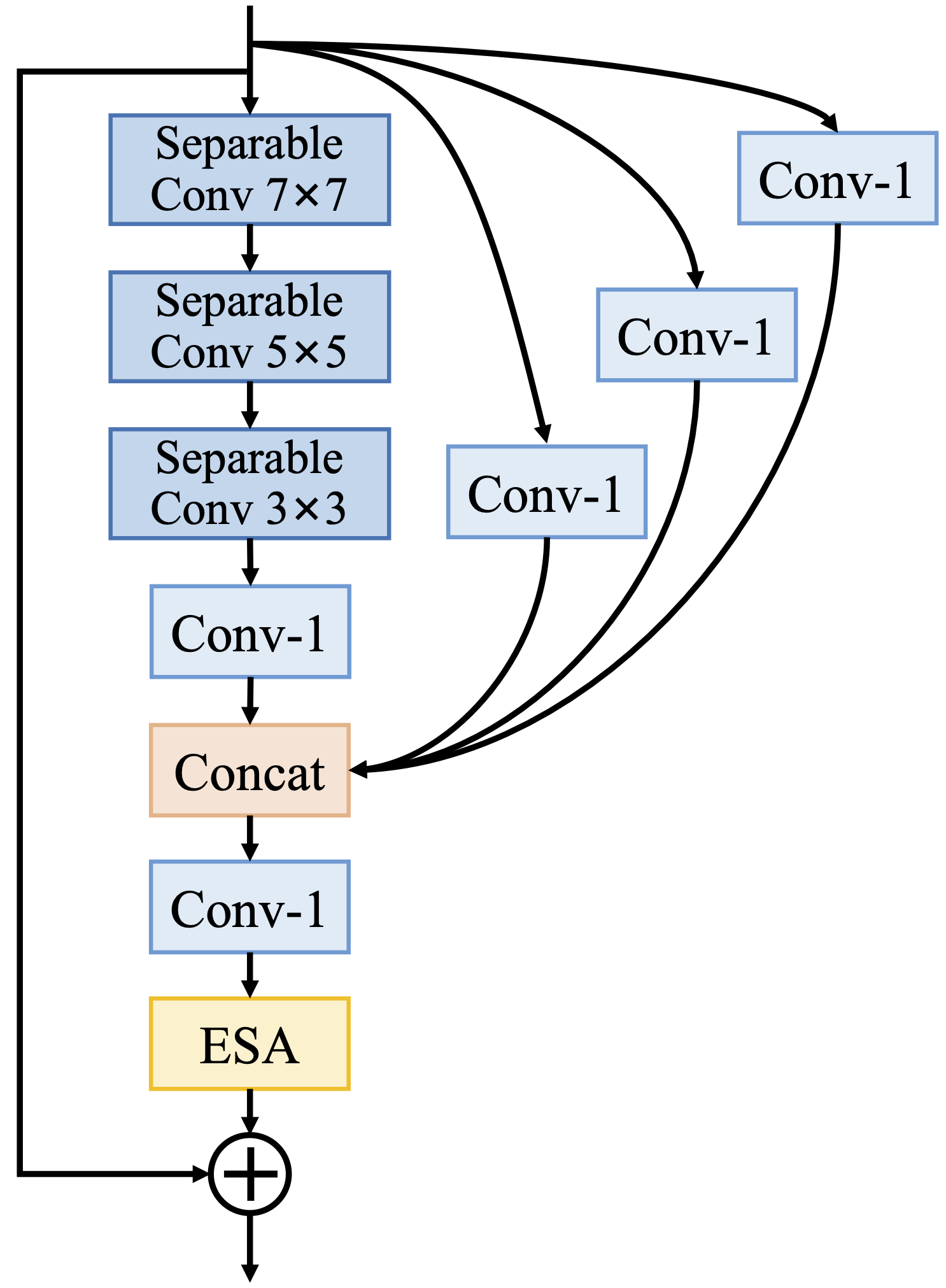}
            \\
            (a) The network structure&
			(b) The cell structure during searching&
			(c) The searched cell structure
		\end{tabular}
	\caption{\textit{Express Team:} The searched cell structure and architecture of network. For brevity, the connection from each cell to the last convolution layer has been omitted.}
	\label{fig:searched_result}
\end{figure*}

\textbf{Search space.} Based on residual feature distillation block of  (RFDB)\cite{RFDN}, the smallest building block, \ie shallow residual blocks (SRB) consist of a $3\times3$ convolution layer and a residual connection as shown in \cref{fig:mixed_block}(a). 
In order to search for a more lightweight structure with competitive performance, the $3\times3$ convolution layer is replaced with a mixed layer as shown in \cref{fig:mixed_block}(b). The mixed layer is composed of multiple operations including separable convolution, dilated convolution, and normal convolution as shown in \cref{tab:Operations}. The input is denoted as $x_k$, and the operation set is denoted as $O$ where each element represents a candidate operation $o(\cdot)$ that is weighted by the architecture parameters $\alpha^k_o$. Then, like DARTS \cite{liu2019darts}, softmax function is used to perform the continuous relaxation of the operation space. Thus, the output of mixed layer $k$ denoted by $f_k(x_k)$ is given as:
\begin{equation}
    f_{k}\left(x_{k}\right)=\sum_{o \in O}\frac{\exp\left(\alpha_{o}^{k}\right)}{\sum_{o^{\prime} \in O} \exp\left(\alpha_{o^{\prime}}^{k}\right)} o\left(x^{k}\right).
\end{equation}
After searching, only the operation with the largest $\alpha^k_o$ is reserved as the best choice of this layer. All three SRBs of each RFDB will be replaced by the searched results. The search space contains $9\times9\times9$ different structures. The network structure and its corresponding cell structure during searching is shown in \cref{fig:searched_result} (a) and \cref{fig:searched_result} (b), respectively.

\textbf{Search strategy.} The differentiable NAS method is applied to the lightweight SISR task. The objective function of the model is defined as
\begin{equation}
    \min _{\theta, \alpha}\left[L_{tr}\left(\theta^{*}(\alpha)+\lambda L_{val}\left(\theta^{*}(\alpha) ; \alpha\right)\right]\right)
\end{equation}
where $\theta$ denotes the weights parameters of the network, and $\lambda$ is a non-negative regularization parameter that balances the importance of the training loss and validation loss. Since the architecture parameter $\alpha$ is continuous, Adam\cite{kingma2014adam} is directly applied to solve problem (2). The parameters $\theta$, $\alpha$ are updated are updated with subsequent iterations:
\begin{equation}
\theta=\theta-\eta_{\theta} \nabla_{\theta} L_{t r}(\theta, \alpha);
\label{eq:theta_update}
\end{equation}
\begin{equation}
    \alpha=\alpha-\eta_{\alpha} \nabla_{\alpha} L_{t r}(\theta, \alpha)+\lambda\nabla_{\alpha} L_{val}(\theta, \alpha).
    \label{eq:A_update}
\end{equation}
The searching and training procedure is summarized in Algorithm 1.

\begin{algorithm}[ht]
\SetKwInput{KwInput}{Input}
\SetKwInput{KwOutput}{Output}
  \caption{\textit{Express Team:} Searching and training algorithm} 
  \label{alg:overall}
  \KwInput{Training set {\small$\mathbb{D}$}}  
  Initialize the super-network {\small$\mathcal{T}$} with architecture parameters {\small$\alpha$}.\\
  Split training set {\small$\mathbb{D}$} into {\small$\mathbb{D}_{train}$} and {\small$\mathbb{D}_{valid}$}.\\
  Train the super-network {\small$\mathcal{T}$} on {\small$\mathbb{D}_{train}$} for several steps to warm up.\\
  \For{{\small$t = 1,2,\ldots, T$}}
  {Sample train batch {\small$\mathbb{B}_{t} = \left \{ {(x_{i}, y_{i})} \right \}_{i=1}^{batch}$} from {\small$\mathbb{D}_{train}$}\\
  Optimize {\small$\theta$} on the {\small$\mathbb{B}_{t}$} by Eq. \eqref{eq:theta_update} \\
  Sample valid batch {\small$\mathbb{B}_{v} = \left \{ {(x_{i}, y_{i})} \right \}_{i=1}^{batch}$} from {\small$\mathbb{D}_{valid}$}\\
  Optimize {\small$\alpha$} on the {\small$\mathbb{B}_{v}$} by Eq. \eqref{eq:A_update} \\
  Save the genotypes of the searched networks
  }
  Train searched networks\\
  \KwOutput{A lightweight SR network {\small$\mathcal{S}$}}  
  \label{alg:algorithm}
\end{algorithm}

\textbf{Searched results.} As shown in \cref{fig:searched_result} (c), the searched cell is composed of a $7\times7$ separable convolution layer, $5\times5$ separable convolution layer, $3\times3$ separable convolution layer, ESA block, and residual connections with information distillation mechanism. Since the number of parameters and FLOPs of the searched results are all fewer than the original $3\times3$ convolution layer, a much smaller (nearly half the original size) model is obtained compared with RFDN \cite{RFDN}.

\textbf{Loss function.} To achieve lightweight and accurate SR models, the loss function is the weighted sum of these three losses: 
\begin{equation}
    L_{1} =\frac{1}{N} \sum_{i=1}^{N}\left|\left(F_{\theta}\left(I^{L R}\right)-I^{HR}\right)\right|;
\end{equation}
\begin{equation}
    L_{HFEN} =\frac{1}{N} \sum_{i=1}^{N}\left|\nabla F_{\theta}\left(I^{LR}\right)-\nabla I^{H R}\right|;
\end{equation}
\begin{equation}
    L_{P} =\sum_{o \in O} \frac{p_{o}}{\sum_{c \in O} p_{c}}\operatorname{softmax}\left(\alpha_{o}\right);
\end{equation}
\begin{equation}
    L(\theta) =L_{1}+\mu \times L_{HFEN}+\gamma \times L_{P}. \label{eqn:express_loss}
\end{equation}
Specifically, $L_{HFEN}$\cite{chaitanya2017interactive} is a gradient-domain $L_1$ loss and can improve image details; $p_o$ denotes the number of parameters of operation $o$ and $L_P$ utilizes them to weigh the architecture parameter $\alpha$, so as to push the algorithm to search for lightweight operations. The $\mu$ and $\gamma$ are weighting parameters that balance the reconstruction performance and model complexity, respectively. When retraining the searched network, set $\gamma=0$ and remove the last item in the total loss function \cref{eqn:express_loss}.

\subsection{Just Try}

\begin{figure*}
    \centering
    \includegraphics[width=1.0\linewidth]{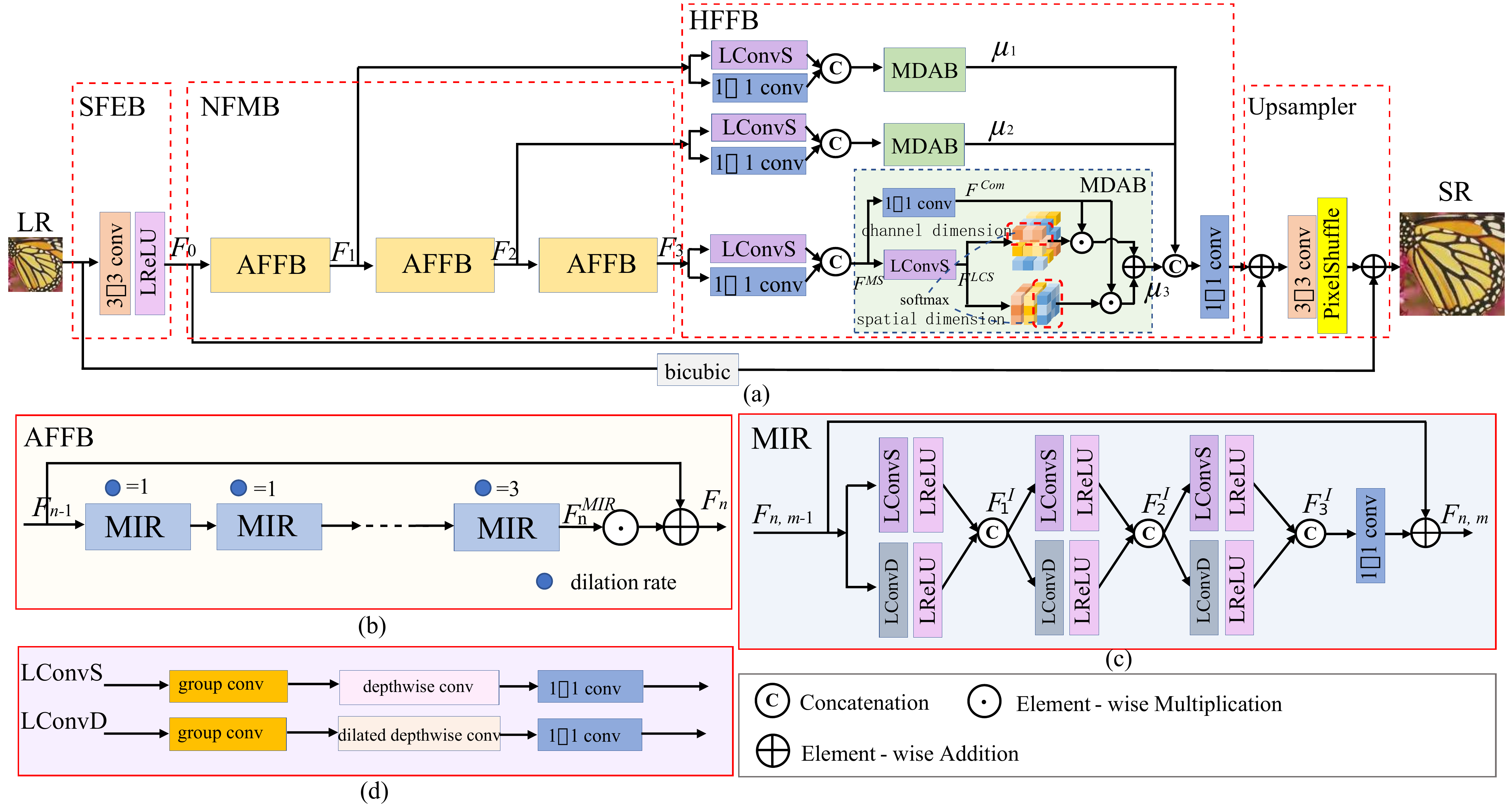}
    \caption{\textit{ncepu\_explorers Team:} (a) The network architecture of the proposed MDAN. (b) Area feature fusion block (AFFB). (c) Multiple interactive residual block (MIR).  (d) Lightweight convolutional units (LConvS / LConvD).}
    \label{fig:MDAN}
\end{figure*}

The team designed a multi-branch network structure LWFANet, The LWFANet extracts the shallow feature with one convolution layer. The extracted feature is sent to the deep feature extraction module which consists of 10 LWFA blocks and a $3 \times 3$ convolution layer. Each LWFA block has four branches, every branch consists of a $1 \times 1$ convolution layer and several $3 \times 3$ convolution layers. The $1 \times 1$ convolutional layer selects the input features and reduces the number of channels to one-fourth of the input channels, the different number of $3 \times 3$ convolutional layers are used to extract the features at different levels. Then multi-level features of every branch are concatenated and used the channel attention mechanism for adaptive aggregation of features $out\_ca$. Then spatial attention is used to get $out\_sa$. The input feature is also enhanced by spatial attention, leading to $x\_sa$. The final output of the LWFA block is obtained sum up $out\_ca$, $out\_sa$ and $x\_sa$. Long skip connections are used to get the final output feature. Then $1 \times 1$ convolution layer is used to reduce the dimension. The upsampling module consists of nearest interpolation and $3 \times 3$ convolution layers.  The reconstructed image is derived after two convolution operations.

\subsection{ncepu\_explorers}

The ncepu\_explorers team proposed the MDAN network architecture shown in \cref{fig:MDAN}, consists of four main parts: shallow feature extraction block (SFEB), nonlinear feature mapping block (NFMB), hierarchical feature fusion block (HFFB), and upsampling block (Upsampler). The SFEB consists of only one 3 $\times$ 3 convolutional layer and one leaky rectified linear unit (LReLU), and the Upsampler uses the sub-pixel convolution. The NFMB cascades $N$ ($N$ = 3) area feature fusion blocks (AFFBs). The HFFB mainly consists of multiple pairs of the lightweight convolutional units (LConvSs) / 1 $\times$ 1 convolutions and multiple multi-dimensional attention blocks (MDABs).

In the MDAN architecture, three AFFBs were cascaded in the NFMB, and the number of input and output channels for each AFFB was 48. Six MIRs were cascaded in each AFFB, and the dilation rates of the dilation convolutions in each MIR were set to 1, 1, 2, 2, 3 and 3. In each MIR, the number of the input channels of both LConvS and LConvD was 48 and the number of the output channels was 24. The group convolutions in both LConvS and LConvD used three groups. The initial values of the learnable parameters $\mu_{1}$, $\mu_{2}$, and $\mu_{3}$ in the HFFB were set to 0.3, 0.3, and 0.4, respectively.

\begin{figure*}
	\centering
	\includegraphics[width=0.8\linewidth]{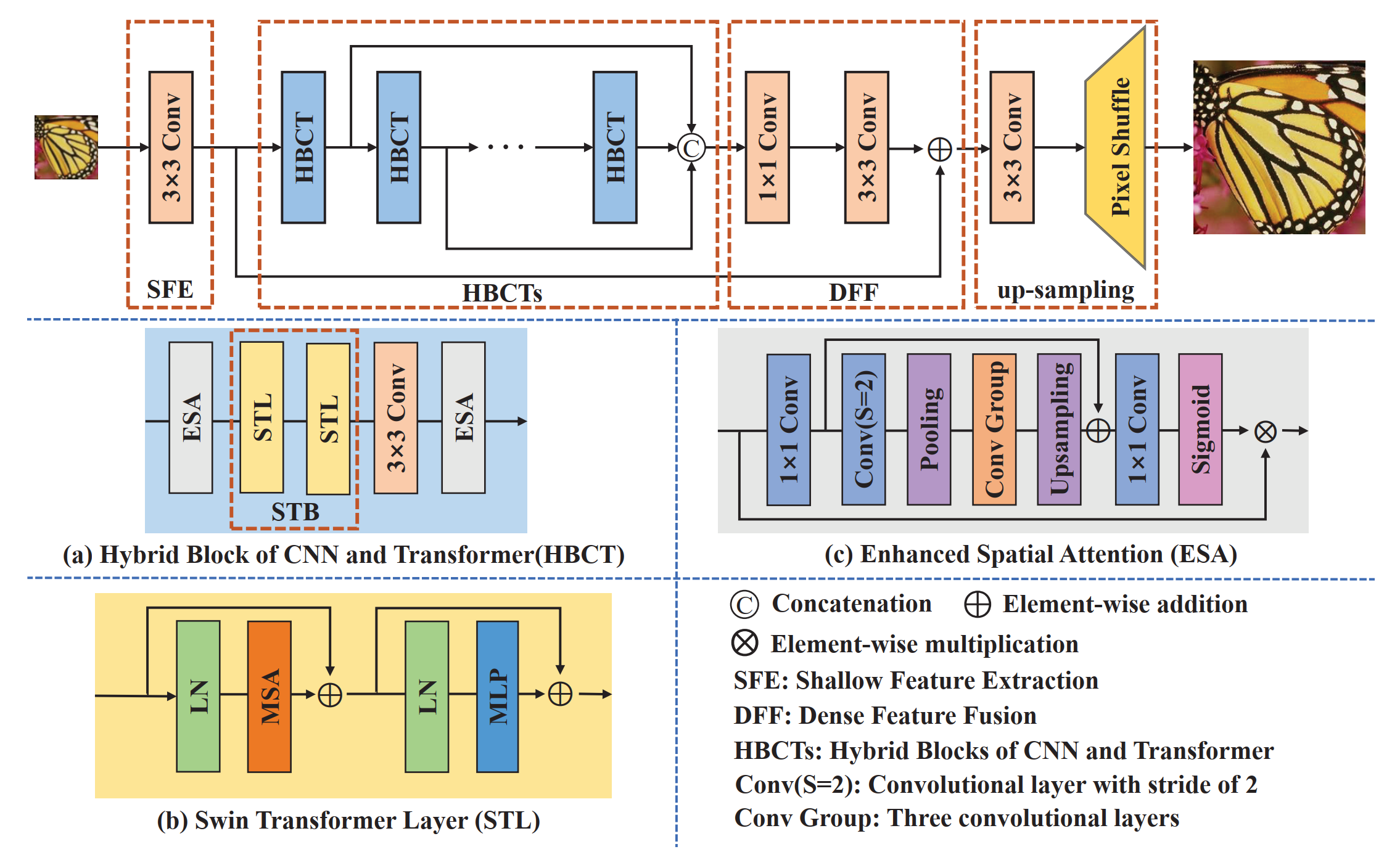}
    \caption{\textit{mju\_mnu Team:} The architecture of the proposed HNCT for lightweight image super-resolution. (a) The module of Hybrid Blocks of CNN and Transformer (HBCTs). (b) Swin Transformer Layer (STL). (c) Enhanced spatial attention module (ESA) proposed in RFANet~\cite{RFANet}.}
    \label{fig:HNCT}
\end{figure*}

\subsection{mju\_mnu}

The team proposed a lightweight SR model namely hybrid
network of CNN and Transformer (HNCT) in \cref{fig:HNCT}, which integrated CNN and transformers to model local and non-local
priors simultaneously. Specifically, HNCT consists of
four parts: shallow feature extraction (SFE) module,
Hybrid Blocks of CNN and Transformer (HBCTs),
dense feature fusion (DFF) module and up-sampling
module. Firstly, shallow features containing low-frequency information are extracted by only one convolution layer in the shallow feature extraction module. Then, four HBCTs are used to extract hierarchical features. Each HBCT contains a Swin Transformer block (STB) with two Swin Transformer layers
inside, a convolutional layer and two enhanced spatial
attention (ESA) modules. Afterwards, these hierarchical features produced by HBCTs are concatenated and
fused to obtain residual features in SFE. Finally, SR
results are generated in the up-sampling module. Integrating CNN and transformer, the HNCT is able to
extract more effective features for SR.

\section*{Acknowledgments}
We thank the NTIRE 2022 sponsors: Huawei, Reality Labs, Bending Spoons, MediaTek, OPPO, Oddity, Voyage81, ETH Zurich (Computer Vision Lab) and University of Wurzburg (CAIDAS).

\appendix

\section{Teams and affiliations}
\label{sec:teams}

\subsection*{NTIRE 2022 team}
\noindent\textit{\textbf{Title: }} NTIRE 2022 Efficient Super-Resolution Challenge\\
\noindent\textit{\textbf{Members: }} \\
Yawei Li$^1$ (\href{mailto:yawei.li@vision.ee.ethz.ch}{yawei.li@vision.ee.ethz.ch}),\\
Kai Zhang$^1$ (\href{mailto:kai.zhang@vision.ee.ethz.ch}{kai.zhang@vision.ee.ethz.ch}),\\
Luc Van Gool$^1$ (\href{mailto:vangool@vision.ee.ethz.ch}{vangool@vision.ee.ethz.ch}),\\
Radu Timofte$^{1,2}$ (\href{mailto:radu.timofte@vision.ee.ethz.ch}{radu.timofte@vision.ee.ethz.ch})\\
\noindent\textit{\textbf{Affiliations: }}\\
$^1$ Computer Vision Lab, ETH Zurich, Switzerland\\
$^2$ University of W\"urzburg, Germany\\

\subsection*{ByteESR}
\noindent\textit{\textbf{Title: }} Residual Local Feature Network For Efficient Super-Resolution\\
\noindent\textit{\textbf{Members: }} \\
Fangyuan Kong (\href{mailto:kongfangyuan@bytedance.com}{kongfangyuan@bytedance.com}),
Mingxi Li, Songwei Liu\\
\noindent\textit{\textbf{Affiliations: }} \\ ByteDance, Shenzhen, China \\
\\

\subsection*{NJU\_Jet}
\noindent\textit{\textbf{Title: }}Fast and Memory-Efficient Network with Window Attention \\
\noindent\textit{\textbf{Members: }} \\
Zongcai Du$^1$ (\href{mailto:151220022@smail.nju.edu.cn}{151220022@smail.nju.edu.cn}),
Ding Liu$^2$\\
\noindent\textit{\textbf{Affiliations: }}\\
$^1$ State Key Laboratory for Novel Software Technology, Nanjing University, China\\
$^2$ ByteDance Inc. \\

\subsection*{NEESR}
\noindent\textit{\textbf{Title: }}Edge-Oriented Feature Distillation Network for Lightweight Image Super Resolution\\
\noindent\textit{\textbf{Members: }}\textit{Chenhui Zhou$^{1}$ \noindent(\href{mailto:username@mail.com}{daomujun@foxmail.com)}}, Jingyi Chen$^{1}$, Qingrui Han$^{1}$\\
\noindent\textit{\textbf{Affiliation: }}\\
$^{1}$ NetEase, Inc.\\

\subsection*{XPixel}
\noindent\textit{\textbf{Title: }}Blueprint Separable Residual Network for Single Image Super-Resolution \\
\noindent\textit{\textbf{Members: }} \\
Zheyuan Li$^{1}$ (\href{mailto:zy.li3@siat.ac.cn}{zy.li3@siat.ac.cn}), Yingqi Liu$^{1}$, Xiangyu Chen$^{1,2}$, Haoming Cai$^{1}$, Yu Qiao$^{3,1}$, Chao Dong$^{3,1}$\\
\noindent\textit{\textbf{Affiliations: }}\\
$^{1}$ Shenzhen Institutes of Advanced Technology, CAS \\
$^{2}$ University of Macau \\
$^{3}$ Shanghai AI Lab, Shanghai, China \\
\\

\subsection*{NJUST\_ESR}
\noindent\textit{\textbf{Title: }}MobileSR: A Mobile-friendly Transformer for Efficient Image Super-Resolution \\
\noindent\textit{\textbf{Members: }} \\
Long Sun$^{1}$ (\href{mailto:cs.longsun@gmail.com}{cs.longsun@gmail.com}), Jinshan Pan$^{1}$, Yi Zhu$^{2}$\\
\noindent\textit{\textbf{Affiliations: }}\\
$^{1}$ Nanjing University of Science and Technology \\
$^{2}$ Amazon Web Services \\
\\

\subsection*{HiImageTeam}
\noindent\textit{\textbf{Title: }}Asymmetric Residual Feature Distillation Network \\
\noindent\textit{\textbf{Members: }} \\
Zhikai Zong (\href{mailto:zzksdu@163.com}{zzksdu@163.com}), Xiaoxiao Liu\\
\noindent\textit{\textbf{Affiliations: }}\\
Qingdao Hi-image Technologies Co.,Ltd (Hisense Visual Technology Co.,Ltd.)
\\

\subsection*{rainbow}
\noindent{\textbf{Title: }} Improved Information Distillation Network for Efficient Super-Resolution\\
\noindent{\textbf{Members: }}{Zheng Hui    
\\\noindent(\href{mailto:huizheng.hz@alibaba-inc.com}{huizheng.hz@alibaba-inc.com})}, Tao Yang, Peiran Ren, Xuansong Xie, Xian-Sheng Hua\\
\noindent{\textbf{Affiliation: }}\\
Alibaba DAMO Academy, EFC, Yuhang District, Hangzhou, Zhejiang, China\\

\subsection*{Super}
\noindent\textit{\textbf{Title: }} Re-parameterized Pixel Attention Feature Distillation Network\\
\noindent\textit{\textbf{Members: }} \\
Yanbo Wang (\href{mailto:51205901021@stu.ecnu.edu.cn}{51205901021@stu.ecnu.edu.cn}), Xiaozhong Ji$^{1,2}$, Chuming Lin$^{2}$, Donghao Luo$^{2}$, Ying Tai$^{2}$, Chengjie Wang$^{2}$,
Zhizhong Zhang$^{1}$, Yuan Xie$^{1}$\\
\noindent\textit{\textbf{Affiliations: }}\\
$^{1}$ East China Normal University \\
$^{2}$ Youtu Lab, Tencent\\

\subsection*{MegSR}
\noindent{\textbf{Title: }} Feature Distillation Network Pruning for Efficient Image Super-Resolution\\
\noindent{\textbf{Members: }}{Shen Cheng$^1$    
\\\noindent(\href{mailto:chengshen@megvii.com}{chengshen@megvii.com})}, Ziwei Luo$^1$, Lei Yu$^1$, Zhihong Wen$^1$, Qi Wu$^1$, Youwei Li$^1$, Haoqiang Fan$^1$,  Jian Sun$^1$ and Shuaicheng Liu$^{2,1}$\\
\noindent{\textbf{Affiliation: }}\\
$^1$ Megvii Technology\\
$^2$ University of Electronic Science and Technology of China\\

\subsection*{VMCL\_Taobao}
\noindent{\textbf{Title: }} Multi-scale Information Distillation Network for Efficient Super-resolution\\
\noindent{\textbf{Members: }}{Yuanfei Huang$^1$    
\\\noindent(\href{mailto:yfhuang@bnu.edu.cn}{yfhuang@bnu.edu.cn})}, Meiguang Jin$^2$, Hua Huang$^1$\\
\noindent{\textbf{Affiliation: }}\\
$^1$ School of Artificial Intelligence, Beijing Normal University\\
$^2$ Alibaba Group\\

\subsection*{Bilibili AI}
\noindent\textit{\textbf{Title: }}RepRFDN: Using Re-parameterization technology into lightweight image super resolution \\
\noindent\textit{\textbf{Members: }} \\
Jing Liu  (\href{mailto:liujing04@bilibili.com}{liujing04@bilibili.com}), Xinjian Zhang\\
\noindent\textit{\textbf{Affiliations: }}\\
Bilibili AI\\

\subsection*{NKU-ESR}
\noindent\textit{\textbf{Title: }} Edge-enhanced Feature Distillation Network for Efficient Super-Resolution\\
\noindent\textit{\textbf{Members: }} \\
Yan Wang (\href{mailto:wyrmy@foxmail.com}{wyrmy@foxmail.com}) \\
\noindent\textit{\textbf{Affiliations: }}\\
Nankai-Baidu Joint Lab, Nankai University, Tianjin, China\\

\subsection*{NJUST\_RESTORATION}
\noindent\textit{\textbf{Title: }} Adaptive Feature Distillation Network for Lightweight Super-Resolution\\
\noindent\textit{\textbf{Members: }} \\
Lingshun Kong (\href{mailto:konglingshun@njust.edu.cn}{konglingshun@njust.edu.cn}), Jinshan Pan\\
\noindent\textit{\textbf{Affiliations: }}\\
Nanjing University of Science and Technology
\\

\subsection*{TOVBU}
\noindent\textit{\textbf{Title: }} Faster Residual Feature Distillation Network for Efficient Super Resolution\\
\noindent\textit{\textbf{Members: }} \\
Gen Li (\href{mailto:leegeun@yonsei.ac.kr}{leegeun@yonsei.ac.kr}), Yuanfan Zhang, Zuowei
Cao, Lei Sun\\
\noindent\textit{\textbf{Affiliations: }}\\
Platform Technologies, Tencent Online Video\\

\subsection*{Alpan}
\noindent{\textbf{Members: }}{Panaetov Alexander 
\\\noindent(\href{mailto:aapanaetov@edu.hse.ru}{aapanaetov@edu.hse.ru})}\\
\noindent{\textbf{Affiliation: }}\\
Higher School of Economics (@edu.hse.ru), Huawei Moscow Research Center (@huawei.com)\\

\subsection*{Dragon}
\noindent{\textbf{Title: }} Double Branch Network With Enhanced Spatial Attention\\
\noindent{\textbf{Members: }}{Yucong Wang     
\\\noindent(\href{mailto:1401121556@qq.com}{1401121556@qq.com})}, Minjie Cai\\
\noindent{\textbf{Affiliation: }}\\
Hunan University\\

\subsection*{TieGuoDun Team}
\noindent\textit{\textbf{Title: }} \\
\noindent\textit{\textbf{Members: }} \\
Shuhao Zhang (\href{mailto:zhangshuha0@163.com}{zhangshuha0@163.com}), Yuhao Zhang\\
\noindent\textit{\textbf{Affiliations: }}\\
Xidian University\\

\subsection*{xilinxSR}
\noindent\textit{\textbf{Title: }} Efficient Image Super-Resolution with Collapsible Linear Blocks\\
\noindent\textit{\textbf{Members: }} \\
Li Wang  (\href{mailto:liwa@xilinx.com}{liwa@xilinx.com}), Lu Tian\\
\noindent\textit{\textbf{Affiliations: }}\\
Xilinx Technology Beijing Limited\\

\subsection*{cipher}
\noindent{\textbf{Title: }} ResDN: Residual Distillation Network for Single Image Super-Resolution\\
\noindent{\textbf{Members: }}{Zheyuan Wang$^1$    
\\\noindent(\href{mailto:cipherwon@163.com}{cipherwon@163.com})}, Hongbing Ma$^2$  \\
\noindent{\textbf{Affiliation: }}\\
$^1$ College of Information Science and Engineering, Xinjiang University, Urumqi,China\\
$^2$ Department of Electronic Engineering, Tsinghua University,
 Beijing,China\\
 
\subsection*{NJU\_MCG}
\noindent\textit{\textbf{Title: }} Feature Distillation and Expansion Network (FDEN)\\
\noindent\textit{\textbf{Members: }} \\
Jie Liu (\href{mailto:jieliu@smail.nju.edu.cn}{jieliu@smail.nju.edu.cn}), Chao Chen, Yidong Cai, Jie Tang, Gangshan Wu\\
\noindent\textit{\textbf{Affiliations: }}\\
Nanjing University
\\

\subsection*{IMGWLH}
\noindent{\textbf{Members: }}{Weiran Wang$^1$    
\\\noindent(\href{mailto:Wangweirantx@163.com}{Wangweirantx@163.com})}, Shirui Huang, Honglei Lu, Huan Liu, Keyan Wang, Jun Chen  \\
\noindent{\textbf{Affiliation: }}\\
Xidian University\\
McMaster University\\

\subsection*{imgwhl}
\noindent{\textbf{Title: }} Residual Feature Extraction Super-Resolution\\
\noindent{\textbf{Members: }}{Shirui Huang$^{1}$     
\\\noindent(\href{mailto:shiruihh@gmail.com}{shiruihh@gmail.com})}, Weiran Wang, Honglei Lu, Huan Liu, Keyan Wang, Jun Chen\\
\noindent{\textbf{Affiliation: }}\\
$^{1}$ School of Telecommunication Engineering, Xidian University, Xi'an, China\\
$^{2}$ McMaster University\\

\subsection*{whu\_sigma}
\noindent\textit{\textbf{Title: }}Feature Distillation Network of Dilated Convolution for Lightweight Image Super-Resolution \\
\noindent\textit{\textbf{Members: }} \\
Shi Chen$^{1}$ (\href{mailto:chenshi@whu.edu.cn}{chenshi@whu.edu.cn}), Yuchun Miao$^{2}$, Zimo Huang$^{3}$, Lefei Zhang$^{1}$ \\
\noindent\textit{\textbf{Affiliations: }}\\
$^{1}$ School of Computer Science, Wuhan University, Wuhan, China\\
$^{2}$ School of Mathematical Science, University of Electronic Science and Technology of China, Chengdu, China\\
$^{3}$ School of Computer Science, The University of Sydney, Sydney, Australia\\\\

\subsection*{Aselsan Research}
\noindent\textit{\textbf{Title: }} IMDeception: Grouped Information Distilling Super-Resolution Network\\
\noindent\textit{\textbf{Members: }} \\
Mustafa Ayazoğlu (\href{mailto:mayazoglu@aselsan.com.tr}{mayazoglu@aselsan.com.tr}), \\
\noindent\textit{\textbf{Affiliations: }}\\
Aselsan Research, Ankara, Turkey\\

\subsection*{Drinktea}
\noindent\textit{\textbf{Title: }} Attention Augmented lightweight network\\
\noindent\textit{\textbf{Members: }} \\
Wei Xiong (\href{mailto:scun2016@163.com}{scun2016@163.com}), Chengyi Xiong, Fei Wang\\
\noindent\textit{\textbf{Affiliations: }}\\
School of Electronic and Information Engineering, South-Central University for Nationalities, Wuhan, China\\

\subsection*{GDUT\_SR}
\noindent\textit{\textbf{Title: }}Progressive Representation Re-Calibration Network for Lightweight
Super-Resolution \\
\noindent\textit{\textbf{Members: }} \\
Hao Li (\href{mailto:lihao9605@gmail.com}{lihao9605@gmail.com}), Ruimian Wen, Zhijing Yang\\
\noindent\textit{\textbf{Affiliations: }}\\
Guangdong University of Technology\\

\subsection*{Giantpandacv}
\noindent{\textbf{Members: }}{Wenbin Zou     
\\\noindent(\href{alexzou14@foxmail.com}{alexzou14@foxmail.com})}, Weixin Zheng, Tian Ye, Yuncheng Zhang\\
\noindent{\textbf{Affiliation: }}\\
Fujian Normal University, Fuzhou University, Jimei University, China Design Group\\

\subsection*{neptune}
\noindent\textit{\textbf{Title: }} \\
\noindent\textit{\textbf{Members: }} \\
Xiangzhen Kong (\href{mailto:neptune.team.ai@gmail.com}{neptune.team.ai@gmail.com}), \\
\\

\subsection*{TeamInception}
\noindent{\textbf{Title: }} Restormer: Efficient Transformer for Image Super-Resolution\\
\noindent{\textbf{Members: }}{Aditya Arora$^{1}$     
\\\noindent(\href{mailto:adityadvlp@gmail.com}{adityadvlp@gmail.com})}, Syed Waqas Zamir$^{1}$, Salman Khan$^{3}$, Munawar Hayat$^{2}$, Fahad Shahbaz Khan$^{3}$\\
\noindent{\textbf{Affiliation: }}\\
$^{1}$ Inception Institute of Artificial  Intelligence (IIAI), Abu Dhabi, UAE\\
$^{2}$ Monash University, Melbourne, Australia\\
$^{3}$ Mohamed bin Zayed University of AI\\

\subsection*{cceNBgdd}
\noindent\textit{\textbf{Title: }} A Very Lightweight and Efficient Image Super-Resolution Network\\
\noindent\textit{\textbf{Members: }} \\
Dandan Gao (\href{mailto:gdd@ncepu.edu.cn}{gdd@ncepu.edu.cn}), Dengwen Zhou
\\
\noindent\textit{\textbf{Affiliations: }}\\
North China Electric Power University, Changping District, Beijing\\

\subsection*{Express}
\noindent\textit{\textbf{Title: }} Searching Lightweight Network for Efficient Super-resolution\\
\noindent\textit{\textbf{Members: }} \\
Qian Ning (\href{mailto:ningqian@stu.xidian.edu.cn}{ningqian@stu.xidian.edu.cn}), Jingzhu Tang, Han Huang, Yufei Wang, Zhangheng Peng\\
\noindent\textit{\textbf{Affiliations: }}\\
The School of Artificial Intelligence of Xidian University\\

\subsection*{Just Try}
\noindent\textit{\textbf{Title: }} Light Weight Feature Aggregation for Image Super-Resolution\\
\noindent\textit{\textbf{Members: }} \\
Haobo Li$^{1}$  (\href{mailto:qwerdf20191024@gmail.com}{qwerdf20191024@gmail.com}), Wenxue Guan1$^{1}$, Shenghua Gong$^{2}$, Xin Li$^{1}$, Jun Liu$^{1,2}$\\
\noindent\textit{\textbf{Affiliations: }}\\
$^{1}$ College of Computer Science and Technology, Jilin University\\
$^{2}$ School of Electronic and Information Engineering, Beihang University\\

\subsection*{ncepu\_explorers}
\noindent\textit{\textbf{Title: }} MDAN\\
\noindent\textit{\textbf{Members: }} \\
Wanjun Wang (\href{mailto:1206371055@qq.com}{1206371055@qq.com}), Dengwen Zhou\\
\noindent\textit{\textbf{Affiliations: }}\\
School of Control and Computer Engineering, North China Electric Power University, Beijing, China\\

\subsection*{mju\_mnu}
\noindent\textit{\textbf{Title: }} Hybrid network of CNN and Transformer for Lightweight Image Super-Resolution\\
\noindent\textit{\textbf{Members: }} \\
Kun Zeng$^{1}$ (\href{mailto:zengkun301@aliyun.com}{zengkun301@aliyun.com}), Hanjiang Lin$^{2}$, Xinyu Chen; Jinsheng Fang\\
\noindent\textit{\textbf{Affiliations: }}\\
$^{1}$ Minnan Normal University, Zhangzhou, Fujian, China\\
$^{2}$ Minjiang University, Fuzhou, Fujian, China\\

\subsection*{Virtual\_Reality}
\noindent\textit{\textbf{Title: }} Non-Local Fourier Convolution for Efficient Image Super Resolution\\
\noindent\textit{\textbf{Members: }} \\
Abhishek Kumar Sinha (\href{mailto:aks@sac.isro.gov.in}{aks@sac.isro.gov.in}), S. Manthira Moorthi, Debajyoti Dhar \\
\noindent\textit{\textbf{Affiliations: }}\\
Space Applications Centre, Ahmedabad\\

\subsection*{NTU607QCO-ESR}
\noindent\textit{\textbf{Title: }} Re-parameterized and pruned model for Efficient SR\\
\noindent\textit{\textbf{Members: }} \\
Hao-Hsiang Yang$^{1}$ (\href{mailto:islike8399@gmail.com}{islike8399@gmail.com}), Zhi-Kai Huang$^{1}$, Wei-Ting Chen$^{2}$, Hua-En Chang$^{1}$, Sy-Yen Kuo$^{1}$ \\
\noindent\textit{\textbf{Affiliations: }}\\
$^{1}$ Department of Electrical Engineering, National Taiwan University, Taiwan \\ 
$^{2}$ Graduate Institute of Electronics Engineering, National Taiwan University, Taiwan\\

\subsection*{Strong Tiger}
\noindent\textit{\textbf{Title: }} Multi-Directional Gradient Network for Real-time Super Resolution
\\
\noindent\textit{\textbf{Members: }} \\
Wei Tan (\href{mailto:tanwei0699@163.com}{tanwei0699@163.com}), \\
\noindent\textit{\textbf{Affiliations: }}\\
DAMO Academy, Alibaba Group\\

\subsection*{VAP}
\noindent{\textbf{Title: }} CL-RFDN: Collapsible Lightweight Residual Feature Distillation Network for Efficient Image Super-Resolution\\
\noindent{\textbf{Members: }}{Hao Chen     
\\\noindent(\href{mailto:xu.qian5@zte.com.cn}{xu.qian5@zte.com.cn})}, Qian Xu\\
\noindent{\textbf{Affiliation: }}\\
ZTE, Nanjing, China\\

\subsection*{Multicog}
\noindent{\textbf{Title: }} Modified Residual Feature Distillation Network\\
\noindent{\textbf{Members: }}{Pratik Narang$^{1}$     
\\\noindent(\href{mailto:pratik.narang@pilani.bits-pilani.ac.in}{pratik.narang@pilani.bits-pilani.ac.in})}, Usneek Singh$^{1}$, Syed Sameen$^{1}$, Harsh Khaitan$^{2}$\\
\noindent{\textbf{Affiliation: }}\\
$^{1}$ BITS Pilani, Pilani, Rajasthan\\
$^{2}$ Kwikpic Tech. Services, Kolkata, India\\

\subsection*{Set5baby}
\noindent{\textbf{Title: }} Self Residual Feature Distillation Network\\
\noindent{\textbf{Members: }}{Liu Yinghua$^{1}$     
\\\noindent(\href{mailto:727630081@qq.com}{727630081@qq.com})}, Zhang Tianlin$^{2}$, Zhang Xiaoming$^{3}$\\
\noindent{\textbf{Affiliation: }}\\
$^{1}$ Computer Vision Institute, Shenzhen University, Shenzhen, China\\
$^{2}$ CAS Key Laboratory of Electronic and Information Technology for Complex Aerospace Systems, National
Space Science Center, Chinese Academy of Science (CAS)\\
$^{3}$
Institue of Artificial Intelligence, Southwest Jiaotong University\\

\subsection*{NWPU SweetDreamLab}
\noindent{\textbf{Title: }} Branch Mixed Distillation Network for Efficient Super-Resolution\\
\noindent{\textbf{Members: }}{Dingxuan Meng     
\\\noindent(\href{mailto:mdxgalaxy20@gmail.com}{mdxgalaxy20@gmail.com})}, Chunwei Tian\\
\noindent{\textbf{Affiliation: }}\\
Northwestern Polytechnical University\\

\subsection*{SSL}
\noindent{\textbf{Title: }} RFDNeXt\\
\noindent{\textbf{Members: }}{Mashrur M. Morshed     
\\\noindent(\href{mailto:mashrurmorshed@iutdhaka.
edu}{mashrurmorshed@iutdhaka.
edu})}, Ahmad Omar Ahsan\\
\noindent{\textbf{Affiliation: }}\\
Islamic University of Technology,
Dhaka, Bangladesh\\

{\small
\bibliographystyle{ieee_fullname}
\bibliography{main}
}

\end{document}